\documentclass[letterpaper]{article} 
\usepackage{aaai25}  
\usepackage{times}  
\usepackage{helvet}  
\usepackage{courier}  
\usepackage[hyphens]{url}  
\usepackage{graphicx} 
\usepackage{natbib}  
\usepackage{caption} 
\usepackage{algorithm}
\usepackage{algorithmic}

\usepackage{xcolor}
\newcommand{\answerYes}[1]{\textcolor{blue}{#1}} 
\newcommand{\answerNo}[1]{\textcolor{teal}{#1}} 
\newcommand{\answerNA}[1]{\textcolor{gray}{#1}}

\usepackage{amsmath}
\usepackage{booktabs} 
\usepackage{tcolorbox}
\usepackage{bm}
\usepackage{multirow}
\usepackage{subcaption}
\usepackage{enumitem}

\usepackage{newfloat}
\usepackage{listings}
\DeclareCaptionStyle{ruled}{labelfont=normalfont,labelsep=colon,strut=off} 
\floatstyle{ruled}
\newfloat{listing}{tb}{lst}{}
\floatname{listing}{Listing}
\title{Model Hubs and Beyond: Analyzing Model Popularity, Performance, and Documentation}
\author {
    Pritam Kadasi\textsuperscript{\rm 1},
    Sriman Reddy Kondam\textsuperscript{\rm 1},
    Srivathsa Vamsi Chaturvedula\textsuperscript{\rm 1},
    Rudranshu Sen\equalcontrib\textsuperscript{\rm 2}, \\
    Agnish Saha\equalcontrib\textsuperscript{\rm 2}, 
    Soumavo Sikdar\equalcontrib\textsuperscript{\rm 2},
    Sayani Sarkar\equalcontrib\textsuperscript{\rm 2},
    Suhani Mittal\textsuperscript{\rm 4}\thanks{Work done while at IIT Gandhinagar as part of B.Tech project.},
    Rohit Jindal\textsuperscript{\rm 3}, \\
    Mayank Singh\textsuperscript{\rm 1}
}
\affiliations {
    \textsuperscript{\rm 1}Indian Institute of Technology Gandhinagar, Gujarat, India\\
    \textsuperscript{\rm 2}Jadavpur University, West Bengal, India, \\
    \textsuperscript{\rm 3}Vellore Institute of Technology, Tamil Nadu, India \\
    \textsuperscript{\rm 4}Citicorp Services India Private Limited, Pune, Maharashtra, India \\
    \{pritam.k, singh.mayank\}@iitgn.ac.in
}

\begin{document}

\maketitle

\begin{abstract}
 With the massive surge in ML models on platforms like Hugging Face, users often lose track and struggle to choose the best model for their downstream tasks, frequently relying on model popularity indicated by download counts, likes, or recency. We investigate whether this popularity aligns with actual model performance and how the comprehensiveness of model documentation correlates with both popularity and performance. In our study, we evaluated a comprehensive set of 500 Sentiment Analysis models on Hugging Face. This evaluation involved massive annotation efforts, with human annotators completing nearly 80,000 annotations, alongside extensive model training and evaluation. Our findings reveal that model popularity does not necessarily correlate with performance. Additionally, we identify critical inconsistencies in model card reporting: approximately 80\% of the models analyzed lack detailed information about the model, training, and evaluation processes. Furthermore, about 88\% of model authors overstate their models' performance in the model cards. Based on our findings, we provide a checklist of guidelines for users to choose good models for downstream tasks.
\end{abstract}

%
\begin{links}
    \link{Datasets}{https://github.com/pskadasi/model-hubs}
\end{links}
\section{Introduction}
\label{sec:intro}

In the aftermath of the groundbreaking Transformers paper \cite{NIPS2017_3f5ee243}, the field of Natural Language Processing (NLP) has undergone a rapid evolution, witnessing an unprecedented surge in research and development. This surge has given rise to a diverse array of models spanning various AI domains, including both vision \cite{10.1145/3505244} and NLP \cite{wolf-etal-2020-transformers} tasks. To facilitate the dissemination and utilization of these models, researchers have established ``Model Hubs'' -- platforms designed to curate and categorize collections of open models and datasets, fostering sharing and collaboration for a wide range of open models. Among these hubs, the Hugging Face (HF)\footnote{\url{https://huggingface.co/}} stands out as the largest model repository on the internet, boasting over 930K models as of September $10^{\text{th}}$, 2024 and expected to surpass 1M mark\footnotemark.

\footnotetext{\url{https://x.com/ClementDelangue/status/1674709427557998593?s=20}}

This surge in model availability can be attributed to the influence of social media hype, leading to a significant increase in model downloads/likes. However, this phenomenon resembles the ``rich get richer'' dynamic \cite{barabasi}, where popular\footnote{Throughout the rest of the manuscript, we commonly refer to models as ``popular'' based on their \#downloads and \#likes. Essentially, the more \#downloads and \#likes a model has, the more popular it is perceived to be.} models attract more attention and downloads, perpetuating their dominance. Despite the widespread adoption of these models, users often engage with them in downstream tasks without conducting rigorous verification and validation, raising concerns about their reliability and performance \cite{10.1109/ICSE48619.2023.00206}. To investigate this, we evaluate models with high download and like counts for a specific task -- ``Sentiment Analysis,'' on HF. We expand our evaluation to include analysis on different dataset difficulty categories~\cite{swayamdipta-etal-2020-dataset}. Our goal is to explore how models obtained from HF perform across various dataset difficulty categories, considering that users may utilize these downloaded models in different contexts.

Furthermore, in response to the growing need for transparency, accountability, and reproducible NLP research, researchers have introduced datasheets \cite{gebru2021datasheets} and model cards \cite{10.1145/3287560.3287596} as mechanisms to provide detailed insights into model characteristics and associated datasets. However, despite the availability of model cards and datasheets as a platform for authors to furnish comprehensive information about their models and datasets, a substantial number of authors fail to provide sufficient details, leading to a lack of transparency and trustworthiness \cite{Singh2023UnlockingMI, liu2024automatic}. The absence of crucial details in model cards not only hampers reproducibility but also hinders the progress of NLP research. To address these concerns, we conduct a thorough manual inspection of model cards for models with high download/like counts on HF, uncovering instances of missing information. We further extend this investigation to understand the impact of missing information in model cards on their download/like count and to determine its correlation with the model's performance.

Given the prevalence of models on platforms like HF, it is pertinent to investigate whether these platforms serve merely as dumping grounds\footnote{\url{https://x.com/sd_marlow/status/1674967110739435525?s=20}}\footnote{\url{https://x.com/KastelBert/status/1674761863798693892?s=20}} for models or if they genuinely contribute to advancing NLP research. Our investigation aims to shed light on these challenges and prompt discussions on the importance of transparent documentation, open models, and reproducible research practices in the NLP. 

Overall, our contributions can be summarized as follows: 

\begin{enumerate}
    \item We thoroughly evaluate the 500 most downloaded and liked models, specifically the top-ranked models based on the \#downloads, which we refer to as $M_{500}$, for the Sentiment Analysis (SA) task from HF to determine if high popularity, as represented by \#downloads and \#likes, actually guarantees better model performance. Our key findings reveal that model performance showed a weak positive correlation and, in some cases, even exhibited a negative correlation with download and like counts.

    \item Further, we conduct a large-scale manual study to verify the completeness of information in Model Cards for $M_{500}$ by carefully annotating every element in each section, requiring massive annotation efforts. Our investigation reveals that more than half of the $M_{500}$ lack key details about the training process, and a striking 96\% neglect to provide any information about critical aspects such as Bias, Risks, and Limitations. We observe a high popularity of highly documented models and vice versa.
    
    \item To the best of our knowledge, this is the first large-scale exhaustive study conducted on models from the HF platform. Additionally, we create a novel Reddit SA dataset to evaluate the performance of top models on a completely unseen dataset. The results reveal poor generalization performance and a weak positive correlation with popularity. 
\end{enumerate}

\section{Related Works}

Prior studies have examined various aspects of the HF platform, including pre-trained models (PTMs) reuse and popularity's influence on adoption \cite{10.1109/ICSE48619.2023.00206, Taraghi2024DeepLM}. Other works have assessed platform vulnerabilities \cite{10297271}, environmental impact \cite{10304801}, and ML model maintenance \cite{Castao2023AnalyzingTE}. Additionally, efforts have been made to categorize PTMs for software engineering tasks, such as automatic classification using a public HF dump \cite{disipio2024automated}. In contrast, our study investigates the relationship between model popularity and performance on HF, questioning whether high popularity guarantees better performance.

Related works \cite{10.1145/3287560.3287596, 10.1145/3531146.3533108} have proposed model cards for experts and non-experts, while \cite{liu2024automatic} addresses information incompleteness in model cards and proposes an automated generation approach using LLMs. Another study \cite{10.1145/3643916.3644412} has investigated the transparency of pre-trained transformer models on HF, highlighting limited exposure of training datasets and biases. Our work differs by examining how information incompleteness in model cards influences model popularity and performance. In summary, our research evaluates popular HF models, exploring their relationship with model card completeness and performance.

\section{Experiments}
\label{sec:exps}

\subsection{Models Filtering Criteria}
\label{sec:resource}

We filter the most popular models on HF focusing on text classification models within NLP due to their abundance\footnote{\url{https://huggingface.co/tasks}}. Specifically, within the text classification task, we opt for SA as a representative use case because of its extensive exploration in both industry and academia. We use HFApi\footnote{\url{https://huggingface.co/docs/huggingface_hub/package_reference/hf_api}} to extract metadata for all text classification models, searching for the keyword ``sentiment'' in their model names. During our exploration of HF platform, we identify models like \texttt{ProsusAI/finbert}, which, despite lacking the explicit ``sentiment'' keyword, are SA models. To cover such cases comprehensively, we broaden our search (using HFApi) to models without the sentiment keyword. Within these models, we identify SA models by manually examining their output labels (refer to \S\ref{apx:mf}). For our experiments, we confine our analysis to three-class SA models to maintain simplicity and avoid complexity introduced by models with more than three class labels. Following this process, we identify and include 1207 models\footnotemark that meet our criteria (refer to \S\ref{apx:mf} for more details).

\footnotetext{These are the models filtered as of August $26^{\text{th}}$, 2024}

Continuing with our analysis, we employ a ranking system \(r\) based on metrics such as \#downloads, \#likes, and the date (last modified) of the model. We determine the rank \(r\) based on these metrics. Specifically, for downloads and likes, we calculate \(r_{d}\) and \(r_{l}\) by arranging the number of downloads and likes in descending order and assigning ranks from 1 to 1207. Similarly, for the date, we calculate \(r_{da}\) by arranging the dates in descending order, ensuring that more recent dates receive higher ranks.

Subsequently, we select the top 500 models based on \(r_{d}\), out of the three possible ranks (\(r_{d}\), \(r_{l}\), and \(r_{da}\)) due to its wide adoption~\cite{10.1109/ICSE48619.2023.00206}, forming the collection \(M_{500}\) (refer to Table~\ref{tab:m100_models} in \S\ref{sec:appendix}). We select the top 500 models for our analysis, as the first 100 alone account for over 99.99\% of the total downloads among the 1207 SA models. While these top models dominate usage, extending the scope to 500 models allowed for broader coverage. Models beyond this threshold have extremely low download counts, following a power-law distribution, and thus contribute minimally to the NLP community. Therefore, we restrict our analysis to the top 500 models.

\subsection{Model Card Inspection}
\label{sec:mcards}
\begin{figure*}[t]
     \centering
     \includegraphics[width=0.7\textwidth]{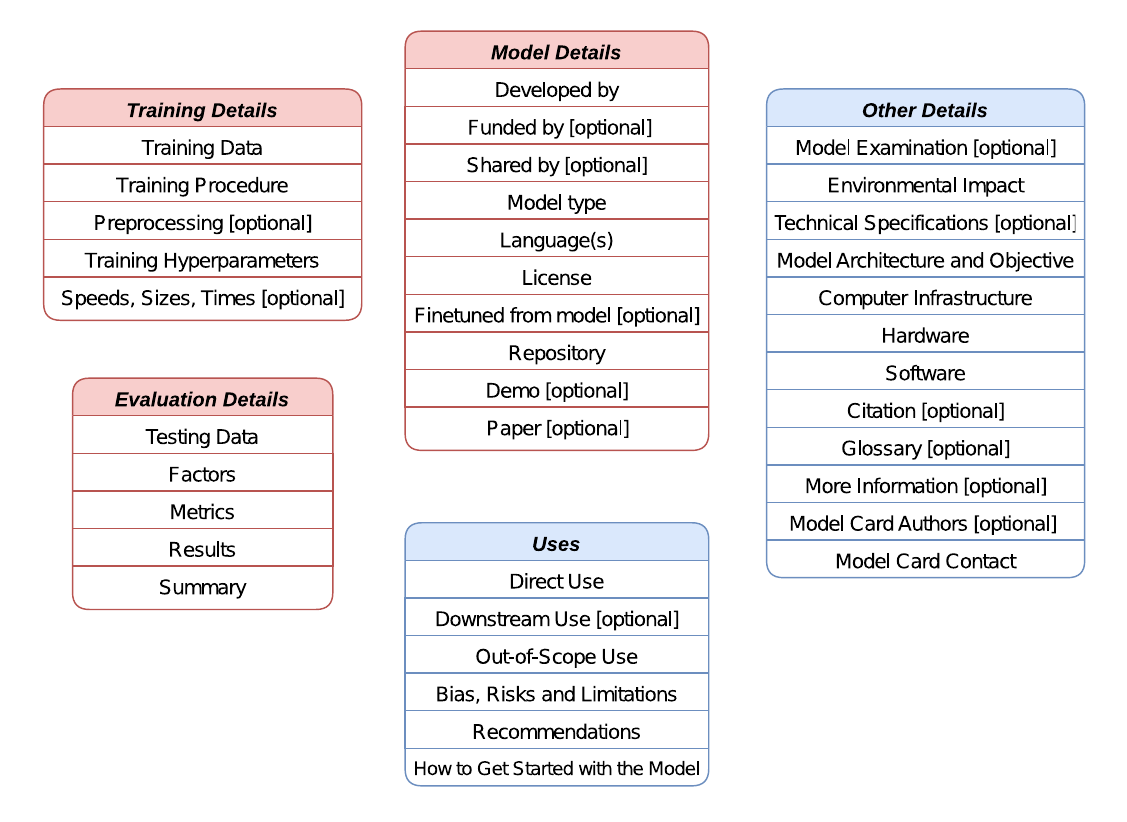}
     \caption{An illustration of a Model Card}
     \label{fig:model_card}
\end{figure*}

We also conduct a manual inspection of $M_{500}$ to identify any discrepancies and incompleteness of information in their model cards (see Figure~\ref{fig:model_card}). Model cards \cite{10.1145/3287560.3287596} are short documents that accompany ML models, furnishing details about the models, it's usage contexts, performance evaluation procedures, and other pertinent information. To assess the comprehensiveness of information within these cards, four annotators manually review each model's model card on HF (refer to \S\ref{sec:ann_info} for annotator details). They annotate each of the 39 elements in a model card across sections, such as ``Training Data'', ``Training Procedure'' and ``Training Hyperparameters'' in the \textbf{Training Details} section, by referring to the HF model card template\footnotemark. Overall, they annotate approximately 78,000 elements (500 models × 39 elements × 4 annotators). Each element is categorized as Full Information Available (FIA), Partial Information Available (PIA), or No Information Available (NIA). For instance, if a model card section contains all the required information according to the template, it is labeled as FIA. If an annotator needs to search for information, such as referring to the research paper URL provided instead of the requested information, or if the information is not provided exactly as per the template or is located in other sections, it is categorized as PIA. If no information is available, it is labeled as NIA. Instances where all annotators disagree result in a ``U'' label, indicating uncertainty.

\footnotetext{\url{https://huggingface.co/docs/hub/model-card-annotated}}
To illustrate the labeling process in detail, we use the model card for \texttt{distilbert/distilgpt2\footnotemark} as an ideal example for comparison, given that it contains nearly all relevant details. For the \textbf{Training Details} section of this model card, we label ``Training Data'' and ``Training Procedure'' as FIA, since comprehensive information is explicitly provided. However, for ``Training Hyperparameters'', we label NIA, as there is no explicit mention of this information in the model card.

\footnotetext{\url{https://huggingface.co/distilbert/distilgpt2}}
In contrast, consider an example from $M_{500}$: \texttt{philschmid/distilbert-base-multilingual-\\cased-sentiment\footnotemark}. \footnotetext{\url{https://huggingface.co/philschmid/distilbert-base-multilingual-cased-sentiment}}
For the \textbf{Training Details} section, an annotator labels ``Training Data'' as PIA, since there is no direct mention of the training data in the ``Training Data'' element of the model card section. However, relevant information about the dataset (\texttt{amazon\_multi\_reviews}\footnotemark) is provided in the model description. For ``Training Procedure'', the label is NIA because no related information is stated, while ``Training Hyperparameters'' is labeled as FIA, as all details regarding hyperparameters are explicitly mentioned in the model card.

\footnotetext{\url{https://huggingface.co/datasets/mteb/amazon_reviews_multi}}
Similarly, annotators systematically label all sections of model cards. The final label for each element is determined by majority vote among annotators. These final labels are then used to calculate the percentages as displayed in Figures~\ref{fig:mi} and~\ref{fig:mc_stack}.

\subsection{Datasets}
\label{sec:data}

We utilize three distinct datasets (\(\mathcal{D}\)): TweetEval\footnote{\url{https://huggingface.co/datasets/cardiffnlp/tweet_eval}} (\(\mathcal{D}_{t}\)) \cite{barbieri-etal-2020-tweeteval}, Amazon Multi Reviews\footnote{\url{https://huggingface.co/datasets/defunct-datasets/amazon_reviews_multi}} (\(\mathcal{D}_{a}\)) \cite{keung-etal-2020-multilingual}, and Financial Phrasebank\footnote{\url{https://huggingface.co/datasets/takala/financial_phrasebank}} (\(\mathcal{D}_{f}\)) \cite{malo2014fp}. Notably, these datasets hold widespread recognition and are commonly employed as benchmark datasets \cite{barbieri-etal-2020-tweeteval, zhang2023sentiment, shah-etal-2022-flue, wang2023fingpt, hussain2021towards, kurihara-etal-2022-jglue}. Furthermore, approximately 72\% of the SA models within $M_{500}$ are originally trained on these datasets, among those in  $M_{500}$ that disclosed their training datasets (around 15.4\%,) which is around ~55 models. While \(\mathcal{D}_{t}\) and \(\mathcal{D}_{f}\) are representative of three-class SA datasets, 
\(\mathcal{D}_{a}\) initially comprises a five-class (0--4) SA classification schema. To bring uniformity in our study, we transform \(\mathcal{D}_{a}\) into a three-class SA dataset by excluding instances with labels 1 and 3. Further details regarding these datasets are provided in Table~\ref{tab:datasets}.

\begin{table*}[!tbh]
\centering
\small

\begin{tabular}{@{}cccccc@{}}
\toprule
\textbf{Dataset}              & \textbf{\begin{tabular}[c]{@{}c@{}}\#Instances\\ (Train/Val/Test)\end{tabular}} & \textbf{\begin{tabular}[c]{@{}c@{}}\#Class\\ Labels\end{tabular}} & \textbf{\begin{tabular}[c]{@{}c@{}}\#Models\\ Trained\end{tabular}} & \textbf{\begin{tabular}[c]{@{}c@{}}\#Downloads\\ All Time\end{tabular}} & \textbf{\begin{tabular}[c]{@{}c@{}}\#Instances\\ (Easy/Ambi/Hard)\end{tabular}} \\ \midrule
\( \bm{\mathcal{D}_{t}} \)           & 45,615/12,284/2,000                                                             & 3                                                                 &620                                                                 & 2,057,849               & 35,793/16,013/8,093                                                             \\
\( \bm{\mathcal{D}_{a}} \) & 120,000/3,000/3,000                                                             & 3                                                                 &152                                                                & 219,774               & 105,052/12,479/8,469                                                            \\
\( \bm{\mathcal{D}_{f}} \) & 4,846/-/-                                                                       & 3                                                                 &130                                                                  & 462,906               & 4,682/163/1                                                                     \\ \bottomrule
\end{tabular}

\caption{Dataset Statistics\protect\footnotemark}
\label{tab:datasets}
\end{table*}

\footnotetext{As of February $6^{\text{th}}$, 2024, sourced from HF datasets webpage. Additionally, $\mathcal{D}_{a}$ is sourced from \url{https://huggingface.co/datasets/mteb/amazon_reviews_multi}, since the official dataset has been taken down. For $\mathcal{D}_{f}$, we split the data in the ratio of 70:10:20 for train, val and test sets repectively.}

\begin{table*}[!htb]
\centering
\small
\begin{tabular}{@{}ccccccccccc@{}}

\toprule
\textbf{}                              &               & \multicolumn{3}{c}{$\bm{\mathcal{D}_{t}}$}    &  & \multicolumn{3}{c}{$\bm{\mathcal{D}_{a}}$}    &  & $\bm{\mathcal{D}_{f}}$ \\ \cmidrule(lr){3-5} \cmidrule(lr){7-9} \cmidrule(l){11-11} 
                                       & \textbf{}     & \textbf{Easy} & \textbf{Ambi} & \textbf{Hard} &  & \textbf{Easy} & \textbf{Ambi} & \textbf{Hard} &  & \textbf{Easy}          \\ \midrule
\multicolumn{1}{c|}{\textbf{}}         & $\bm{r_{d}}$  & 0.270*        & 0.241*        & 0.085         &  & 0.255*        & 0.243*        & 0.241*        &  & 0.196*                 \\
\multicolumn{1}{c|}{\textbf{Evaluate}} & $\bm{r_{l}}$  & 0.164*        & 0.130*        & 0.071         &  & 0.136*        & 0.147*        & 0.156*        &  & 0.187*                 \\
\multicolumn{1}{c|}{\textbf{}}         & $\bm{r_{da}}$ & -0.136*       & -0.126*       & -0.018        &  & -0.087        & -0.083        & -0.125*       &  & -0.098*                \\ \midrule
\multicolumn{1}{c|}{\textbf{}}         & $\bm{r_{d}}$  & -0.089        & -0.145*        & 0.147*         &  & -0.125*       & -0.142*        & 0.206*        &  & -0.069                 \\
\multicolumn{1}{c|}{\textbf{Finetune}} & $\bm{r_{l}}$  & -0.143*       & -0.165*       & 0.161*        &  & -0.177*       & -0.181*       & 0.259*        &  & -0.074                 \\
\multicolumn{1}{c|}{}                  & $\bm{r_{da}}$ & 0.019         & 0.016         & -0.080        &  & 0.001         & -0.008        & -0.054        &  & -0.010                 \\ \bottomrule

\end{tabular}
\caption{Correlation between $r$ and the performance (F1) of models in $M_{500}$ across different dataset categories. Values highlighted with * represent statistical significant values with p-value $< 0.05$. For $\mathcal{D}_{f}$, we did not perform any evaluation strategy for ambiguous and hard instances, since the \#instances categorized as ambiguous and hard is very low, which is insufficient for training and evaluation, therefore values are not calculated and displayed.}
\label{tab:corr_hf}
\end{table*}

\subsection{Training Strategies}
\label{sec:train_strategy}

We explore two distinct strategies for evaluating $M_{500}$ as outlined below:
\paragraph{Evaluate}
In this strategy, we assess the performance of $M_{500}$ models without any prior training on the datasets $\mathcal{D}$.

\paragraph{Finetune}
In this strategy, we finetune $M_{500}$ on the train set and evaluate its performance on the corresponding test set of $\mathcal{D}$.

\subsection{Training and Evaluation Methodology}

We thoroughly evaluate the $M_{500}$ models to assess their performance across different categories of dataset difficulty (Easy, Ambiguous, and Hard) using dataset cartography~\cite{swayamdipta-etal-2020-dataset}. This evaluation helps us understand how these models perform across various datasets based on their difficulty levels, given that models can be used in downstream tasks involving datasets of varying difficulty.

Dataset cartography \cite{swayamdipta-etal-2020-dataset, kadasi-singh-2023-unveiling} framework captures training dynamics during training to categorize instances into easy, ambiguous, and hard instances. It categorizes the instances based on confidence measured as the mean probability of the correct label across epochs, and variability, represented by the variance of the aforementioned confidence. This analysis creates the datamaps as displayed in Figure~\ref{fig:cart_all}, where it has three distinct regions: easy-to-learn (\textbf{Easy}), ambiguous-to-learn (\textbf{Ambiguous}), and hard-to-learn (\textbf{Hard}) for the model.

We train the RoBERTa\footnotemark~\cite{liu2019roberta} model, capturing training dynamics through dataset cartography on $\mathcal{D}$, which comprises train, valid, and test splits (refer to Table~\ref{tab:datasets} for \#instances categorized as easy, ambiguous and hard). Next, we apply Evaluate and Finetune strategies (see \S\ref{sec:train_strategy}) to the test splits of instance difficulty categories.

Before conducting experiments, we perform hyperparameter tuning on the RoBERTa model for $\mathcal{D}$ to identify optimal hyperparameters (refer to \S\ref{apx:hp} for more details). These optimal hyperparameters are then consistently used across all experiments. 

\footnotetext{We also experimented with other models such as XLNet \cite{yang2020xlnet}, BERT \cite{devlin-etal-2019-bert}, and ALBERT \cite{lan2020albert} to categorize instances into different difficulty categories. However, as the average overlap percentage of instances between these models in their respective difficulty categories is higher, we decided to stick with RoBERTa.}

\paragraph{Evaluation Metrics}

To present our results effectively, we calculate F1 scores. Additionally, we calculate Spearman rank correlation~\cite{spearman} to examine the relationship between model performance and their corresponding $r$ for $\mathcal{D}$ across different difficulty categories. We use the following terminology in Table~\ref{tab:corr_range} throughout the rest of the paper.

\begin{table}[th]
\centering
\small
    \begin{tabular}{@{}cc@{}}
    \toprule
    \textbf{Strength} & \textbf{Coefficient Interval} \\ \midrule
    Very Weak         & 0.00 - 0.199                  \\
    Weak              & 0.20 - 0.399                  \\
    Medium            & 0.40 - 0.599                  \\
    Strong            & 0.60 - 0.799                  \\
    Very Strong       & 0.80 - 1.000                  \\ \bottomrule
    \end{tabular}
\caption{Summary of Spearman correlation coefficient strengths}
\label{tab:corr_range}
\end{table}

\section{Results and Discussions}
\label{sec:res}

\subsection{Does High Popularity Necessarily Correlate With Better Model Performance?}
\label{sec:rq1}

With reference to Table~\ref{tab:corr_hf}, for $\mathcal{D}_{t}$, we observe weak to very weak correlations between model performance on easy and ambiguous instances and the metrics $r_{d}$, $r_{l}$, and $r_{da}$. This indicates that for this dataset, models highly ranked in terms of $r_{d}$ and $r_{l}$ and low ranked in terms of $r_{da}$ (older models) are likely to perform better. For hard instances, we do not draw any conclusions since the scores are statistically insignificant. 

This weak correlation is due to models that are highly ranked with respect to $r_{d}$ perform pooorer as compared to some of the models that are low ranked (e.g., \texttt{oliverguhr/german-sentiment-bert\footnote{\url{https://huggingface.co/oliverguhr/german-sentiment-bert}}}, $r_{d}$: 12, f1-rank\footnotemark: 266 on $\mathcal{D}_{t}$ (easy)). This trend could be attributed to ``popularity bias''. Models that are popular (highly ranked) might be favored due to their visibility regardless of their true performance. Community preferences, familiarity with certain model architectures, or advertising on social media by model card authors could play a significant role. Some users may prioritize factors like ease of integration, pretraining details, or compatibility over actual downstream performance.

\footnotetext{Rank assigned by sorting F1-scores of models in the ascending order, 1: high-ranked, 500: least-ranked}

Model downloads may also reflect the utility of a model for specific tasks rather than its overall performance. For instance, a model might perform exceptionally well for a specialized task but is still broadly downloaded because it fits specific user needs (e.g., \texttt{ProsusAI/finbert\footnote{\url{https://huggingface.co/ProsusAI/finbert}}}, $r_{d}$: 5). Conversely, a high-performing model on certain benchmarks may not meet the requirements of a large portion of users, leading to fewer downloads (e.g., \texttt{cardiffnlp/xlm-v-base-tweet-sentiment-ar\footnote{\url{https://huggingface.co/cardiffnlp/xlm-v-base-tweet-sentiment-ar}}}, $r_{d}$: 395).

After applying the fine-tuning strategy, most models that performed poorly in the Evaluate strategy showed performance gains, resulting in a shift from very weak positive to very weak negative correlation with $r_{d}$ and $r_{l}$ across all difficulty categories.

\begin{figure}[tb]
     \centering
     \includegraphics[width=\linewidth]{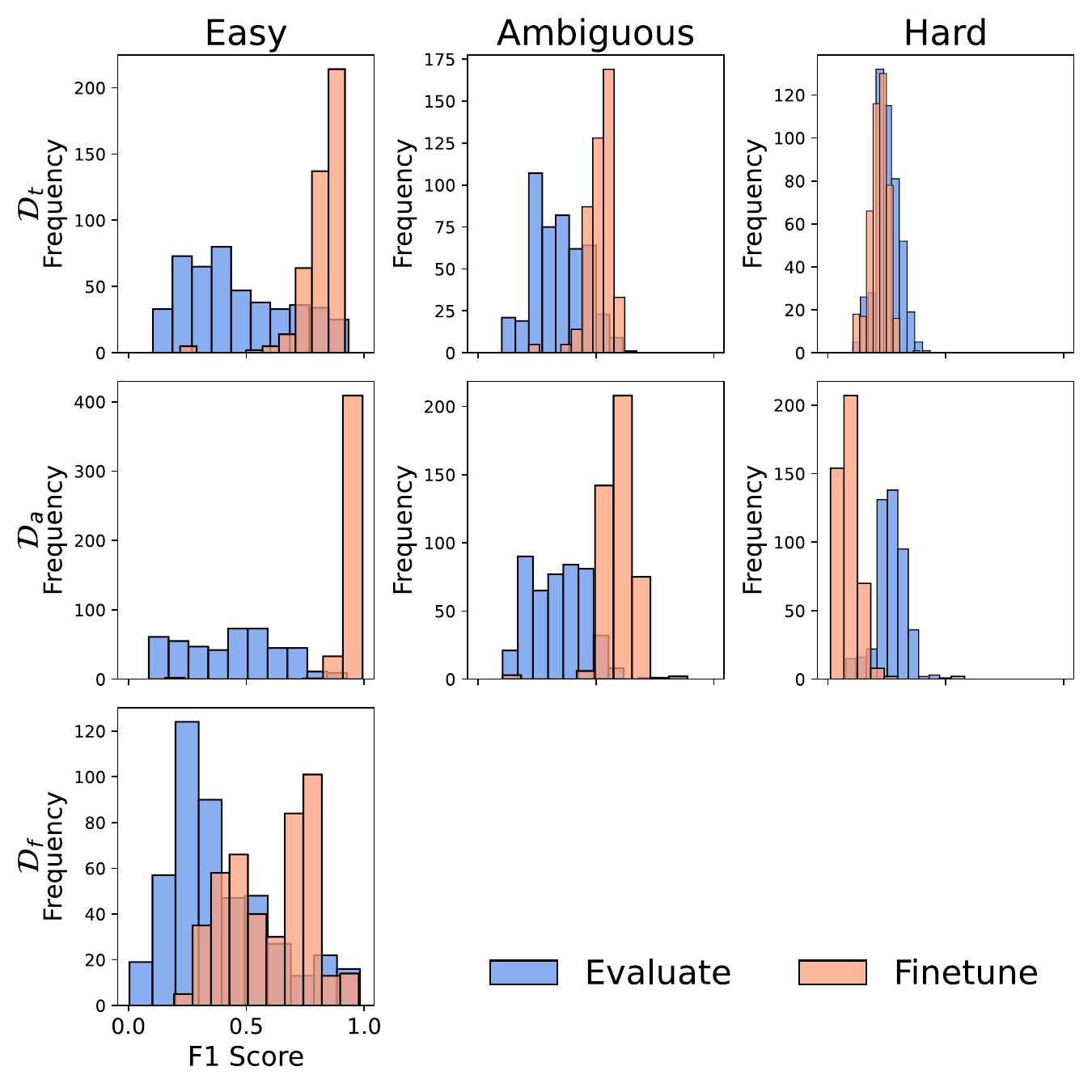}
     \caption{Distribution of F1 scores for models in $M_{500}$ across different difficulty categories in dataset $\mathcal{D}$, under both the Evaluate and Finetune strategies.}
     \label{fig:dist_f1_all}
\end{figure}

For $\mathcal{D}_{a}$, similar to $\mathcal{D}_{t}$, we observe weak to very weak correlations between model performance and the metrics $r$. In the fine-tuning strategy as depicted in Table~\ref{tab:corr_hf}, we observe a very weak negative correlation. Here, fine-tuning models has degraded the correlation further. For $\mathcal{D}_{f}$, we observe a very weak correlation with $r$ compared to $\mathcal{D}_{t}$ and $\mathcal{D}_{a}$.

To further investigate the reasons behind these correlation changes, we analyze the distribution of F1 scores, as shown in the distribution plot (see Figure~\ref{fig:dist_f1_all}). Following Finetuning, a significant proportion of models exhibit improved performance, particularly for easy and ambiguous instances. Interestingly, for hard instances, there is minimal difference in the distribution of F1 scores between the Evaluate and Finetune strategies. Specifically for dataset $\mathcal{D}_{a}$, performance declined after Finetuning, indicating that Finetuning on hard instances did not necessarily enhance model performance. Despite being perceived as top-performing models as evident from their popularity, these high-ranked models within $M_{500}$ did not consistently outperform other low-ranked (or less popular) models when handling instances of varying difficulty or from different domains.

\begin{tcolorbox}
    \textbf{Takeaway:} 
    Contrary to common belief, high popularity does not always guarantee superior model performance.
\end{tcolorbox}

\subsection{Manual Inspection of \bm{$M_{500}$}}

\begin{table*}[!htbp]
\centering
\small
\begin{tabular}{@{}cccclccclclccc@{}}

\toprule
\textbf{}                         & \multicolumn{3}{c}{$\bm{\mathcal{D}_{t}}$}    &  & \multicolumn{3}{c}{$\bm{\mathcal{D}_{a}}$}    &  & $\bm{\mathcal{D}_{f}}$ &  & \multicolumn{3}{c}{\textbf{r}}              \\ \cmidrule(lr){2-4} \cmidrule(lr){6-8} \cmidrule(lr){10-10} \cmidrule(l){12-14} 
                                  & \textbf{Easy} & \textbf{Ambi} & \textbf{Hard} &  & \textbf{Easy} & \textbf{Ambi} & \textbf{Hard} &  & \textbf{Easy}          &  & $\bm{r_{d}}$ & $\bm{r_{l}}$ & $\bm{r_{da}}$ \\ \midrule
\multicolumn{1}{c|}{\textbf{FIA}} & 0.231*        & 0.221*        & 0.023         &  & 0.192*        & 0.196*        & 0.169*        &  & 0.151*                 &  & 0.619*       & 0.527*       & -0.081       \\
\multicolumn{1}{c|}{\textbf{PIA}} & 0.112*         & 0.085         & -0.071        &  & 0.067         & 0.076         & 0.081         &  & 0.066                  &  & 0.508*       & 0.458*       & -0.158*       \\
\multicolumn{1}{c|}{\textbf{NIA}} & -0.180*       & -0.168*       & -0.027        &  & -0.155*       & -0.166*       & -0.145*       &  & -0.165*                &  & -0.591*      & -0.491*      & 0.067        \\ \bottomrule
\end{tabular}

\caption{Correlation values computed between the percentage of FIA/PIA/NIA (across all sections of the model card) and model performance (F1). Additionally, it includes correlation values calculated between the percentage of FIA/PIA/NIA and $r$ (model popularity). Values highlighted with * represent statistical significant values with p-value $<0.05$. For $\mathcal{D}_{f}$, we did not perform any evaluation strategy for ambiguous and hard instances, since the \#instances categorized as ambiguous and hard is very low, which is insufficient for training and evaluation, therefore values are not calculated and displayed.}
\label{tab:mc_perf_rank_corr}
\end{table*}

\begin{figure}[!htb]
     \centering
     \includegraphics[width=\linewidth]{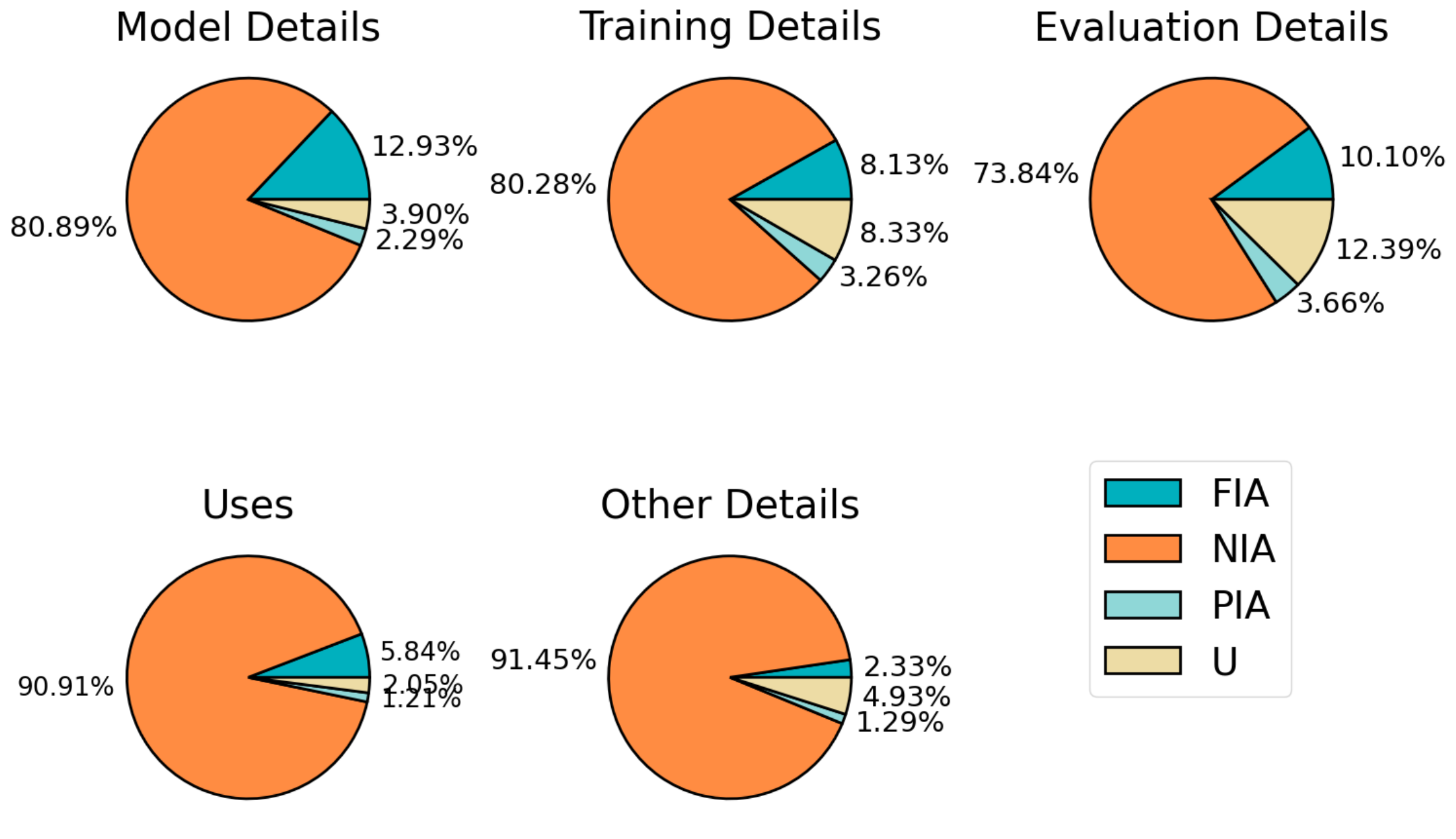}
     \caption{Results of manual inspection of model cards as described in \S\ref{sec:resource}. Each pie chart corresponds to a specific section within the model card.}
     \label{fig:mi}
\end{figure}
\begin{figure}[!htb]
     \centering
     \includegraphics[width=0.9\linewidth]{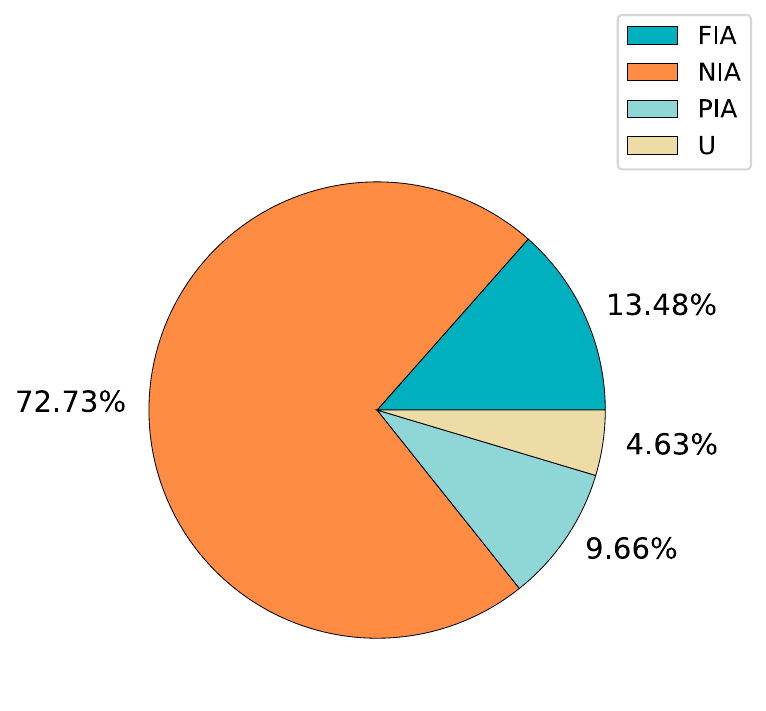}
     \caption{Results of manual inspection of class labels for the $M_{500}$ models.}
     \label{fig:label_pie}
\end{figure}

As detailed in \S\ref{sec:resource}, post-inspection of $M_{500}$, depicted in Figure~\ref{fig:mi}, a substantial lack of information emerges. Approximately 81\% of models lack all the details in the \textbf{Model Details} section of the model card, while 80\% do not provide insights into specific training processes and corresponding details. Additionally, about 74\% did not provide any information about model evaluation (\textbf{Evaluation Details}). Notably, 91\% of models lack complete information concerning \textbf{Uses}  and \textbf{Other Details}.

\begin{figure*}[htbp]
     \centering
     \includegraphics[width=0.8\textwidth]{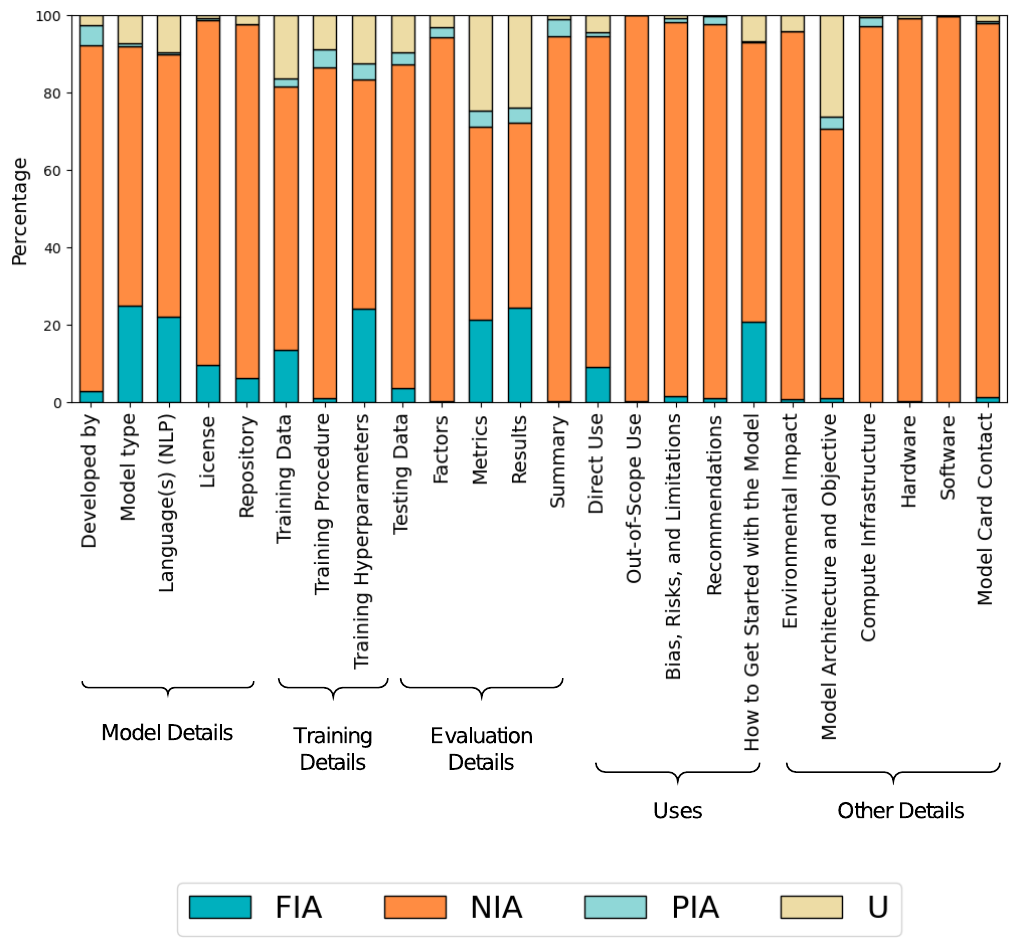}
     \caption{Results of manual inspection of model cards (granular analysis), as detailed in \S\ref{sec:resource}. Each bar corresponds to a specific element within a section of the model card.}
     \label{fig:mc_stack}
\end{figure*}


A more granular analysis of Model Card sections concerning Figure~\ref{fig:mc_stack}, focusing on mandatory (non-optional) elements per section (see Figure~\ref{fig:model_card}), reveals significant gaps. For \textbf{Training Details}, nearly 68\% of models provide no information about the training data, approximately 85\% omit details about the training process, and about 59\% fail to disclose hyperparameter information. Similar patterns are observed in \textbf{Evaluation} and \textbf{Model Details}. Within the \textbf{Uses} section, essential details such as \textbf{Direct Use} (85\%), \textbf{Out-of-Scope Use} (99\%), \textbf{Bias Risks, and Limitations} (96\%), and \textbf{Recommendations} (96\%) are predominantly absent. The absence of these critical details raises questions about the models' suitability for downstream tasks.

During the inspection, we observe the absence of label mapping, particularly for categorical integer labels. This absence hinders the interpretation of which categorical label corresponds to which sentiments (Positive, Negative, or Neutral). In cases where no label information is present, we perform queries using HF Inference API\footnote{\url{https://huggingface.co/docs/api-inference/index}} on their model page, categorizing these models under PIA. In instances where label outputs are ambiguous, such as \texttt{Label\_0}, \texttt{Label\_1}, and \texttt{Label\_2}, it becomes challenging to determine the corresponding sentiment. These issues are categorized as NIA and are observed in 23\% of models, as illustrated in Figure~\ref{fig:label_pie}.

\subsubsection{Does the Performance of a Model Correlate With How Well Documented Its Model Cards Are?}
\label{sec:rq2}

\begin{figure*}[!htb]
     \centering
     \includegraphics[width=\textwidth]{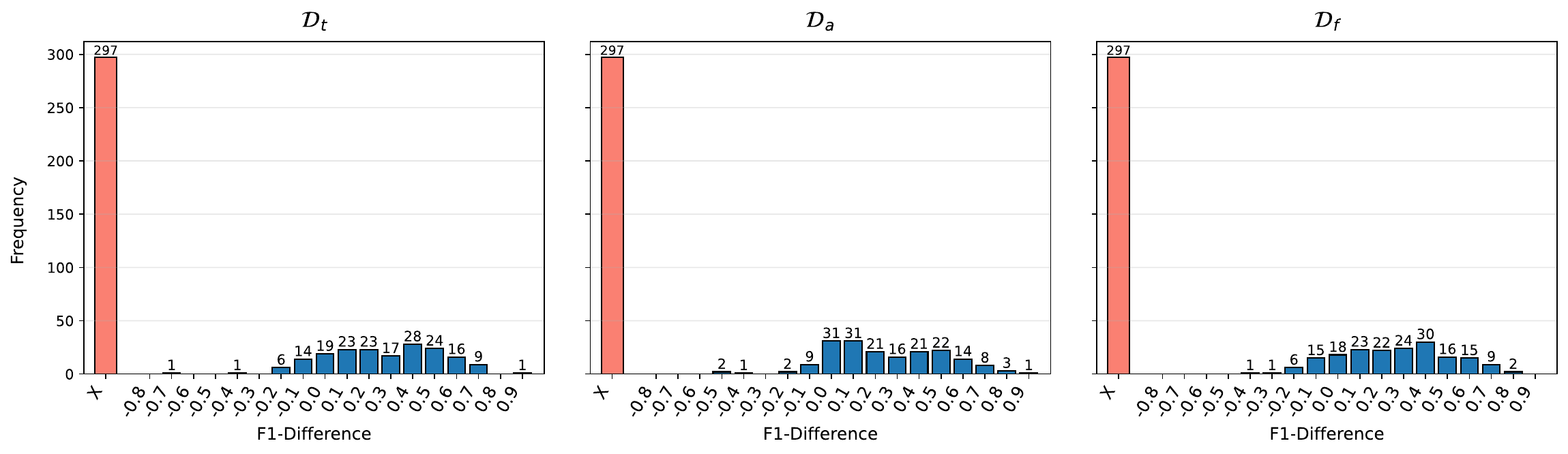}
     \caption{Difference between reported performance (F1) and empirical performance (F1) scores across datasets in $\mathcal{D}$, with F1-Difference on the X-axis and Frequency on y-axis. Empirical performance refers to the performance of $M_{500}$ models evaluated on the full dataset of $\mathcal{D}$, while reported performance indicates the performance reported by the authors in their model cards. Cross (X) mark on the X-axis, with red colored bar represent the \#models whose scores are not reported. For the mapping of model IDs to model names, refer Table~\ref{tab:m100_models} to Table~\ref{tab:m500_models} in the \S\ref{sec:appendix}.}
     \label{fig:mc_perf_rep_emp}
\end{figure*}

We conducted a qualitative study of the $M_{500}$ model cards through manual inspection as detailed in \S\ref{sec:mcards}. Our investigation reveals a weak positive correlation between the comprehensiveness of the information provided in the model cards and the performance of the respective models, as evident from Table~\ref{tab:mc_perf_rank_corr}. 
This suggests that as the amount of full information available in the model cards (FIA) increases, there is a slight tendency for models to perform better on easy and ambiguous instances as compared to hard instances. However, the correlation is weak, so this relationship is not strong or highly predictive. This trend remains consistent across all categories of instances (easy, ambiguous, and hard) in dataset $\mathcal{D}$, with one exception: for hard instances of $\mathcal{D}_t$, the correlation score is statistically insignificant.

Interestingly, while some well-documented models perform well, others with similarly comprehensive documentation might not. Conversely, less-documented models occasionally outperform their well-documented counterparts, further weakening the overall correlation. This variability highlights that documentation alone is not a reliable indicator of model performance.

Overall, our findings indicate that models with more available information (FIA) did not consistently perform better in downstream tasks. Conversely, models with insufficient information (NIA) show a very weak negative correlation, suggesting that while these models tend to perform slightly worse than those with more information, the weak correlation implies that this is not always the case. This trend is consistent across all categories in dataset $\mathcal{D}$, except for hard instances of $\mathcal{D}_t$ for which the correlation score is statistically insignificant. We did not draw any conclusions from PIA scores due to their statistical insignificance.

\begin{tcolorbox}
    \textbf{Takeaway:} 
    Comprehensive documentation does not necessarily gurantee better model performance.
\end{tcolorbox}

\subsubsection{Are Popular Models Well Documented?}
\label{sec:rq3}

We continue our analysis by examining the correlation between the amount of information provided in model cards and $r$. Our analysis reveals a notable correlation between the comprehensiveness of information in model cards and their model rankings, as demonstrated in Table~\ref{tab:mc_perf_rank_corr}. Specifically, models with more information available (FIA/PIA) in their model cards tend to receive higher rankings in terms of $r$. Conversely, for models with insufficient information (NIA), the correlation is negative, indicating lower rankings (as $r_{d}$ increases [eg., from $r_{d}$:1 to $r_{d}$:5], completeness rank ($r_{c}$)\footnotemark decreases (eg., from $r_{c}$:5 to $r_{c}$:1)). 
This suggests as amount of information in model card decreases, there is a tendency for models to be less popular. These trends are consistent across all datasets. These observations underscore the importance of providing comprehensive information in model cards, as it appears to be correlated with the perceived trustworthiness and popularity of models for downstream tasks. It is important to note that this correlation does not imply causation. The relationship could be bidirectional, where more comprehensive documentation might enhance model popularity, and higher popularity could lead maintainers to invest more time in documenting the model. Further research is needed to explore this potential bidirectional relationship.

\footnotetext{Completeness rank ($r_{c}$) is calculated by sorting percentage of FIA/PIA/NIA in descending order and assigning the ranks in ascending order, from 1 to 500}

\begin{tcolorbox}
    \textbf{Takeaway:} 
    Models accompanied by more comprehensive details in their model cards tended to have higher popularity among users and vice-versa.
\end{tcolorbox}

\subsubsection{Does the Reported Performance Accurately Reflect Its True Capabilities?}
\label{sec:rq4}

We further investigate whether the performance scores reported in the model cards of $M_{500}$ accurately reflect the models' true performance. 

As almost all of the models have not specified details about their evaluation datasets, we are unable to assess their performance directly on those datasets. However, as a proxy, we evaluate them on similar domains. For example, Model X is evaluated on Dataset Y (unknown), and we record its performance (as mentioned in the model card). We then compare this by evaluating Model X on a benchmark dataset from the same domain as Dataset Y.

To explore this further, we examine the performance of $M_{500}$ on the complete dataset (combining train, validation, and test sets) of $\mathcal{D}$. Initially, we assess what percentage of models reported their scores on HF model cards. Our findings reveal that approximately 60\% of models failed to report their scores\footnote{We include models that have reported the macro F1 score for any given dataset and exclude the rest, which amounts to 10\% of models. Additionally, we exclude approximately 4\% of models from our analysis due to errors during model evluations.}. Among those who did report their scores, we assess the validity of the reported scores. To accomplish this, we conduct evaluations of $M_{500}$ models on the complete datasets of $\mathcal{D}$. Following the evaluations, we observe that a notable proportion ($\approx$ 88\%) of models performed worse on $\mathcal{D}$ compared to their original performance on some other dataset (it may or may not be $\mathcal{D}$) as depicted in Figure~\ref{fig:mc_perf_rep_emp}. These findings suggest that many model authors do not thoroughly evaluate their models across diverse dataset domains. Even if evaluations are conducted, we recommend that model authors accurately document the downstream usage of their models, their potential performance across various dataset domains, and any associated limitations.

\begin{tcolorbox}
    \textbf{Takeaway:} 
    A staggering 88\% of model authors overstated scores higher than what is reported in the model cards. 
\end{tcolorbox}

\subsection{A Case Study on the Reddit Dataset}
\label{sec:case_study}

In the preceding sections, we examined the performance of $M_{500}$ on popular benchmark datasets. Here, we further assess $M_{500}$'s capabilities by evaluating its performance on a new, unseen dataset. Specifically, we conduct a case study using a real-world dataset sourced from social media posts on Reddit\footnotemark.

\subsubsection{Creation of Reddit Dataset}
\label{sec:reddit_data}

We utilized the PRAW\footnote{\url{https://praw.readthedocs.io/en/stable/}} library to scrape comments from the top 100 posts (of all time) of ten subreddits: Afghanistan (r/afghanistan), China (r/china), Bhutan (r/bhutan), Bangladesh (r/bangladesh), India (r/india), Sri Lanka (r/srilanka), Pakistan (r/pakistan), Maldives (r/maldives), Myanmar (r/myanmar), and Nepal (r/nepal). These subreddits primarily feature political content and this process gathered a total of 16,000 comments, including nested ones. After merging the comments, those containing URLs were excluded from subsequent sampling. Our focus narrowed to comments with a token count (excluding stopwords) ranging from 5 to 50 to maintain the quality of the text. We employed three SA models--\texttt{\url{cardiffnlp/twitter-roberta-base-sentiment-latest}}\footnote{\url{https://huggingface.co/cardiffnlp/twitter-roberta-base-sentiment-latest}}, \texttt{\url{finiteautomata/bertweet-base-sentiment-analysis}}\footnote{\url{https://huggingface.co/finiteautomata/bertweet-base-sentiment-analysis}}, and \texttt{\url{Seethal/sentiment_analysis_generic_dataset}}\footnote{\url{https://huggingface.co/Seethal/sentiment_analysis_generic_dataset}} chosen for their popularity\footnote{As of August $20^{\text{th}}$, 2023} to determine the majority sentiment for each comment. To ensure balanced representation, we randomly sampled 1,000 comments, aiming for an equal distribution of positive, negative, and neutral majority sentiments. The average sentence length for this dataset is 21 (words).

We then recruit four undergraduate students proficient in English and familiar with SA tasks to annotate the 1,000 instances. Before annotation, all annotators receive detailed instructions and examples of pre-annotated instances. Inter-annotator agreement is assessed using Krippendorff's $\alpha$, yielding a score of 0.46. 

\begin{table}[th]
\centering
\small
    \begin{tabular}{@{}cccc@{}}
    \toprule
    Dataset & Avg. & Min & Max \\ \midrule
    Reddit  & 20.8 & 6   & 50  \\
    $\mathcal{D}_{t}$      & 22.3 & 1   & 88  \\
    $\mathcal{D}_{a}$      & 42   & 3   & 808 \\
    $\mathcal{D}_{f}$      & 23   & 2   & 81  \\ \bottomrule
    \end{tabular}
\caption{The table displays average, min. and max. sentence length (\#words) in the datasets studied, an insight into the data distribution.}
\label{tab:data_dist}
\end{table}

\subsubsection{Evaluation and Analysis}
\label{sec:reddit_data_eval}

Next, we evaluate the performance of the $M_{500}$ on the 1,000-instance Reddit dataset and calculate the correlation between $r$ and the $M_{500}$ performance scores. We observe a very weak correlation between $r$ and $M_{500}$ performance, with correlation scores of 0.266, 0.136, and -0.1 for $r_{d}$, $r_{l}$, and $r_{da}$ respectively, all of which are statistically significant (p $< 0.05$). We also report the mean and standard deviation of the reported F1 scores: (0.74, 0.17), compared to the mean and standard deviation of empirical F1 scores on the Reddit dataset: (0.41, 0.16) which indeed indicate poor generalization. 

The weak generalization observed may be attributed to a domain mismatch, as the Reddit dataset likely differs substantially in style, tone, and structure from the datasets on which the models were originally trained or fine-tuned. Training datasets are often curated from formal or task-specific texts, whereas Reddit's informal, conversational, and heterogeneous content poses distinct challenges that these models may not be equipped to address effectively. Furthermore, a potential popularity bias could influence model rankings, with higher-ranked models being favored for factors unrelated to their actual performance, such as ease of integration, familiarity within the community, or promotional efforts by model card authors. This bias may contribute to an inflated perception of their generalization capabilities.

Furthermore, the nature of Reddit comments, often characterized by informal language, slang, sarcasm, and context-dependent sentiment adds an additional layer of complexity. These features may be harder for models to interpret accurately, especially if they have not been exposed to similar linguistic characteristics during training.

Overall, these findings indicate that, although considered the top performing models due to their popularity, the $M_{500}$ models demonstrate weak correlations and poor generalization on this dataset. This highlights the importance of finetuning on diverse and representative datasets to improve generalization across varying domains and dataset difficulty types.

\begin{tcolorbox}
    \textbf{Takeaway:} 
    Although perceived as top models, $M_{500}$ displays poor generalization performance.
\end{tcolorbox}

\footnotetext{\url{https://www.reddit.com/}}

\subsection{Key Recommendations}
\label{sec:reco}
This section provides empirically-backed recommendations for two key user groups: those downloading pre-trained models for downstream tasks and those contributing models to public repositories. While some of these recommendations might be intuitively known, our study emphasizes their importance through empirical evidence.

\noindent \textbf{For users downloading models:}
\begin{itemize}
    \item \textit{Don't rely solely on popularity:} Avoid selecting models based solely on their high number of downloads or likes. Popularity does not always correlate with performance quality. For instance, \texttt{oliverguhr/german-sentiment-bert\footnotemark} ($r_{d}$: 12, F1-rank: 266 on $\mathcal{D}_{t}$ (easy)) demonstrates poor generalization despite its high popularity (refer to \S\ref{sec:rq1}).
    \footnotetext{\url{https://huggingface.co/oliverguhr/german-sentiment-bert}}

    \item \textit{Check for comprehensive documentation:} While detailed and well-structured documentation can aid in understanding a model's usage, do not assume that well-documented models always perform better. For example, \texttt{HerbertAIHug/Finetuned-Roberta-Base-\\Sentiment-identifier\footnotemark}\footnotetext{\url{https://huggingface.co/HerbertAIHug/Finetuned-Roberta-Base-Sentiment-identifier}} has detailed documentation but does not consistently outperform less-documented models (refer to \S\ref{sec:rq2}). Ideal documentation examples include \texttt{distilbert/distilgpt2\footnotemark}\footnotetext{\url{https://huggingface.co/distilbert/distilgpt2}}.

    \item \textit{Evaluate model suitability:} Consider the model's documented performance on relevant benchmarks or similar tasks to ensure its suitability for a specific use case (refer to \S\ref{sec:rq4}). For instance, \texttt{cross-encoder/ms-marco-MiniLM-L-6-v2\footnotemark}\footnotetext{\url{https://huggingface.co/cross-encoder/ms-marco-MiniLM-L-6-v2}} and \texttt{sismetanin/sbert-ru-sentiment-krnd\footnotemark}\footnotetext{\url{https://huggingface.co/sismetanin/sbert-ru-sentiment-krnd}} demonstrate strong task-specific performance.
\end{itemize}

\noindent \textbf{For users contributing models to public repositories:}
\begin{itemize}
    \item \textit{Emphasize thorough documentation:} Comprehensive documentation is critical to making a model popular, trusted, and widely accepted on platforms like HF. Include details on training data, procedures, hyperparameters, and any limitations or biases (refer to \S\ref{sec:rq3}). Examples of well-documented models include \texttt{cardiffnlp/twitter-roberta-base-\\sentiment-latest\footnotemark}.
    \footnotetext{\url{https://huggingface.co/cardiffnlp/twitter-roberta-base-sentiment-latest}}

    \item \textit{Highlight performance metrics:} Provide detailed performance metrics and benchmarks to help users understand the model's strengths and weaknesses. Transparency in performance builds credibility and trust (refer to \S\ref{sec:rq4}). For example, \texttt{oliverguhr/german-sentiment-bert\footnotemark}\footnotetext{\url{https://huggingface.co/oliverguhr/german-sentiment-bert}} and \texttt{sismetanin/sbert-ru-sentiment-krnd\footnotemark}\footnotetext{\url{https://huggingface.co/sismetanin/sbert-ru-sentiment-krnd}} explicitly document their performance metrics.

    \item \textit{Engage with the community:} Actively respond to user feedback and queries. Community engagement improves the model's reputation and can highlight areas for improvement. For example, \texttt{google-bert/bert-base-uncased\footnotemark} maintains an active discussion platform\footnotetext{\url{https://huggingface.co/google-bert/bert-base-uncased/discussions}}.
\end{itemize}

\section{Conclusions}

In our study, we evaluated popular SA models on HF, exploring their performance across various datasets and difficulty categories to identify any correlation between popularity and performance. Surprisingly, we found no strong connection between popularity and overall model performance, with models lacking detailed information in their cards being less popular and performing worse. Our findings highlight the importance of thorough documentation and rigorous evaluation processes in NLP, emphasizing that popularity metrics alone are insufficient for assessing a model's suitability. Instead, we must develop more robust metrics that prioritize transparency, reproducibility, and comprehensive evaluation to drive advancements in NLP research and ensure the reliability of models in real-world applications.

\section*{Limitations}
The current study is constrained to sentiment analysis tasks, limiting the generalizability of the findings to other domains within NLP. Similarly, the scope of the study is confined to three datasets, which may restrict the broader applicability of the results and observations. Additionally, resource limitations pose a significant bottleneck, preventing experiments on a very large number of models, large-sized (like Llama models) with respect to the \#parameters, or across extensive text corpora.

\section*{Ethics Statement}

Before participating in the evaluation, all human participants were provided with clear and comprehensive information about the nature and objectives of the study. Explicit informed consent was obtained from each participant before their involvement in the research.

We also recognize that subjective judgments, particularly in assessing model card completeness, may introduce potential biases in our evaluation. While we followed predefined guidelines and incorporated multiple annotators to improve consistency, some level of subjectivity remains. We consider fairness and transparency in our methodology a priority and encourage future work to explore more standardized evaluation frameworks.

\section*{Acknowledgements}
We express our sincere gratitude to the University Grants Commission (UGC), Ministry of Education, Government of India, for funding the lead author's PhD through the UGC-JRF program, which made this work possible. We also thank the anonymous reviewers for their valuable feedback, which has greatly contributed to improving this paper.

\bibliography{aaai25, anthology, custom}

\begin{thebibliography}{33}
\providecommand{\natexlab}[1]{#1}

\bibitem[{Barabási and Albert(1999)}]{barabasi}
Barabási, A.-L.; and Albert, R. 1999.
\newblock Emergence of Scaling in Random Networks.
\newblock \emph{Science}, 286(5439): 509--512.

\bibitem[{Barbieri et~al.(2020)Barbieri, Camacho-Collados, Espinosa~Anke, and Neves}]{barbieri-etal-2020-tweeteval}
Barbieri, F.; Camacho-Collados, J.; Espinosa~Anke, L.; and Neves, L. 2020.
\newblock {T}weet{E}val: Unified Benchmark and Comparative Evaluation for Tweet Classification.
\newblock In Cohn, T.; He, Y.; and Liu, Y., eds., \emph{Findings of the Association for Computational Linguistics: EMNLP 2020}, 1644--1650. Online: Association for Computational Linguistics.

\bibitem[{Casta{\~n}o et~al.(2023)Casta{\~n}o, Mart'inez-Fern'andez, Franch, and Bogner}]{Castao2023AnalyzingTE}
Casta{\~n}o, J.; Mart'inez-Fern'andez, S.; Franch, X.; and Bogner, J. 2023.
\newblock Analyzing the Evolution and Maintenance of ML Models on Hugging Face.
\newblock \emph{ArXiv}, abs/2311.13380.

\bibitem[{Castaño et~al.(2023)Castaño, Martínez-Fernández, Franch, and Bogner}]{10304801}
Castaño, J.; Martínez-Fernández, S.; Franch, X.; and Bogner, J. 2023.
\newblock Exploring the Carbon Footprint of Hugging Face's ML Models: A Repository Mining Study.
\newblock In \emph{2023 ACM/IEEE International Symposium on Empirical Software Engineering and Measurement (ESEM)}, 1--12.

\bibitem[{Crisan et~al.(2022)Crisan, Drouhard, Vig, and Rajani}]{10.1145/3531146.3533108}
Crisan, A.; Drouhard, M.; Vig, J.; and Rajani, N. 2022.
\newblock Interactive Model Cards: A Human-Centered Approach to Model Documentation.
\newblock In \emph{Proceedings of the 2022 ACM Conference on Fairness, Accountability, and Transparency}, FAccT '22, 427–439. New York, NY, USA: Association for Computing Machinery.
\newblock ISBN 9781450393522.

\bibitem[{Devlin et~al.(2019)Devlin, Chang, Lee, and Toutanova}]{devlin-etal-2019-bert}
Devlin, J.; Chang, M.-W.; Lee, K.; and Toutanova, K. 2019.
\newblock {BERT}: Pre-training of Deep Bidirectional Transformers for Language Understanding.
\newblock In Burstein, J.; Doran, C.; and Solorio, T., eds., \emph{Proceedings of the 2019 Conference of the North {A}merican Chapter of the Association for Computational Linguistics: Human Language Technologies, Volume 1 (Long and Short Papers)}, 4171--4186. Minneapolis, Minnesota: Association for Computational Linguistics.

\bibitem[{{FORCE11}(2020)}]{fair}
{FORCE11}. 2020.
\newblock The FAIR Data principles.
\newblock \url{https://force11.org/info/the-fair-data-principles/}.
\newblock Accessed: 2025-01-15.

\bibitem[{Gebru et~al.(2021)Gebru, Morgenstern, Vecchione, Vaughan, Wallach, Iii, and Crawford}]{gebru2021datasheets}
Gebru, T.; Morgenstern, J.; Vecchione, B.; Vaughan, J.~W.; Wallach, H.; Iii, H.~D.; and Crawford, K. 2021.
\newblock Datasheets for datasets.
\newblock \emph{Communications of the ACM}, 64(12): 86--92.

\bibitem[{Hussain et~al.(2021)Hussain, Holla, Mishra, Yannakoudakis, and Shutova}]{hussain2021towards}
Hussain, A.; Holla, N.; Mishra, P.; Yannakoudakis, H.; and Shutova, E. 2021.
\newblock Towards a robust experimental framework and benchmark for lifelong language learning.
\newblock In \emph{Thirty-fifth Conference on Neural Information Processing Systems Datasets and Benchmarks Track (Round 1)}.

\bibitem[{Jiang et~al.(2023)Jiang, Synovic, Hyatt, Schorlemmer, Sethi, Lu, Thiruvathukal, and Davis}]{10.1109/ICSE48619.2023.00206}
Jiang, W.; Synovic, N.; Hyatt, M.; Schorlemmer, T.~R.; Sethi, R.; Lu, Y.-H.; Thiruvathukal, G.~K.; and Davis, J.~C. 2023.
\newblock An Empirical Study of Pre-Trained Model Reuse in the Hugging Face Deep Learning Model Registry.
\newblock In \emph{Proceedings of the 45th International Conference on Software Engineering}, ICSE '23, 2463–2475. IEEE Press.
\newblock ISBN 9781665457019.

\bibitem[{Kadasi and Singh(2023)}]{kadasi-singh-2023-unveiling}
Kadasi, P.; and Singh, M. 2023.
\newblock Unveiling the Multi-Annotation Process: Examining the Influence of Annotation Quantity and Instance Difficulty on Model Performance.
\newblock In Bouamor, H.; Pino, J.; and Bali, K., eds., \emph{Findings of the Association for Computational Linguistics: EMNLP 2023}, 1371--1388. Singapore: Association for Computational Linguistics.

\bibitem[{Kathikar et~al.(2023)Kathikar, Nair, Lazarine, Sachdeva, and Samtani}]{10297271}
Kathikar, A.; Nair, A.; Lazarine, B.; Sachdeva, A.; and Samtani, S. 2023.
\newblock Assessing the Vulnerabilities of the Open-Source Artificial Intelligence (AI) Landscape: A Large-Scale Analysis of the Hugging Face Platform.
\newblock In \emph{2023 IEEE International Conference on Intelligence and Security Informatics (ISI)}, 1--6.

\bibitem[{Keung et~al.(2020)Keung, Lu, Szarvas, and Smith}]{keung-etal-2020-multilingual}
Keung, P.; Lu, Y.; Szarvas, G.; and Smith, N.~A. 2020.
\newblock The Multilingual {A}mazon Reviews Corpus.
\newblock In Webber, B.; Cohn, T.; He, Y.; and Liu, Y., eds., \emph{Proceedings of the 2020 Conference on Empirical Methods in Natural Language Processing (EMNLP)}, 4563--4568. Online: Association for Computational Linguistics.

\bibitem[{Khan et~al.(2022)Khan, Naseer, Hayat, Zamir, Khan, and Shah}]{10.1145/3505244}
Khan, S.; Naseer, M.; Hayat, M.; Zamir, S.~W.; Khan, F.~S.; and Shah, M. 2022.
\newblock Transformers in Vision: A Survey.
\newblock \emph{ACM Comput. Surv.}, 54(10s).

\bibitem[{Kurihara, Kawahara, and Shibata(2022)}]{kurihara-etal-2022-jglue}
Kurihara, K.; Kawahara, D.; and Shibata, T. 2022.
\newblock {JGLUE}: {J}apanese General Language Understanding Evaluation.
\newblock In Calzolari, N.; B{\'e}chet, F.; Blache, P.; Choukri, K.; Cieri, C.; Declerck, T.; Goggi, S.; Isahara, H.; Maegaard, B.; Mariani, J.; Mazo, H.; Odijk, J.; and Piperidis, S., eds., \emph{Proceedings of the Thirteenth Language Resources and Evaluation Conference}, 2957--2966. Marseille, France: European Language Resources Association.

\bibitem[{Lan et~al.(2020)Lan, Chen, Goodman, Gimpel, Sharma, and Soricut}]{lan2020albert}
Lan, Z.; Chen, M.; Goodman, S.; Gimpel, K.; Sharma, P.; and Soricut, R. 2020.
\newblock ALBERT: A Lite BERT for Self-supervised Learning of Language Representations.
\newblock arXiv:1909.11942.

\bibitem[{Liu et~al.(2024)Liu, Li, Jin, and Diab}]{liu2024automatic}
Liu, J.; Li, W.; Jin, Z.; and Diab, M. 2024.
\newblock Automatic Generation of Model and Data Cards: A Step Towards Responsible AI.
\newblock arXiv:2405.06258.

\bibitem[{Liu et~al.(2019)Liu, Ott, Goyal, Du, Joshi, Chen, Levy, Lewis, Zettlemoyer, and Stoyanov}]{liu2019roberta}
Liu, Y.; Ott, M.; Goyal, N.; Du, J.; Joshi, M.; Chen, D.; Levy, O.; Lewis, M.; Zettlemoyer, L.; and Stoyanov, V. 2019.
\newblock RoBERTa: A Robustly Optimized BERT Pretraining Approach.
\newblock arXiv:1907.11692.

\bibitem[{Loshchilov and Hutter(2019)}]{loshchilov2018decoupled}
Loshchilov, I.; and Hutter, F. 2019.
\newblock Decoupled Weight Decay Regularization.
\newblock In \emph{International Conference on Learning Representations}.

\bibitem[{Malo et~al.(2014)Malo, Sinha, Korhonen, Wallenius, and Takala}]{malo2014fp}
Malo, P.; Sinha, A.; Korhonen, P.; Wallenius, J.; and Takala, P. 2014.
\newblock Good debt or bad debt: Detecting semantic orientations in economic texts.
\newblock \emph{Journal of the Association for Information Science and Technology}, 65(4): 782--796.

\bibitem[{Mitchell et~al.(2019)Mitchell, Wu, Zaldivar, Barnes, Vasserman, Hutchinson, Spitzer, Raji, and Gebru}]{10.1145/3287560.3287596}
Mitchell, M.; Wu, S.; Zaldivar, A.; Barnes, P.; Vasserman, L.; Hutchinson, B.; Spitzer, E.; Raji, I.~D.; and Gebru, T. 2019.
\newblock Model Cards for Model Reporting.
\newblock In \emph{Proceedings of the Conference on Fairness, Accountability, and Transparency}, FAT* '19, 220–229. New York, NY, USA: Association for Computing Machinery.
\newblock ISBN 9781450361255.

\bibitem[{Pepe et~al.(2024)Pepe, Nardone, Mastropaolo, Bavota, Canfora, and Di~Penta}]{10.1145/3643916.3644412}
Pepe, F.; Nardone, V.; Mastropaolo, A.; Bavota, G.; Canfora, G.; and Di~Penta, M. 2024.
\newblock How do Hugging Face Models Document Datasets, Bias, and Licenses? An Empirical Study.
\newblock In \emph{Proceedings of the 32nd IEEE/ACM International Conference on Program Comprehension}, ICPC '24, 370–381. New York, NY, USA: Association for Computing Machinery.
\newblock ISBN 9798400705861.

\bibitem[{Shah et~al.(2022)Shah, Chawla, Eidnani, Shah, Du, Chava, Raman, Smiley, Chen, and Yang}]{shah-etal-2022-flue}
Shah, R.; Chawla, K.; Eidnani, D.; Shah, A.; Du, W.; Chava, S.; Raman, N.; Smiley, C.; Chen, J.; and Yang, D. 2022.
\newblock When {FLUE} Meets {FLANG}: Benchmarks and Large Pretrained Language Model for Financial Domain.
\newblock In Goldberg, Y.; Kozareva, Z.; and Zhang, Y., eds., \emph{Proceedings of the 2022 Conference on Empirical Methods in Natural Language Processing}, 2322--2335. Abu Dhabi, United Arab Emirates: Association for Computational Linguistics.

\bibitem[{Singh et~al.(2023)Singh, Lodwal, Malwat, Thakur, and Singh}]{Singh2023UnlockingMI}
Singh, S.; Lodwal, H.; Malwat, H.; Thakur, R.; and Singh, M. 2023.
\newblock Unlocking Model Insights: A Dataset for Automated Model Card Generation.
\newblock \emph{ArXiv}, abs/2309.12616.

\bibitem[{Sipio et~al.(2024)Sipio, Rubei, Rocco, Ruscio, and Nguyen}]{disipio2024automated}
Sipio, C.~D.; Rubei, R.; Rocco, J.~D.; Ruscio, D.~D.; and Nguyen, P.~T. 2024.
\newblock Automated categorization of pre-trained models for software engineering: A case study with a Hugging Face dataset.
\newblock arXiv:2405.13185.

\bibitem[{Spearman(1904)}]{spearman}
Spearman, C. 1904.
\newblock The Proof and Measurement of Association between Two Things.
\newblock \emph{The American Journal of Psychology}, 15(1): 72--101.

\bibitem[{Swayamdipta et~al.(2020)Swayamdipta, Schwartz, Lourie, Wang, Hajishirzi, Smith, and Choi}]{swayamdipta-etal-2020-dataset}
Swayamdipta, S.; Schwartz, R.; Lourie, N.; Wang, Y.; Hajishirzi, H.; Smith, N.~A.; and Choi, Y. 2020.
\newblock Dataset Cartography: Mapping and Diagnosing Datasets with Training Dynamics.
\newblock In Webber, B.; Cohn, T.; He, Y.; and Liu, Y., eds., \emph{Proceedings of the 2020 Conference on Empirical Methods in Natural Language Processing (EMNLP)}, 9275--9293. Online: Association for Computational Linguistics.

\bibitem[{Taraghi et~al.(2024)Taraghi, Dorcelus, Foundjem, Tambon, and Khomh}]{Taraghi2024DeepLM}
Taraghi, M.; Dorcelus, G.; Foundjem, A.~T.; Tambon, F.; and Khomh, F. 2024.
\newblock Deep Learning Model Reuse in the HuggingFace Community: Challenges, Benefit and Trends.
\newblock \emph{ArXiv}, abs/2401.13177.

\bibitem[{Vaswani et~al.(2017)Vaswani, Shazeer, Parmar, Uszkoreit, Jones, Gomez, Kaiser, and Polosukhin}]{NIPS2017_3f5ee243}
Vaswani, A.; Shazeer, N.; Parmar, N.; Uszkoreit, J.; Jones, L.; Gomez, A.~N.; Kaiser, L.~u.; and Polosukhin, I. 2017.
\newblock Attention is All you Need.
\newblock In Guyon, I.; Luxburg, U.~V.; Bengio, S.; Wallach, H.; Fergus, R.; Vishwanathan, S.; and Garnett, R., eds., \emph{Advances in Neural Information Processing Systems}, volume~30. Curran Associates, Inc.

\bibitem[{Wang, Yang, and Wang(2023)}]{wang2023fingpt}
Wang, N.; Yang, H.; and Wang, C. 2023.
\newblock Fin{GPT}: Instruction Tuning Benchmark for Open-Source Large Language Models in Financial Datasets.
\newblock In \emph{NeurIPS 2023 Workshop on Instruction Tuning and Instruction Following}.

\bibitem[{Wolf et~al.(2020)Wolf, Debut, Sanh, Chaumond, Delangue, Moi, Cistac, Rault, Louf, Funtowicz, Davison, Shleifer, von Platen, Ma, Jernite, Plu, Xu, Le~Scao, Gugger, Drame, Lhoest, and Rush}]{wolf-etal-2020-transformers}
Wolf, T.; Debut, L.; Sanh, V.; Chaumond, J.; Delangue, C.; Moi, A.; Cistac, P.; Rault, T.; Louf, R.; Funtowicz, M.; Davison, J.; Shleifer, S.; von Platen, P.; Ma, C.; Jernite, Y.; Plu, J.; Xu, C.; Le~Scao, T.; Gugger, S.; Drame, M.; Lhoest, Q.; and Rush, A. 2020.
\newblock Transformers: State-of-the-Art Natural Language Processing.
\newblock In Liu, Q.; and Schlangen, D., eds., \emph{Proceedings of the 2020 Conference on Empirical Methods in Natural Language Processing: System Demonstrations}, 38--45. Online: Association for Computational Linguistics.

\bibitem[{Yang et~al.(2020)Yang, Dai, Yang, Carbonell, Salakhutdinov, and Le}]{yang2020xlnet}
Yang, Z.; Dai, Z.; Yang, Y.; Carbonell, J.; Salakhutdinov, R.; and Le, Q.~V. 2020.
\newblock XLNet: Generalized Autoregressive Pretraining for Language Understanding.
\newblock arXiv:1906.08237.

\bibitem[{Zhang et~al.(2023)Zhang, Deng, Liu, Pan, and Bing}]{zhang2023sentiment}
Zhang, W.; Deng, Y.; Liu, B.; Pan, S.~J.; and Bing, L. 2023.
\newblock Sentiment Analysis in the Era of Large Language Models: A Reality Check.
\newblock arXiv:2305.15005.

\end{thebibliography}
\newpage

\section*{Paper Checklist}

\begin{enumerate}

\item For most authors...
\begin{enumerate}
    \item  Would answering this research question advance science without violating social contracts, such as violating privacy norms, perpetuating unfair profiling, exacerbating the socio-economic divide, or implying disrespect to societies or cultures?
    \answerYes{Yes}
  \item Do your main claims in the abstract and introduction accurately reflect the paper's contributions and scope?
    \answerYes{Yes}
   \item Do you clarify how the proposed methodological approach is appropriate for the claims made? 
    \answerYes{Yes}
   \item Do you clarify what are possible artifacts in the data used, given population-specific distributions?
    \answerNA{N/A}
  \item Did you describe the limitations of your work?
    \answerYes{Yes}
  \item Did you discuss any potential negative societal impacts of your work?
    \answerNA{N/A}
      \item Did you discuss any potential misuse of your work?
    \answerNA{N/A}
    \item Did you describe steps taken to prevent or mitigate potential negative outcomes of the research, such as data and model documentation, data anonymization, responsible release, access control, and the reproducibility of findings?
    \answerYes{Yes}
  \item Have you read the ethics review guidelines and ensured that your paper conforms to them?
    \answerYes{Yes}
\end{enumerate}

\item Additionally, if your study involves hypotheses testing...
\begin{enumerate}
  \item Did you clearly state the assumptions underlying all theoretical results?
    \answerNA{N/A}
  \item Have you provided justifications for all theoretical results?
    \answerNA{N/A}
  \item Did you discuss competing hypotheses or theories that might challenge or complement your theoretical results?
    \answerNA{N/A}
  \item Have you considered alternative mechanisms or explanations that might account for the same outcomes observed in your study?
    \answerNA{N/A}
  \item Did you address potential biases or limitations in your theoretical framework?
    \answerNA{N/A}
  \item Have you related your theoretical results to the existing literature in social science?
    \answerNA{N/A}
  \item Did you discuss the implications of your theoretical results for policy, practice, or further research in the social science domain?
    \answerNA{N/A}
\end{enumerate}

\item Additionally, if you are including theoretical proofs...
\begin{enumerate}
  \item Did you state the full set of assumptions of all theoretical results?
    \answerNA{N/A}
	\item Did you include complete proofs of all theoretical results?
    \answerNA{N/A}
\end{enumerate}

\item Additionally, if you ran machine learning experiments...
\begin{enumerate}
  \item Did you include the code, data, and instructions needed to reproduce the main experimental results (either in the supplemental material or as a URL)?
    \answerYes{Yes,in the Abstract}
  \item Did you specify all the training details (e.g., data splits, hyperparameters, how they were chosen)?
    \answerYes{Yes, Refer Table~\ref{tab:datasets}}
     \item Did you report error bars (e.g., with respect to the random seed after running experiments multiple times)?
    \answerNA{N/A}
	\item Did you include the total amount of compute and the type of resources used (e.g., type of GPUs, internal cluster, or cloud provider)?
    \answerYes{Yes, in Hyperparameters section. refer Appendix~\ref{apx:hp}}
     \item Do you justify how the proposed evaluation is sufficient and appropriate to the claims made? 
    \answerYes{Yes}
     \item Do you discuss what is ``the cost`` of misclassification and fault (in)tolerance?
    \answerNA{N/A}
  
\end{enumerate}

\item Additionally, if you are using existing assets (e.g., code, data, models) or curating/releasing new assets, \textbf{without compromising anonymity}...
\begin{enumerate}
  \item If your work uses existing assets, did you cite the creators?
    \answerYes{Yes}
  \item Did you mention the license of the assets?
    \answerNA{N/A}
  \item Did you include any new assets in the supplemental material or as a URL?
    \answerNo{No}
  \item Did you discuss whether and how consent was obtained from people whose data you're using/curating?
    \answerNo{No, but all the data and assets have been taken from the Open-Source platforms like Hugging Face}
  \item Did you discuss whether the data you are using/curating contains personally identifiable information or offensive content?
    \answerNA{N/A}
\item If you are curating or releasing new datasets, did you discuss how you intend to make your datasets FAIR (see \citet{fair})?
\answerNA{N/A}
\item If you are curating or releasing new datasets, did you create a Datasheet for the Dataset (see \citet{gebru2021datasheets})? 
\answerNA{N/A}
\end{enumerate}

\item Additionally, if you used crowdsourcing or conducted research with human subjects, \textbf{without compromising anonymity}...
\begin{enumerate}
  \item Did you include the full text of instructions given to participants and screenshots?
    \answerYes{Yes, Refer the Instruction section in Appendix~\ref{apx:annot_instruct}}
  \item Did you describe any potential participant risks, with mentions of Institutional Review Board (IRB) approvals?
    \answerNA{N/A}
  \item Did you include the estimated hourly wage paid to participants and the total amount spent on participant compensation?
    \answerNo{No, all the participants are UG students who were willing to volunteer}
   \item Did you discuss how data is stored, shared, and deidentified?
   \answerNA{N/A}
\end{enumerate}

\end{enumerate}


\newpage

\appendix
\section{Appendix}
\label{sec:appendix}

\subsection{More details on Models Filtering}
\label{apx:mf}

\begin{figure}[!tbh]
     \centering
     \includegraphics[width=\linewidth]{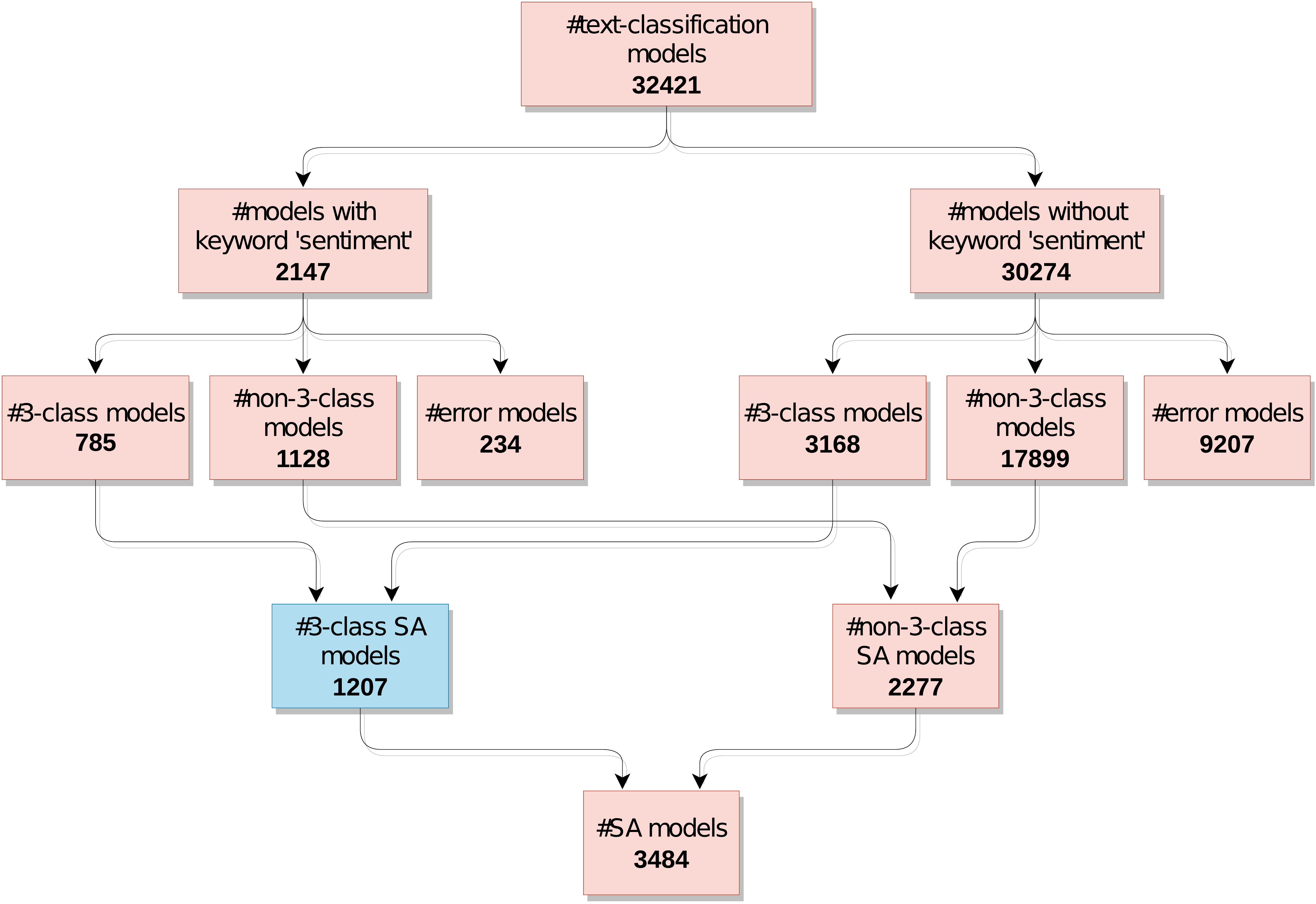}
     \caption{Models Filtering}
     \label{fig:models_tree}
\end{figure}
\begin{table}[ht]
\small
\centering
\resizebox{\linewidth}{!}{
    
\begin{tabular}{@{}cccc@{}}
\toprule
\textbf{Dataset}               & \textbf{Learning Rate} & \textbf{\begin{tabular}[c]{@{}c@{}}Per Device Train \\ Batch Size\end{tabular}} & \textbf{\#Train Epochs} \\ \midrule
\( \bm{\mathcal{D}_{t}} \)            & 1.9298e-05             & 128                                                                             & 5                       \\
\( \bm{\mathcal{D}_{f}} \) & 1.1470e-05             & 16                                                                              & 19                      \\
\( \bm{\mathcal{D}_{a}} \)  & 6.2177e-06             & 32                                                                              & 3                       \\ \bottomrule
\end{tabular}}
\caption{Optimal Hyperparameters}
\label{tab:hparams}
\end{table}

In filtering the models, we initially focused on those categorized under the ''Text Classification'' task on HF. These models are categorized into two groups: those containing the keyword ``sentiment'' and those without it.To distinguish between these groups of models of 3-class and non-3-class, we ran an example input text through each model in the respective groups and analyzed the number of output classes. This process result in four distinct groups (3-class-with-sentiment, 3-class-without-sentiment, non-3-class-with-sentiment, non-3-class-without-sentiment).

For models lacking the ``sentiment'' keyword in their names. we conducted additional checks to determine whether they qualified as SA models. Specifically, we examine their output classes to identify labels such as ``positive,'' ``very positive,'' ``neutral,'' ``negative,'' ``very negative,'' and variations like``pos,'' ``neg,'' ``neu,'' ``positiv,'' ``notr,'' ``negativ,'' or uppercase equivalents (e.g., "POSITIVE," "NEGATIVE"). Substring matching is used to include variations of these labels. Models producing such outputs are classified as SA models. Conversely, models with the ``sentiment'' keyword are directly classified as SA models. We restrict our analysis to 3-class models to maintain simplicity and avoid complexity introduced by models with more than three class labels. This process yielded a total of 1207 3-class SA models as depicted in Figure~\ref{fig:models_tree}, from which the $M_{500}$ is selected based on $r_{d}$.

\begin{table*}
  \centering
  \resizebox{\textwidth}{!}{
    \scalebox{22}{
      \begin{tabular}{|c|c|c|c|}
\hline

\textbf{Model ID} & \multicolumn{1}{c|}{\textbf{Model Name}}                          & \textbf{Model ID} & \multicolumn{1}{c|}{\textbf{Model Name}}                      \\ \hline
M1                & cardiffnlp/twitter-roberta-base-sentiment-latest                  & M51               & KBLab/robust-swedish-sentiment-multiclass                     \\ \hline
M2                & mrm8488/distilroberta-finetuned-financial-news-sentiment-analysis & M52               & amansolanki/autonlp-Tweet-Sentiment-Extraction-20114061       \\ \hline
M3                & cardiffnlp/twitter-roberta-base-sentiment                         & M53               & knkarthick/Sentiment-Analysis                                 \\ \hline
M4                & lxyuan/distilbert-base-multilingual-cased-sentiments-student      & M54               & Ammar-alhaj-ali/arabic-MARBERT-sentiment                      \\ \hline
M5                & ProsusAI/finbert                                                  & M55               & RogerKam/roberta\_RCADE\_fine\_tuned\_sentiment\_covid\_news  \\ \hline
M6                & yiyanghkust/finbert-tone                                          & M56               & Jean-Baptiste/roberta-large-financial-news-sentiment-en       \\ \hline
M7                & avichr/heBERT\_sentiment\_analysis                                & M57               & FinanceInc/auditor\_sentiment\_finetuned                      \\ \hline
M8                & finiteautomata/bertweet-base-sentiment-analysis                   & M58               & akoksal/bounti                                                \\ \hline
M9                & cardiffnlp/twitter-xlm-roberta-base-sentiment                     & M59               & MarieAngeA13/Sentiment-Analysis-BERT                          \\ \hline
M10               & finiteautomata/beto-sentiment-analysis                            & M60               & Venkatesh4342/distilbert-helpdesk-sentiment                   \\ \hline
M11               & Seethal/sentiment\_analysis\_generic\_dataset                     & M61               & Souvikcmsa/BERT\_sentiment\_analysis                          \\ \hline
M12               & oliverguhr/german-sentiment-bert                                  & M62               & CAMeL-Lab/bert-base-arabic-camelbert-ca-sentiment             \\ \hline
M13               & CAMeL-Lab/bert-base-arabic-camelbert-da-sentiment                 & M63               & ka05ar/banglabert-sentiment                                   \\ \hline
M14               & ahmedrachid/FinancialBERT-Sentiment-Analysis                      & M64               & cardiffnlp/xlm-twitter-politics-sentiment                     \\ \hline
M15               & cardiffnlp/twitter-xlm-roberta-base-sentiment-multilingual        & M65               & alexandrainst/da-sentiment-base                               \\ \hline
M16               & koheiduck/bert-japanese-finetuned-sentiment                       & M66               & RecordedFuture/Swedish-Sentiment-Fear                         \\ \hline
M17               & blanchefort/rubert-base-cased-sentiment                           & M67               & JP040/bert-german-sentiment-twitter                           \\ \hline
M18               & MonoHime/rubert-base-cased-sentiment-new                          & M68               & philschmid/distilbert-base-multilingual-cased-sentiment       \\ \hline
M19               & rohanrajpal/bert-base-multilingual-codemixed-cased-sentiment      & M69               & tkurtulus/TurkishAirlines-SentimentAnalysisModel              \\ \hline
M20               & snunlp/KR-FinBert-SC                                              & M70               & zainalq7/autotrain-NLU\_crypto\_sentiment\_analysis-754123133 \\ \hline
M21               & cointegrated/rubert-tiny-sentiment-balanced                       & M71               & sahri/indonesiasentiment                                      \\ \hline
M22               & sbcBI/sentiment\_analysis\_model                                  & M72               & yangheng/deberta-v3-large-absa-v1.1                           \\ \hline
M23               & soleimanian/financial-roberta-large-sentiment                     & M73               & Tejas3/distillbert\_base\_uncased\_80\_equal                  \\ \hline
M24               & ayameRushia/bert-base-indonesian-1.5G-sentiment-analysis-smsa     & M74               & turing-usp/FinBertPTBR                                        \\ \hline
M25               & ayameRushia/roberta-base-indonesian-1.5G-sentiment-analysis-smsa  & M75               & mr4/phobert-base-vi-sentiment-analysis                        \\ \hline
M26               & citizenlab/twitter-xlm-roberta-base-sentiment-finetunned          & M76               & elozano/tweet\_sentiment\_eval                                \\ \hline
M27               & yangheng/deberta-v3-base-absa-v1.1                                & M77               & cardiffnlp/bertweet-base-sentiment                            \\ \hline
M28               & philschmid/distilbert-base-multilingual-cased-sentiment-2         & M78               & hakonmh/sentiment-xdistil-uncased                             \\ \hline
M29               & climatebert/distilroberta-base-climate-sentiment                  & M79               & FinScience/FS-distilroberta-fine-tuned                        \\ \hline
M30               & j-hartmann/sentiment-roberta-large-english-3-classes              & M80               & rohanrajpal/bert-base-codemixed-uncased-sentiment             \\ \hline
M31               & Sigma/financial-sentiment-analysis                                & M81               & seara/rubert-base-cased-russian-sentiment                     \\ \hline
M32               & blanchefort/rubert-base-cased-sentiment-rusentiment               & M82               & cardiffnlp/xlm-roberta-base-sentiment-multilingual            \\ \hline
M33               & w11wo/indonesian-roberta-base-sentiment-classifier                & M83               & cardiffnlp/camembert-base-tweet-sentiment-fr                  \\ \hline
M34               & mdhugol/indonesia-bert-sentiment-classification                   & M84               & LondonStory/txlm-roberta-hindi-sentiment                      \\ \hline
M35               & HooshvareLab/bert-fa-base-uncased-sentiment-digikala              & M85               & shashanksrinath/News\_Sentiment\_Analysis                     \\ \hline
M36               & christian-phu/bert-finetuned-japanese-sentiment                   & M86               & xaqren/sentiment\_analysis                                    \\ \hline
M37               & lucas-leme/FinBERT-PT-BR                                          & M87               & emre/turkish-sentiment-analysis                               \\ \hline
M38               & CAMeL-Lab/bert-base-arabic-camelbert-mix-sentiment                & M88               & RogerKam/roberta\_fine\_tuned\_sentiment\_newsmtsc            \\ \hline
M39               & pin/senda                                                         & M89               & RecordedFuture/Swedish-Sentiment-Violence                     \\ \hline
M40               & sbcBI/sentiment\_analysis                                         & M90               & deepset/bert-base-german-cased-sentiment-Germeval17           \\ \hline
M41               & neuraly/bert-base-italian-cased-sentiment                         & M91               & cardiffnlp/xlm-roberta-base-tweet-sentiment-pt                \\ \hline
M42               & hw2942/bert-base-chinese-finetuning-financial-news-sentiment-v2   & M92               & KernAI/stock-news-distilbert                                  \\ \hline
M43               & poom-sci/WangchanBERTa-finetuned-sentiment                        & M93               & Timothy1337/finetuning-sentiment-all\_df                      \\ \hline
M44               & fergusq/finbert-finnsentiment                                     & M94               & blanchefort/rubert-base-cased-sentiment-rurewiews             \\ \hline
M45               & CouchCat/ma\_sa\_v7\_distil                                       & M95               & Voicelab/herbert-base-cased-sentiment                         \\ \hline
M46               & peerapongch/baikal-sentiment-ball                                 & M96               & svalabs/twitter-xlm-roberta-bitcoin-sentiment                 \\ \hline
M47               & samayash/finetuning-financial-news-sentiment                      & M97               & CAMeL-Lab/bert-base-arabic-camelbert-msa-sentiment            \\ \hline
M48               & seara/rubert-tiny2-russian-sentiment                              & M98               & bardsai/finance-sentiment-de-base                             \\ \hline
M49               & aari1995/German\_Sentiment                                        & M99               & kinit/slovakbert-sentiment-twitter                            \\ \hline
M50               & mdraw/german-news-sentiment-bert                                  & M100              & hanifnoerr/Fine-tuned-Indonesian-Sentiment-Classifier         \\ \hline
\end{tabular}
    }
  }
  \caption{Split-$1$ of $M_{500}$ Models}
  \label{tab:m100_models}
\end{table*}

\begin{table*}
  \centering
  \resizebox{\textwidth}{!}{
    \scalebox{22}{
      \begin{tabular}{|c|c|c|c|}
\hline

\textbf{Model ID} & \textbf{Model Name}                                             & \textbf{Model ID} & \textbf{Model Name}                                                                \\ \hline
M101              & ganeshkharad/gk-hinglish-sentiment                              & M151              & eevvgg/StanceBERTa                                                                 \\ \hline
M102              & Kyle1668/boss-sentiment-bert-base-uncased                       & M152              & mrcaelumn/yelp\_restaurant\_review\_sentiment\_analysis                            \\ \hline
M103              & oferweintraub/bert-base-finance-sentiment-noisy-search          & M153              & MMG/xlm-roberta-base-sa-spanish                                                    \\ \hline
M104              & RashidNLP/Finance-Sentiment-Classification                      & M154              & finiteautomata/beto-headlines-sentiment-analysis                                   \\ \hline
M105              & atowey01/hostel-reviews-sentiment-model                         & M155              & KernAI/community-sentiment-bert                                                    \\ \hline
M106              & bardsai/twitter-sentiment-pl-base                               & M156              & sismetanin/rubert\_conversational-ru-sentiment-rureviews                           \\ \hline
M107              & clips/republic                                                  & M157              & cardiffnlp/twitter-roberta-base-2021-124m-sentiment                                \\ \hline
M108              & larskjeldgaard/senda                                            & M158              & Tobias/bert-base-german-cased\_German\_Hotel\_sentiment                            \\ \hline
M109              & bardsai/finance-sentiment-zh-base                               & M159              & tyqiangz/indobert-lite-large-p2-smsa                                               \\ \hline
M110              & data354/camembert-fr-covid-tweet-sentiment-classification       & M160              & Venkatesh4342/distilbert-helpdesk-sentence-sentiment                               \\ \hline
M111              & nimaafshar/parsbert-fa-sentiment-twitter                        & M161              & seninoseno/rubert-base-cased-sentiment-study-feedbacks-solyanka                    \\ \hline
M112              & dnzblgn/BART\_Sentiment\_Classification                         & M162              & hazrulakmal/distilbert-optimised-finetuned-financial-sentiment                     \\ \hline
M113              & cardiffnlp/roberta-base-sentiment                               & M163              & Softechlb/Sent\_analysis\_CVs                                                      \\ \hline
M114              & rohanrajpal/bert-base-en-es-codemix-cased                       & M164              & Anthos23/FS-distilroberta-fine-tuned                                               \\ \hline
M115              & mstafam/fine-tuned-bert-financial-sentiment-analysis            & M165              & francisco-perez-sorrosal/distilbert-base-uncased-finetuned-with-spanish-tweets-clf \\ \hline
M116              & sismetanin/xlm\_roberta\_large-ru-sentiment-rureviews           & M166              & sismetanin/rubert-ru-sentiment-rureviews                                           \\ \hline
M117              & Davlan/naija-twitter-sentiment-afriberta-large                  & M167              & Tobias/bert-base-uncased\_English\_Hotel\_sentiment                                \\ \hline
M118              & blanchefort/rubert-base-cased-sentiment-med                     & M168              & z-dickson/multilingual\_sentiment\_newspaper\_headlines                            \\ \hline
M119              & abnersampaio/sentiment                                          & M169              & techthiyanes/Bert\_Bahasa\_Sentiment                                               \\ \hline
M120              & cardiffnlp/roberta-base-tweet-sentiment-en                      & M170              & RogerKam/roberta\_fine\_tuned\_sentiment\_sst3                                     \\ \hline
M121              & BVK97/Discord-NFT-Sentiment                                     & M171              & nikunjbjj/jd-resume-model                                                          \\ \hline
M122              & alger-ia/dziribert\_sentiment                                   & M172              & eevvgg/PaReS-sentimenTw-political-PL                                               \\ \hline
M123              & pysentimiento/roberta-es-sentiment                              & M173              & IAyoub/finetuning-bert-sentiment-reviews-2                                         \\ \hline
M124              & l3cube-pune/MarathiSentiment                                    & M174              & Cristian-dcg/beto-sentiment-analysis-finetuned-onpremise                           \\ \hline
M125              & m3hrdadfi/albert-fa-base-v2-sentiment-multi                     & M175              & NYTK/sentiment-ohb3-hubert-hungarian                                               \\ \hline
M126              & bardsai/finance-sentiment-pl-fast                               & M176              & cardiffnlp/xlm-v-base-tweet-sentiment-de                                           \\ \hline
M127              & Jorgeutd/bert-base-uncased-finetuned-surveyclassification       & M177              & warwickai/fin-perceiver                                                            \\ \hline
M128              & nickwong64/bert-base-uncased-finance-sentiment                  & M178              & mdeniz1/turkish-sentiment-analysis-bert-base-turkish-uncased                       \\ \hline
M129              & ncduy/phobert-large-finetuned-vietnamese\_students\_feedback    & M179              & TFLai/turkish-bert-128k-sentiment                                                  \\ \hline
M130              & Yah216/Sentiment\_Analysis\_CAMelBERT\_msa\_sixteenth\_HARD     & M180              & benjaminbeilharz/bert-base-uncased-sentiment-classifier                            \\ \hline
M131              & mwz/RomanUrduClassification                                     & M181              & RashidNLP/Amazon-Deberta-Base-Sentiment                                            \\ \hline
M132              & eevvgg/sentimenTw-political                                     & M182              & blinjrm/finsent                                                                    \\ \hline
M133              & yj2773/hinglish11k-sentiment-analysis                           & M183              & Kapiche/twitter-roberta-base-sentiment-latest                                      \\ \hline
M134              & cardiffnlp/bert-base-multilingual-cased-sentiment-multilingual  & M184              & bardsai/finance-sentiment-zh-fast                                                  \\ \hline
M135              & pysentimiento/roberta-targeted-sentiment-analysis               & M185              & EMBEDDIA/sloberta-tweetsentiment                                                   \\ \hline
M136              & Souvikcmsa/SentimentAnalysisDistillBERT                         & M186              & cassiepowell/RoBERTa-large-mnli-for-agreement                                      \\ \hline
M137              & candra/indobertweet-sentiment2                                  & M187              & oandreae/financial\_sentiment\_model                                               \\ \hline
M138              & sismetanin/sbert-ru-sentiment-rureviews                         & M188              & cardiffnlp/xlm-roberta-base-tweet-sentiment-fr                                     \\ \hline
M139              & rohanrajpal/bert-base-en-hi-codemix-cased                       & M189              & papepipopu/trading\_ai                                                             \\ \hline
M140              & sismetanin/sbert-ru-sentiment-krnd                              & M190              & Elron/deberta-v3-large-sentiment                                                   \\ \hline
M141              & DunnBC22/distilbert-base-uncased-Financial\_Sentiment\_Analysis & M191              & meghanabhange/Hinglish-Bert-Class                                                  \\ \hline
M142              & niksmer/RoBERTa-RILE                                            & M192              & EMBEDDIA/bertic-tweetsentiment                                                     \\ \hline
M143              & l3cube-pune/marathi-sentiment-md                                & M193              & Sonny4Sonnix/twitter-roberta-base-sentimental-analysis-of-covid-tweets             \\ \hline
M144              & ZiweiChen/FinBERT-FOMC                                          & M194              & EMBEDDIA/rubert-tweetsentiment                                                     \\ \hline
M145              & RogerKam/roberta\_fine\_tuned\_sentiment\_financial\_news       & M195              & arjuntheprogrammer/distilbert-base-multilingual-cased-sentiment-2                  \\ \hline
M146              & akahana/indonesia-sentiment-roberta                             & M196              & bowipawan/bert-sentimental                                                         \\ \hline
M147              & amphora/KorFinASC-XLM-RoBERTa                                   & M197              & zabiullah/autotrain-customers\_email\_sentiment-3449294006                         \\ \hline
M148              & abhishek/autonlp-swahili-sentiment-615517563                    & M198              & slisowski/stock\_sentiment\_hp                                                     \\ \hline
M149              & m3hrdadfi/albert-fa-base-v2-sentiment-digikala                  & M199              & bardsai/finance-sentiment-pl-base                                                  \\ \hline
M150              & classla/bcms-bertic-parlasent-bcs-ter                           & M200              & adam-chell/tweet-sentiment-analyzer \\ \hline
\end{tabular}
    }
  }
  \caption{Split-$2$ of $M_{500}$ Models}
  \label{tab:m200_models}
\end{table*}

\begin{table*}
  \centering
  \resizebox{\textwidth}{!}{
    \scalebox{21}{
      \begin{tabular}{|c|c|c|c|}
\hline

\textbf{Model ID} & \multicolumn{1}{c|}{\textbf{Model Name}}                          & \textbf{Model ID} & \multicolumn{1}{c|}{\textbf{Model Name}}                      \\ \hline

M201              & Manauu17/enhanced\_roberta\_sentiments\_es                                                        & M251              & ymcnabb/dutch\_threeway\_sentiment\_classification\_v2                                                          \\ \hline
M202              & \begin{tabular}[c]{@{}c@{}}lucaordronneau/twitter-roberta-base-sentiment-latest-finetuned\\-FG-SINGLE\_SENTENCE-NEWS\end{tabular}           & M252              & sudhanvasp/Sentiment-Analysis                                                                                   \\ \hline
M203              & cardiffnlp/twitter-roberta-base-dec2021-sentiment                                                 & M253              & hazrulakmal/bert-base-uncased-finetuned                                                                         \\ \hline
M204              & Splend1dchan/bert-base-uncased-slue-goldtrascription-e3-lr1e-4                                    & M254              & ramnika003/autotrain-sentiment\_analysis\_project-705021428                                                     \\ \hline
M205              & everyl12/crisis\_sentiment\_roberta                                                               & M255              & veb/twitch-roberta-base-sentiment-latest                                                                        \\ \hline
M206              & DunnBC22/bert-base-uncased-Twitter\_Sentiment\_Analysis\_v2                                       & M256              & Abubakari/finetuned-Sentiment-classfication-ROBERTA-model                                                       \\ \hline
M207              & Hyeonseo/ko-finance\_news\_classifier                                                             & M257              & kullackaan/sentiment-tweets                                                                                     \\ \hline
M208              & bardsai/twitter-sentiment-pl-fast                                                                 & M258              & chrommium/sbert\_large-finetuned-sent\_in\_news\_sents\_3lab                                                    \\ \hline
M209              & EMBEDDIA/english-tweetsentiment                                                                   & M259              & rasmodev/Covid-19\_Sentiment\_Analysis\_RoBERTa\_Model                                                          \\ \hline
M210              & DSI/human-directed-sentiment                                                                      & M260              & Mawulom/Fine-Tuned-Bert\_Base\_Cased\_Sentiment\_Analysis                                                       \\ \hline
M211              & hw2942/bert-base-chinese-finetuning-financial-news-sentiment                                      & M261              & ikoghoemmanuell/finetuned\_sentiment\_modell                                                                    \\ \hline
M212              & Tejas3/distillbert\_base\_uncased\_80                                                             & M262              & mtyrrell/CPU\_Transport\_GHG\_Classifier                                                                        \\ \hline
M213              & bardsai/finance-sentiment-es-base                                                                 & M263              & mlkorra/obgv-gender-bert-hi-en                                                                                  \\ \hline
M214              & sismetanin/rubert\_conversational-ru-sentiment-sentirueval2016                                    & M264              & KarelDO/lstm.CEBaB\_confounding.uniform.absa.5-class.seed\_42                                                   \\ \hline
M215              & arize-ai/distilbert\_reviews\_with\_language\_drift                                               & M265              & \begin{tabular}[c]{@{}c@{}}francisco-perez-sorrosal/dccuchile-distilbert-base-spanish-\\uncased-finetuned-with-spanish-tweets-clf\end{tabular}            \\ \hline
M216               & \begin{tabular}[c]{@{}c@{}} Theivaprakasham/sentence-transformers-\\paraphrase-MiniLM-L6-v2-twitter\_sentiment  \end{tabular}                  & M266              & Hyeonseo/finance\_news\_classifier                                                                              \\ \hline
M217              & gabrielyang/finance\_news\_classifier-KR\_v7                                                      & M267              & LYTinn/finetuning-sentiment-model-tweet-gpt2                                                                    \\ \hline
M218              & researchaccount/sa\_sub4                                                                          & M268              & \begin{tabular}[c]{@{}c@{}}francisco-perez-sorrosal/dccuchile-distilbert-base-spanish-\\uncased-finetuned-with-spanish-tweets-clf-cleaned-ds\end{tabular} \\ \hline
M219              & \begin{tabular}[c]{@{}c@{}}Abdelrahman-Rezk/emotion-english-distilroberta-base-fine\_\\tuned\_for\_amazon\_reviews\_english\_3\end{tabular} & M269              & cassiepowell/LaBSE-for-agreement                                                                                \\ \hline
M220              & \begin{tabular}[c]{@{}c@{}}TankuVie/bert-base-multilingual-\\uncased-vietnamese\_sentiment\_analysis \end{tabular}                           & M270              &
\begin{tabular}[c]{@{}c@{}}
francisco-perez-sorrosal/distilbert-base-multilingual-\\cased-finetuned-with-spanish-tweets-clf  
\end{tabular} \\ \hline
M221              & researchaccount/sa\_sub1                                                                          & M271              & cardiffnlp/xlm-v-base-tweet-sentiment-en                                                                        \\ \hline
M222              & laurens88/finetuning-crypto-tweet-sentiment-test2                                                 & M272              & \begin{tabular}[c]{@{}c@{}}francisco-perez-sorrosal/distilbert-base-multilingual-\\cased-finetuned-with-spanish-tweets-clf-cleaned-ds    \end{tabular}    \\ \hline
M223              & Narsil/finbert2                                                                                   & M273              & dpeinado/twitter-roberta-base-sentiment-latest-apple-opinions                                                   \\ \hline
M224              & Anthos23/my-awesome-model                                                                         & M274              & course5i/SEAD-L-6\_H-384\_A-12-mnli                                                                             \\ \hline
M225              & aXhyra/presentation\_sentiment\_1234567                                                           & M275              & DunnBC22/distilbert-base-uncased-US\_Airline\_Twitter\_Sentiment\_Analysis                                      \\ \hline
M226              & hw2942/bert-base-chinese-finetuning-financial-news-sentiment-test                                 & M276              & Psunrise/finetuning-customer-sentiment-model-300-samples                                                        \\ \hline
M227              & ScriptEdgeAI/MarathiSentiment-Bloom-560m                                                          & M277              & AGudden/xlm-roberta-base-finetuned-marc                                                                         \\ \hline
M228              & Teeto/reviews-classification                                                                      & M278              & course5i/SEAD-L-6\_H-256\_A-8-mnli                                                                              \\ \hline
M229              & hilmansw/indobert-finetuned-sentiment-happiness-index                                             & M279              & eduardopds/distilbert-base-uncased-tweets                                                                       \\ \hline
M230              & Narsil/finbert-slow                                                                               & M280              & maclean-connor96/feedier-french-books                                                                           \\ \hline
M231              & BramVanroy/bert-base-multilingual-cased-hebban-reviews                                            & M281              & abnersampaio/sentimentv2                                                                                        \\ \hline
M232              & cnut1648/biolinkbert-mnli                                                                         & M282              & CultureBERT/roberta-large-clan                                                                                  \\ \hline
M233              & sismetanin/rubert\_conversational-ru-sentiment-krnd                                               & M283              & CultureBERT/roberta-large-hierarchy                                                                             \\ \hline
M234              & Cloudy1225/stackoverflow-roberta-base-sentiment                                                   & M284              & jayantapaul888/twitter-data-microsoft-deberta-base-mnli-sentiment-finetuned-memes                               \\ \hline
M235              & poerwiyanto/bert-base-indonesian-522M-finetuned-sentiment                                         & M285              & francisco-perez-sorrosal/distilbert-base-uncased-finetuned-with-spanish-tweets-clf-cleaned-ds                   \\ \hline
M236              & mirfan899/da\_spacy\_sentiment                                                                    & M286              & traptrip/rosatom\_survey\_sentiment\_classifier                                                                 \\ \hline
M237              & GhylB/Sentiment\_Analysis\_DistilBERT                                                             & M287              & tanoManzo/roberta-attitude                                                                                      \\ \hline
M238              & Vasanth/bert-stock-sentiment-analyzer                                                             & M288              & Christiansg/finetuning-sentiment\_spanish-amazon-group23                                                        \\ \hline
M239              & cardiffnlp/xlm-v-base-tweet-sentiment-pt                                                          & M289              & sasha/autotrain-RobertaBaseTweetEval-1281048989                                                                 \\ \hline
M240              & sakasa007/finetuning-sentiment-text-mining                                                        & M290              & IngeniousArtist/distilbert-finance                                                                              \\ \hline
M241              & CK42/sentiment\_analysis\_sbcBI                                                                   & M291              & Anthos23/FS-finbert-fine-tuned-f1                                                                               \\ \hline
M242              & kartashoffv/vashkontrol-sentiment-rubert                                                          & M292              & cardiffnlp/xlm-roberta-base-tweet-sentiment-es                                                                  \\ \hline
M243              & dogruermikail/bert-fine-tuned-stock-sentiment-uncased                                             & M293              & cardiffnlp/xlm-roberta-base-tweet-sentiment-en                                                                  \\ \hline
M244              & researchaccount/sa\_sub5                                                                          & M294              & hw2942/bert-base-chinese-finetuning-financial-news-sentiment-test1                                              \\ \hline
M245              & madmancity/bert2                                                                                  & M295              & sasha/autotrain-RobertaBaseTweetEval-1281048990                                                                 \\ \hline
M246              & bright1/fine-tuned-twitter-Roberta-base-sentiment                                                 & M296              & CultureBERT/roberta-large-adhocracy                                                                             \\ \hline
M247              & debashish68/roberta-sent-generali                                                                 & M297              & ruanchaves/bert-base-portuguese-cased-porsimplessent                                                            \\ \hline
M248              & BramVanroy/xlm-roberta-base-hebban-reviews                                                        & M298              & sasha/autotrain-RobertaBaseTweetEval-1281048986                                                                 \\ \hline
M249              & fassahat/anferico-bert-for-patents-finetuned-557k-patent-sentences                                & M299              & bardsai/finance-sentiment-ja-base                                                                               \\ \hline
M250              &\begin{tabular}[c]{@{}c@{}} KarelDO/lstm.CEBaB\_confounding.food\_service\\\_positive.absa.5-class.seed\_42    \end{tabular}                  & M300              & ruanchaves/bert-base-portuguese-cased-assin-entailment \\ \hline

\end{tabular}
    }
  }
  \caption{Split-$3$ of $M_{500}$ Models}
  \label{tab:m300_models}
\end{table*}

\begin{table*}
  \centering
  \resizebox{\textwidth}{!}{
    \scalebox{21}{
      \begin{tabular}{|c|c|c|c|}

\hline
\textbf{Model ID} & \textbf{Model Name}                                                         & \textbf{Model ID} & \textbf{Model Name}                                                                                                 \\ \hline
M301              & ruanchaves/mdeberta-v3-base-assin-entailment                                & M351              & deansaco/Roberta-base-financial-sentiment-analysis                                                                  \\ \hline
M302              & Nelver28/sentiment-analysis-positive-mixed-negative                         & M352              & bvint/autotrain-sphere-lecture-demo-1671659193                                                                      \\ \hline
M303              & hazrulakmal/benchmark-finetuned-distilbert                                  & M353              & Svetlana0303/Classfication\_RoBERTa                                                                                 \\ \hline
M304              & gyesibiney/covid-tweet-sentimental-Analysis-roberta                         & M354              & jayantapaul888/twitter-data-pysentimiento-robertuito-sentiment-finetuned-memes                                      \\ \hline
M305              & Rem59/autotrain-Test\_2-789524315                                           & M355              & dshin/my\_awesome\_model                                                                                            \\ \hline
M306              & ozoora/rubert-4.1poi                                                        & M356              & ongknsro/ACARISBERT-DistilBERT                                                                                      \\ \hline
M307              & profoz/deploy-mlops-demo                                                    & M357              & Cincin-nvp/NusaX-senti\_XLM-R                                                                                       \\ \hline
M308              & CultureBERT/roberta-large-market                                            & M358              & tanoManzo/distilbert-attitude                                                                                       \\ \hline
M309              & philschmid/finbert-tone-endpoint-ds                                         & M359              & slickdata/finetuned-Sentiment-classfication-ROBERTA-model                                                           \\ \hline
M310              & ruanchaves/mdeberta-v3-base-porsimplessent                                  & M360              & vectorizer/sentiment\_analysis\_93k\_entries                                                                        \\ \hline
M311              & ldeb/solved-finbert-tone                                                    & M361              & Bennet1996/finetuning-ESG-sentiment-model-distilbert                                                                \\ \hline
M312              & Venkatesh4342/bert-base-uncased-finetuned-fin                               & M362              & \begin{tabular}[c]{@{}c@{}}adnanakbr/bert-base-multilingual-uncased-sentiment-fine\_\\tuned\_for\_amazon\_english\_reviews\_on\_200K\_review\_v2\end{tabular} \\ \hline
M313              & cardiffnlp/xlm-roberta-base-tweet-sentiment-ar                              & M363              & intanm/sa10-clm-20230403-001-3                                                                                      \\ \hline
M314              & DingYao/autotrain-fbert-singlish-5-1943965533                               & M364              & cardiffnlp/mbert-base-tweet-sentiment-fr                                                                            \\ \hline
M315              & sasha/autotrain-BERTBase-TweetEval-1281248998                               & M365              & chinmayapani/panich                                                                                                 \\ \hline
M316              & MavisAJ/Sentiment\_analysis\_roberta\_model                                 & M366              & davidchiii/news-headlines                                                                                           \\ \hline
M317              & UchihaMadara/phobert-finetuned-sentiment-analysis                           & M367              & Svetlana0303/Classfication\_longformer                                                                              \\ \hline
M318              & cardiffnlp/mbert-base-tweet-sentiment-pt                                    & M368              & HerbertAIHug/Finetuned-Roberta-Base-Sentiment-identifier                                                            \\ \hline
M319              & Venkatesh4342/xlm-roberta-helpdesk-sentiment                                & M369              & cardiffnlp/xlm-v-base-tweet-sentiment-fr                                                                            \\ \hline
M320              & sasha/autotrain-BERTBase-TweetEval-1281248996                               & M370              & Svetlana0303/Classfication\_AlBERT                                                                                  \\ \hline
M321              & DunnBC22/fnet-large-Financial\_Sentiment\_Analysis\_v3                      & M371              & gr8testgad-1/sentiment\_analysis                                                                                    \\ \hline
M322              & DingYao/autotrain-fbert-singlish-1755361190                                 & M372              & cardiffnlp/mbert-base-tweet-sentiment-ar                                                                            \\ \hline
M323              & IsaacSarps/sentiment\_analysis                                              & M373              & ecabott/nepali-sentiment-analyzer                                                                                   \\ \hline
M324              & sasha/autotrain-DistilBERT-TweetEval-1281148991                             & M374              & cruiser/distilbert\_model\_kaggle\_200\_epoch                                                                       \\ \hline
M325              & cardiffnlp/xlm-v-base-tweet-sentiment-es                                    & M375              & nlp-chula/sentiment-finnlp-th                                                                                       \\ \hline
M326              & cardiffnlp/xlm-roberta-base-tweet-sentiment-de                              & M376              & giotvr/portuguese-nli-3-labels                                                                                      \\ \hline
M327              & DingYao/autotrain-fbert-singlish-2-1937065404                               & M377              & Pendo/finetuned-Sentiment-classfication-ROBERTA-Base-model                                                          \\ \hline
M328              & dogruermikail/bert-fine-tuning-sentiment-stocks-analyis-uncased             & M378              & raygx/BERT-NepSA-domainAdapt                                                                                        \\ \hline
M329              & l3cube-pune/marathi-sentiment-subtitles                                     & M379              & incredible45/News-Sentimental-model-Buy-Neutral-Sell                                                                \\ \hline
M330              & ultraleow/cloud4bert                                                        & M380              & cardiffnlp/xlm-v-base-tweet-sentiment-ar                                                                            \\ \hline
M331              & aarnphm/multi-length-text-classification-pipeline                           & M381              & intanm/mlm\_v1\_20230327\_fin\_sa\_10                                                                               \\ \hline
M332              & israel/testing\_model                                                       & M382              & RogerB/kin-sentiC                                                                                                   \\ \hline
M333              & raygx/BERT-NepSA-T2                                                         & M383              & intanm/sa100-mlm-20230403-001-2                                                                                     \\ \hline
M334              & devtanumisra/finetuning-sentiment-model-deberta-smote                       & M384              & raoh/yiran-nlp4                                                                                                     \\ \hline
M335              & cardiffnlp/xlm-roberta-base-tweet-sentiment-it                              & M385              & CMunch/fine\_tuned\_dota                                                                                            \\ \hline
M336              & sairahul5223/autotrain-auto-train-intent-classification-20220928-1584756071 & M386              & asaderu-ai/ssclass\_best                                                                                            \\ \hline
M337              & sasha/autotrain-DistilBERT-TweetEval-1281148992                             & M387              & salohnana2018/ABSA-single-domainAdapt-bert-base-MARBERT2-HARD                                                       \\ \hline
M338              & cardiffnlp/mbert-base-tweet-sentiment-it                                    & M388              & cruiser/roberta\_tweet\_eval\_finetuned                                                                             \\ \hline
M339              & antypasd/twitter-roberta-base-sentiment-earthquake                          & M389              & sara-nabhani/ML-ns-bert-base-uncased                                                                                \\ \hline
M340              & QuophyDzifa/Sentiment-Analysis-Model                                        & M390              & raygx/distilBERT-NepSA                                                                                              \\ \hline
M341              & fassahat/anferico-bert-for-patents-finetuned-150k-sentences                 & M391              & Ausbel/Vaccine-tweet-sentiments-analysis-model-2                                                                    \\ \hline
M342              & Himanshusingh/finetunedfinbert-model                                        & M392              & salohnana2018/MARBERTV02ABSAnew                                                                                     \\ \hline
M343              & SiddharthaM/twitter-data-bert-base-multilingual-uncased-hindi-only-memes    & M393              & duwuonline/mymodel-classify-sentiment                                                                               \\ \hline
M344              & cardiffnlp/xlm-v-base-tweet-sentiment-it                                    & M394              & AlonCohen/RuSentNE-test                                                                                             \\ \hline
M345              & fassahat/distilbert-base-uncased-finetuned-557k-patent-sentences            & M395              & memotirre90/Equipo16\_gpt2-HotelSentiment                                                                           \\ \hline
M346              & DunnBC22/fnet-base-Financial\_Sentiment\_Analysis                           & M396              & neojex/testing                                                                                                      \\ \hline
M347              & brema76/vaccine\_event\_it                                                  & M397              & xyu1163/Testmodel\_sentiment\_with\_3\_labels                                                                       \\ \hline
M348              & UholoDala/tweet\_sentiments\_analysis\_bert                                 & M398              & jcy204/wind\_model2                                                                                                 \\ \hline
M349              & W4nkel/microsoftTurkishTrain                                                & M399              & dhikaardianto/TiktokSentimentIndoBertTwitter                                                                        \\ \hline
M350              & aisyahhrazak/distilbert-mooc-review-sentiment                               & M400              & midwinter73/dipterv6                                                                                                \\ \hline

\end{tabular}
    }
  }
  \caption{Split-$4$ of $M_{500}$ Models}
  \label{tab:m400_models}
\end{table*}

 \begin{table*}
  \centering
  \resizebox{\textwidth}{!}{
    \scalebox{22}{
      \begin{tabular}{|c|c|c|c|}

\hline
\textbf{Model ID} & \multicolumn{1}{c|}{\textbf{Model Name}}                          & \textbf{Model ID} & \multicolumn{1}{c|}{\textbf{Model Name}}                                                                                                    \\ \hline
M401              & salohnana2018/ARABERT02-best-trail                                    & M451              & yanezh/twiiter\_try11\_fold2                                                                                          \\ \hline
M402              & cruiser/final\_model                                                  & M452              & Yt99/SFinBERT                                                                                                         \\ \hline
M403              & bekbote/autotrain-dl-phrasebank-53436126044                           & M453              & yanezh/twiiter\_try5\_fold0                                                                                           \\ \hline
M404              & senfu/bert-base-uncased-top-pruned-mnli                               & M454              & abnersampaio/sentiment3                                                                                               \\ \hline
M405              & abnersampaio/sentiment3.5                                             & M455              & yanezh/twiiter\_try10\_fold4                                                                                          \\ \hline
M406              & intanm/fin-sa-post-trained-indobert-base-finreport                    & M456              & intanm/fin-sa-post-trained-indobert-base-finnews-p2-001                                                               \\ \hline
M407              & kreynolds03/Solomon                                                   & M457              & yanezh/twiiter\_try13\_fold1                                                                                          \\ \hline
M408              & rod16/v1\_finetuning-sentiment-model-news-samples                     & M458              & yanezh/twiiter\_try4\_fold2                                                                                           \\ \hline
M409              & luissgtorres/Bert\_sentiment\_analysis\_Indata                        & M459              & langecod/Financial\_Phrasebank\_RoBERTa                                                                               \\ \hline
M410              & intanm/mlm\_v1\_20230327\_fin\_sa\_70                                 & M460              & BeChi87/train\_model\_Lastest1                                                                                        \\ \hline
M411              & Svetlana0303/Classfication\_electra                                   & M461              & intanm/fin-sa-post-trained-indobert-base-combined                                                                     \\ \hline
M412              & intanm/mlm\_v1\_20230327\_fin\_sa\_20                                 & M462              & evendivil/finetuning-sentiment-model-3000-samples                                                                     \\ \hline
M413              & Sonny4Sonnix/movie\_sentiment\_trainer                                & M463              & yanezh/twiiter\_try13\_fold3                                                                                          \\ \hline
M414              & A-Funakoshi/bert-finetuned-multilingual-sentiments                    & M464              & gabrielkytz/finetuning-sentiment-model-3000-samples                                                                   \\ \hline
M415              & salohnana2018/ARABERT02                                               & M465              & ErisGrey/my\_models                                                                                                   \\ \hline
M416              & intanm/mlm\_v1\_20230327\_fin\_sa\_100                                & M466              & HerbertAIHug/finetuned\_sentiment\_analysis\_modell                                                                   \\ \hline
M417              & intanm/mlm\_v1\_20230327\_fin\_sa\_90                                 & M467              & yanezh/twiiter\_try5\_fold3                                                                                           \\ \hline
M418              & A-Funakoshi/bert-base-japanese-v3-wrime-sentiment                     & M468              & \begin{tabular}[c]{@{}c@{}}Abdelrahman-Rezk/emotion-english-distilroberta-base-\\fine\_tuned\_for\_amazon\_english\_reviews\_V03\_on\_100K\_review \end{tabular}\\ \hline
M419              & Q317/EmoraBert                                                        & M469              & yanezh/twiiter\_try13\_fold4                                                                                          \\ \hline
M420              & NewtonKimathi/Covid\_Vaccine\_Sentiment\_Analysis\_Roberta\_Model     & M470              & yanezh/twiiter\_try4\_fold3                                                                                           \\ \hline
M421              & yanezh/twiiter\_try13\_fold2                                          & M471              & yanezh/twiiter\_try6\_fold1                                                                                           \\ \hline
M422              & quesmed/tone                                                          & M472              & BeChi87/train\_model\_Lastestttt                                                                                      \\ \hline
M423              & intanm/baseline\_fin\_sa\_60                                          & M473              & intanm/fin-sa-post-trained-indobert-large-finnews-p2-002                                                              \\ \hline
M424              & intanm/mlm\_v1\_20230327\_fin\_sa\_60                                 & M474              & jordankrishnayah/robertaSentimentBiasTestwithMBIC                                                                     \\ \hline
M425              & raygx/distilGPT-NepSA                                                 & M475              & \begin{tabular}[c]{@{}c@{}}adnanakbr/emotion-english-distilroberta-base-\\fine\_tuned\_for\_amazon\_english\_reviews\_on\_200K\_review \end{tabular}            \\ \hline
M426              & gsl22/sentiment-analysis-v1                                           & M476              & BeChi87/train\_model\_noStopWord                                                                                      \\ \hline
M427              & intanm/100-finSA-finLM-v1-001                                         & M477              & \begin{tabular}[c]{@{}c@{}}adnanakbr/emotion-english-distilroberta-base-\\fine\_tuned\_for\_amazon\_english\_reviews\_on\_200K\_review\_v3 \end{tabular}         \\ \hline
M428              & intanm/baseline\_fin\_sa\_100                                         & M478              & intanm/fin-sa-post-trained-indobert-base-finreports-p2-001                                                            \\ \hline
M429              & venetis/deberta-v3-base-finetuned-3d-sentiment                        & M479              & venetis/distilroberta-base-finetuned-3d-sentiment                                                                     \\ \hline
M430              & manvik28/FinBERT\_Tuned                                               & M480              & gabrielkytz/novo                                                                                                      \\ \hline
M431              & NewtonKimathi/Covid\_Vaccine\_Sentiment\_Analysis\_Bert\_based\_Model & M481              & sshs/phobert-base-vietnamese-sentiment                                                                                \\ \hline
M432              & raygx/GNePT-NepSA                                                     & M482              & BeChi87/train\_model\_Lastest\_new                                                                                    \\ \hline
M433              & santis2/phrasebank-sentiment-analysis                                 & M483              & jordankrishnayah/fullManifesto-ROBERTAsentiment2e                                                                     \\ \hline
M434              & venetis/electra-base-discriminator-finetuned-3d-sentiment             & M484              & BeChi87/train\_model\_Lastest\_21                                                                                     \\ \hline
M435              & AlonCohen/RuSentNE-iter-2                                             & M485              & BeChi87/train\_model\_Lastest\_16\_8\_lan2                                                                            \\ \hline
M436              & intanm/baseline\_fin\_sa\_70                                          & M486              & BeChi87/train\_model\_Lastest\_16\_8\_lan3                                                                            \\ \hline
M437              & jcy204/cold\_model                                                    & M487              & maegancp/finetuning-sentiment-model                                                                                   \\ \hline
M438              & Beanz1935/finetuning-sentiment-model-3000-samples                     & M488              & mrfakename/distilroberta-financial-news-tweets-sentiment-analysis                                                     \\ \hline
M439              & yanezh/twiiter\_try15\_fold2                                          & M489              & Jonathan0528/bert-base-uncased-financial-news-sentiment                                                               \\ \hline
M440              & intanm/baseline\_fin\_sa\_30                                          & M490              & rhythm00/HF\_tutor                                                                                                    \\ \hline
M441              & Robher51/Model3                                                       & M491              & \begin{tabular}[c]{@{}c@{}}salohnana2018/ABSA-Pair-domainAdapt-CAMEL-MSA-HARD-\\ImbalancedHandling-absoluteWeights-gamma1-dropout01     \end{tabular}           \\ \hline
M442              & yanezh/twiiter\_try15\_fold4                                          & M492              & \begin{tabular}[c]{@{}c@{}}salohnana2018/ABSA-Pair-domainAdapt-CAMEL-MSA-HARD-\\ImbalancedHandling-inv-prop-weights-gamma1            \end{tabular}                 \\ \hline
M443              & tiwanyan/distil\_bert\_fine\_tune                                     & M493              & nuriafari/my\_model                                                                                                   \\ \hline
M444              & intanm/fin-sa-post-trained-indobert-base-finnews-5                    & M494              & \begin{tabular}[c]{@{}c@{}} salohnana2018/ABSA-SentencePair-DAPT-HARD-SemEval-\\bert-base-Camel-MSA-run3        \end{tabular}                                             \\ \hline
M445              & yanezh/twiiter\_try15\_fold3                                          & M495              & \begin{tabular}[c]{@{}c@{}}salohnana2018/ABSA-Pair-domainAdapt-Qarib-HARD-ImbalancedHandling-\\absoluteWeights-gamma1-dropout01     \end{tabular}                \\ \hline
M446              & cruiser/distilbert\_final\_config\_dropout                            & M496              & \begin{tabular}[c]{@{}c@{}}salohnana2018/ABSA-Pair-domainAdapt-Camel-MSA-\\HARD-ImbalancedHandling-absoluteWeights-gamma1-run2      \end{tabular}                 \\ \hline
M447              & Q317/EmoraBert4                                                       & M497              & \begin{tabular}[c]{@{}c@{}}salohnana2018/ABSA-Pair-domainAdapt-Camel-MSA-\\HARD-ImbalancedHandling-absoluteWeights-gamma1-fulldata    \end{tabular}               \\ \hline
M448              & raygx/xlmRoBERTa-NepSA                                                & M498              & \begin{tabular}[c]{@{}c@{}}salohnana2018/ABSA-Pair-domainAdapt-Camel-MSA-\\HARD-ImbalancedHandling-absoluteWeights-gamma1-run1  \end{tabular}                     \\ \hline
M449              & yanezh/twiiter\_try10\_fold0                                          & M499              & \begin{tabular}[c]{@{}c@{}}salohnana2018/ABSA-Pair-domainAdapt-CAMEL-MSA-\\HARD-ImbalancedHandling-norm\_inv\_const5-gamma3     \end{tabular}                     \\ \hline
M450              & yanezh/twiiter\_try11\_fold3                                          & M500              & Anwaarma/autotrain-testsen-65460136124                                                                                \\ \hline
\end{tabular}
    }
    
  }
  \caption{Split-$5$ of $M_{500}$ Models}
  \label{tab:m500_models}
\end{table*}

\subsection{In-Domain/Out-Domain Evaluation of \bm{$M_{500}$}}
\label{apx:dom_eval}

Since most of the model authors have not reported their training datasets, it becomes difficult to determine whether the model evaluation on a test dataset is in-domain or out-domain. This is the reason we avoided conducting an in-domain/out-domain evaluation. However, in this section, we present the results (refer Table~\ref{tab:corr_hf_extended}) based on the available information for the models in their model cards. The results may not be conclusive since the training data for most models in $M_{500}$ is unknown. Furthermore, as depicted in Figure~\ref{fig:te_e} and similar figures that follow, it is evident that performance is not solely dependent on whether the evaluation is In-Domain or Out-Domain. For instance, Figure~\ref{fig:te_e} illustrates that some models in Out-Domain evaluation perform better than models in In-Domain evaluation.

\begin{table*}[!htbp]
\resizebox{\textwidth}{!}{
\centering

\begin{tabular}{@{}cccccccccccccccccccc@{}}
\toprule
\multirow{3}{*}{}                                       & \multirow{3}{*}{}      & \multicolumn{6}{c}{$\bm{\mathcal{D}_{t}}$}                                                       & \multicolumn{6}{c}{$\bm{\mathcal{D}_{a}}$}                                             & \multicolumn{6}{c}{$\bm{\mathcal{D}_{f}}$}                                            \\ \cmidrule(l){3-8}\cmidrule(l){9-14}\cmidrule(l){15-20} 
                                                        &                        & \multicolumn{3}{c}{\textbf{In-Domain}}        & \multicolumn{3}{c}{\textbf{Out-Domain}}       & \multicolumn{3}{c}{\textbf{In-Domain}}        & \multicolumn{3}{c}{\textbf{Out-Domain}}       & \multicolumn{3}{c}{\textbf{In-Domain}}        & \multicolumn{3}{c}{\textbf{Out-Domain}}       \\ \cmidrule(l){3-5}\cmidrule(l){6-8}\cmidrule(l){9-11}\cmidrule(l){12-14}\cmidrule(l){15-17}\cmidrule(l){18-20} 
                                                        &                        & \textbf{Easy} & \textbf{Ambi} & \textbf{Hard} & \textbf{Easy} & \textbf{Ambi} & \textbf{Hard} & \textbf{Easy} & \textbf{Ambi} & \textbf{Hard} & \textbf{Easy} & \textbf{Ambi} & \textbf{Hard} & \textbf{Easy} & \textbf{Ambi} & \textbf{Hard} & \textbf{Easy} & \textbf{Ambi} & \textbf{Hard} \\ \midrule
\multicolumn{1}{c|}{\multirow{3}{*}{\rotatebox{90}{\textbf{Evaluate}}}} & $\bm{r_{d}}$ & 0.058         & 0.235         & 0.271         & 0.214*        & 0.186*        & 0.057         & -0.714        & -0.714        & -0.642        & 0.217*        & 0.188*        & 0.162*        & 0.113         & -             & -             & 0.107*        & -             & -             \\
\multicolumn{1}{c|}{}                                   & $\bm{r_{l}}$     & -0.035        & 0.051         & 0.250         & 0.159*        & 0.121*        & 0.077         & -0.883*       & -0.883*       & -0.270        & 0.140*        & 0.154*        & 0.173*        & 0.230         & -             & -             & 0.153*        & -             & -             \\
\multicolumn{1}{c|}{}                                   & $\bm{r_{da}}$     & 0.105         & 0.146         & 0.153         & -0.098*       & -0.089        & -0.063        & 0.178         & 0.178         & 0.393         & -0.084        & -0.081        & -0.135*       & -0.576*       & -             & -             & -0.086        & -             & -             \\ \midrule
\multicolumn{1}{c|}{\multirow{3}{*}{\rotatebox{90}{\textbf{Finetune}}}} & $\bm{r_{d}}$ & 0.018         & 0.34          & 0.075         & -0.048        & -0.115*       & 0.093         & -0.535        & -0.535        & 0.107         & -0.087        & -0.068        & 0.151*        & 0.268         & -             & -             & -0.098*       & -             & -             \\
\multicolumn{1}{c|}{}                                   & $\bm{r_{l}}$     & -0.097        & 0.074         & 0.083         & -0.162*       & -0.187*       & 0.175*        & -0.558        & -0.883*       & 0.324         & -0.161*       & -0.168*       & 0.247*        & 0.263         & -             & -             & -0.128*       & -             & -             \\
\multicolumn{1}{c|}{}                                   & $\bm{r_{da}}$     & 0.099         & 0.147         & 0.281         & 0.047         & 0.031         & -0.122*       & 0.642         & 0.321         & 0.393         & -0.011        & -0.015        & -0.052        & -0.433*       & -             & -             & 0.009         & -             & -             \\ \bottomrule
\end{tabular}

}
\caption{Correlation between $r$ and the performance (F1) of models in $M_{500}$ across different dataset domains and categories. Values highlighted with * represent statistical significant values with p-value $< 0.05$. For $\mathcal{D}_{f}$, we did not perform any evaluation strategy for ambiguous and hard instances, since the \#instances categorized as ambiguous and hard is very low, which is insufficient for training and evaluation, therefore values are not calculated and displayed.}
\label{tab:corr_hf_extended}
\end{table*}

\subsection{Granular analysis of model card sections}
\label{apx:sec_corr}

A more granular analysis extending \S\ref{sec:rq2} with respect to sections of model cards is presented in Table~\ref{tab:mc_sections_corr}.

\begin{table*}[!htbp]
\resizebox{\textwidth}{!}{
\centering

\begin{tabular}{@{}cccccccccccc@{}}
\toprule
\multicolumn{2}{c}{\multirow{2}{*}{\textbf{}}}                                   & \multicolumn{3}{c}{$\bm{\mathcal{D}_{a}}$}               & \multicolumn{3}{c}{$\bm{\mathcal{D}_{a}}$}              & $\bm{\mathcal{D}_{f}}$   & \multicolumn{3}{c}{\textbf{r}}           \\ \cmidrule(l){3-5}\cmidrule(l){6-8}\cmidrule(l){9-9}\cmidrule(l){10-12} 
\multicolumn{2}{c}{}                                                             & \textbf{Easy} & \textbf{Ambi} & \textbf{Hard} & \textbf{Easy} & \textbf{Ambi} & \textbf{Hard} & \textbf{Easy} & $\bm{r_{d}}$ & $\bm{r_{l}}$ & $\bm{r_{da}}$ \\ \midrule
\multicolumn{1}{c|}{\multirow{3}{*}{\textbf{Model Details}}}      & \textbf{FIA} & 0.173*        & 0.159*        & 0.064         & 0.132*        & 0.139*        & 0.135*        & 0.102*        & 0.603*      & 0.521*      & -0.039       \\
\multicolumn{1}{c|}{}                                             & \textbf{PIA} & 0.076         & 0.054         & -0.051        & 0.038         & 0.046         & 0.047         & -0.006        & 0.504*      & 0.408*      & -0.159*      \\
\multicolumn{1}{c|}{}                                             & \textbf{NIA} & -0.189*       & -0.173*       & -0.088        & -0.139*       & -0.160*       & -0.180*       & -0.139*       & -0.667*     & -0.521*     & 0.079        \\ \midrule
\multicolumn{1}{c|}{\multirow{3}{*}{\textbf{Training Details}}}   & \textbf{FIA} & 0.073         & 0.063         & 0.020         & 0.097*        & 0.067         & 0.017         & 0.074         & 0.384*      & 0.272*      & 0.112*       \\
\multicolumn{1}{c|}{}                                             & \textbf{PIA} & 0.052         & 0.038         & -0.055        & 0.047         & 0.027         & 0.030         & 0.074         & 0.343*      & 0.335*      & -0.152*      \\
\multicolumn{1}{c|}{}                                             & \textbf{NIA} & -0.079        & -0.078        & -0.015        & -0.110*       & -0.085        & -0.038        & -0.105*       & -0.448*     & -0.327*     & -0.019       \\ \midrule
\multicolumn{1}{c|}{\multirow{3}{*}{\textbf{Evaluation Details}}} & \textbf{FIA} & 0.119*        & 0.133*        & 0.048         & 0.093*        & 0.101*        & 0.125*        & 0.090         & 0.199*      & 0.147*      & -0.081       \\
\multicolumn{1}{c|}{}                                             & \textbf{PIA} & 0.099*        & 0.078         & -0.069        & 0.073         & 0.051         & 0.032         & 0.083         & 0.289*      & 0.307*      & -0.102*      \\
\multicolumn{1}{c|}{}                                             & \textbf{NIA} & -0.165*       & -0.160*       & -0.011        & -0.134*       & -0.140*       & -0.116*       & -0.152*       & -0.305*     & -0.226*     & 0.067        \\ \midrule
\multicolumn{1}{c|}{\multirow{3}{*}{\textbf{Uses}}}               & \textbf{FIA} & 0.137*        & 0.112*        & -0.082        & 0.109*        & 0.088         & 0.087         & 0.049         & 0.501*      & 0.444*      & -0.171*      \\
\multicolumn{1}{c|}{}                                             & \textbf{PIA} & 0.118*        & 0.090         & -0.079        & 0.098*        & 0.082         & 0.039         & 0.128*        & 0.278*      & 0.280*      & -0.118*      \\
\multicolumn{1}{c|}{}                                             & \textbf{NIA} & -0.176*       & -0.149*       & 0.068         & -0.144*       & -0.121*       & -0.119*       & -0.094*       & -0.458*     & -0.444*     & 0.195*       \\ \midrule
\multicolumn{1}{c|}{\multirow{3}{*}{\textbf{Other Details}}}      & \textbf{FIA} & 0.146*        & 0.128*        & -0.015        & 0.131*        & 0.114*        & 0.081         & 0.070         & 0.411*      & 0.334*      & -0.173*      \\
\multicolumn{1}{c|}{}                                             & \textbf{PIA} & 0.079         & 0.060         & -0.043        & 0.062         & 0.047         & 0.060         & 0.110*        & 0.327*      & 0.344*      & -0.088*      \\
\multicolumn{1}{c|}{}                                             & \textbf{NIA} & -0.028        & -0.019        & 0.088         & -0.024        & -0.009        & 0.011         & -0.033        & -0.068      & -0.095*     & 0.012        \\ \bottomrule
\end{tabular}
}
\caption{Correlation values computed between the percentage of FIA/PIA/NIA (across all sections of the model card) and model performance (F1). Additionally, it includes correlation values calculated between the percentage of FIA/PIA/NIA and $r$ (model popularity). Values highlighted with * represent statistical significant values with p-value $<0.05$. For $\mathcal{D}_{f}$, we did not perform any evaluation strategy for ambiguous and hard instances, since the \#instances categorized as ambiguous and hard is very low, which is insufficient for training and evaluation, therefore values are not calculated and displayed.}
\label{tab:mc_sections_corr}
\end{table*}

\section{Annotator Information}
\label{sec:ann_info}
We recruited annotators for two tasks in our study. Task 1 involved the Manual Inspection study outlined in \S\ref{sec:mcards}, while Task 2 focused on curating the Reddit dataset as described in \S\ref{sec:case_study}. We invited undergraduate students to volunteer for these tasks, ensuring their anonymity and providing no compensation. Detailed annotation instructions can be found in \S\ref{apx:annot_instruct} for Task 2. Annotators were not given strict time limits for annotation. They were proficient in English and familiar with the tasks. For Task 2, in the event of a tie between annotator labels, we randomly selected a label based on their order of labeling as the final decision.

\section{Hyperparameters}
\label{apx:hp}

We did hyperparameter tuning on RoBERTa model for all datasets in $\mathcal{D}$, the optimal parameters are detailed in 
Table~\ref{tab:hparams}. The optimization process employs the AdamW optimizer \cite{loshchilov2018decoupled}. To ensure reproducibility, we conduct all our experiments using the open-source HF Transformers library\footnote{\url{https://huggingface.co/docs/transformers/}} \cite{wolf-etal-2020-transformers}. Additionally, all experiments are executed using 4 $\times$ Tesla V100 GPUs.

\section{Performance Plots}
\label{apx:pp}
\begin{figure*}
    \centering
    \begin{subfigure}[b]{0.95\textwidth}
        \centering
        \includegraphics[width=\textwidth]{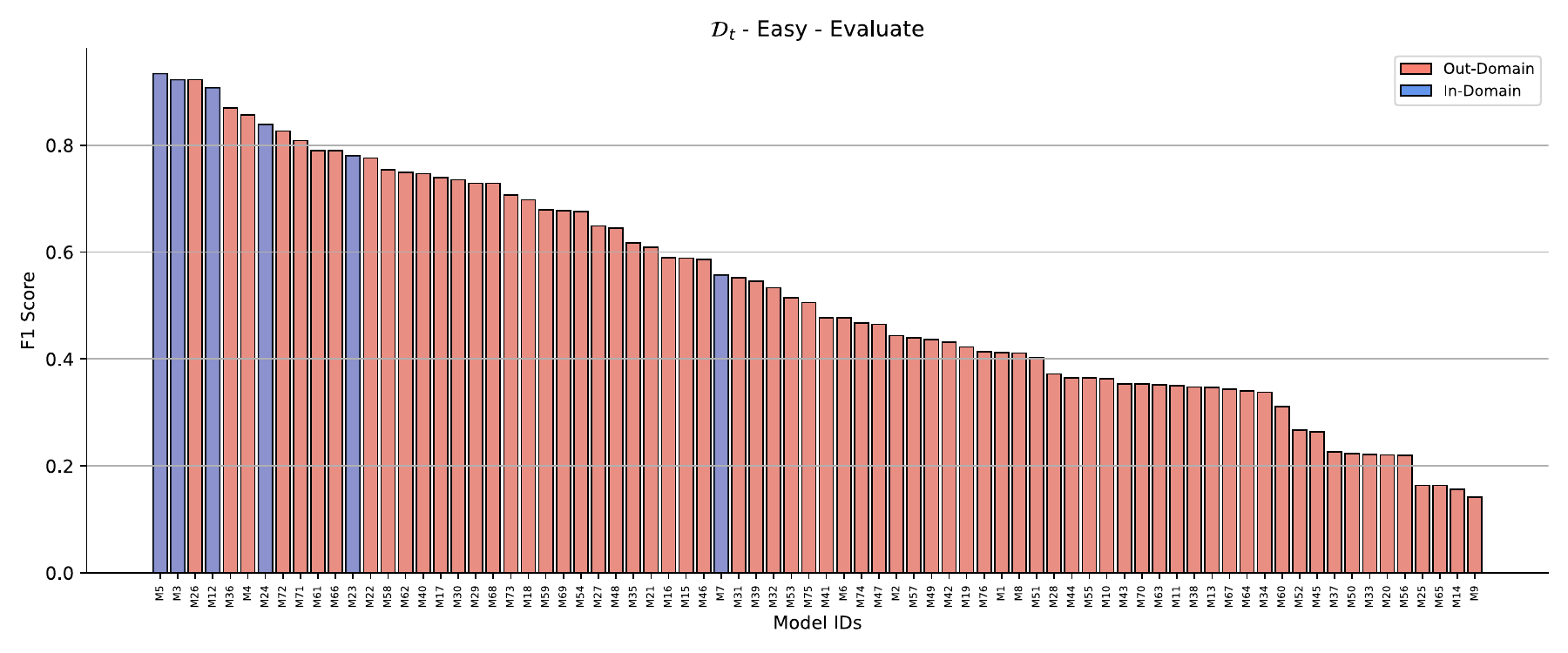}
    \end{subfigure}
    \hfill
    \begin{subfigure}[b]{\textwidth}
        \centering
        \includegraphics[width=\textwidth]{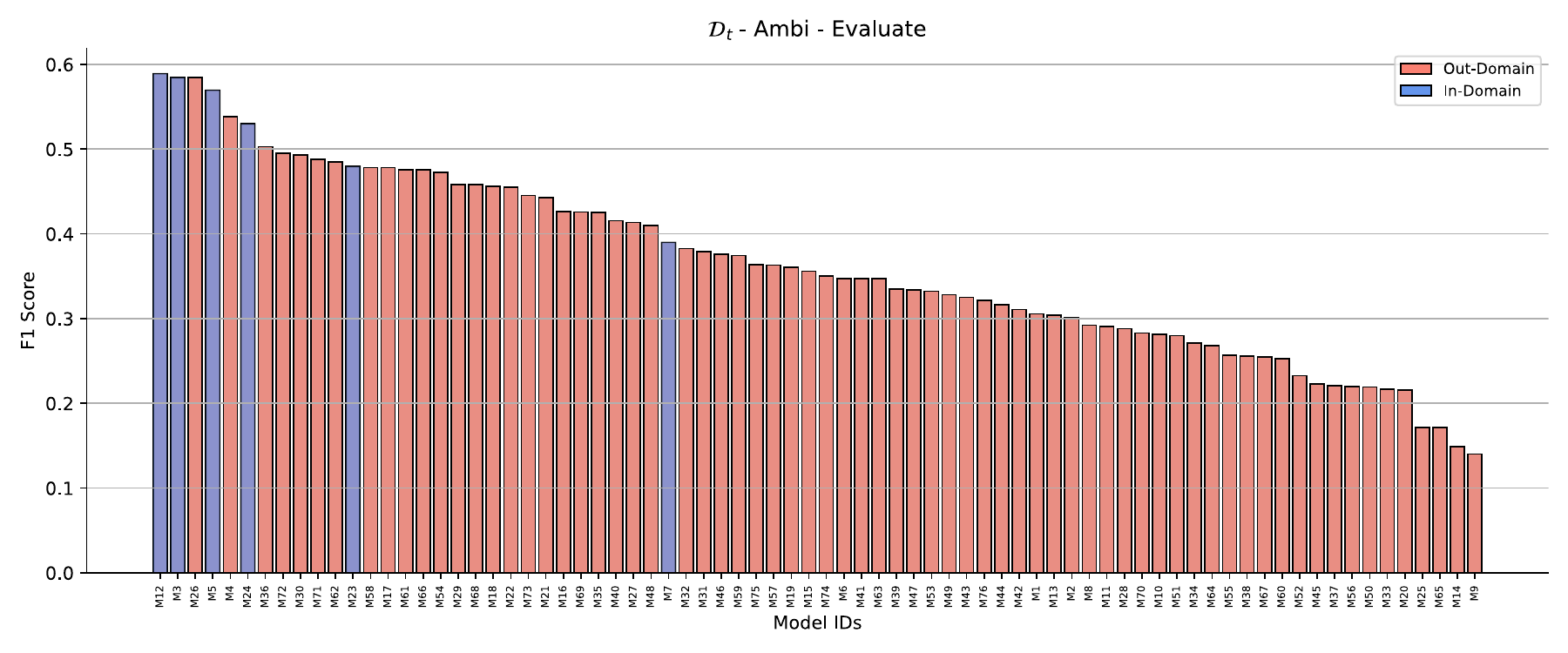}
    \end{subfigure}
    \hfill
    \begin{subfigure}[b]{\textwidth}
        \centering
        \includegraphics[width=\textwidth]{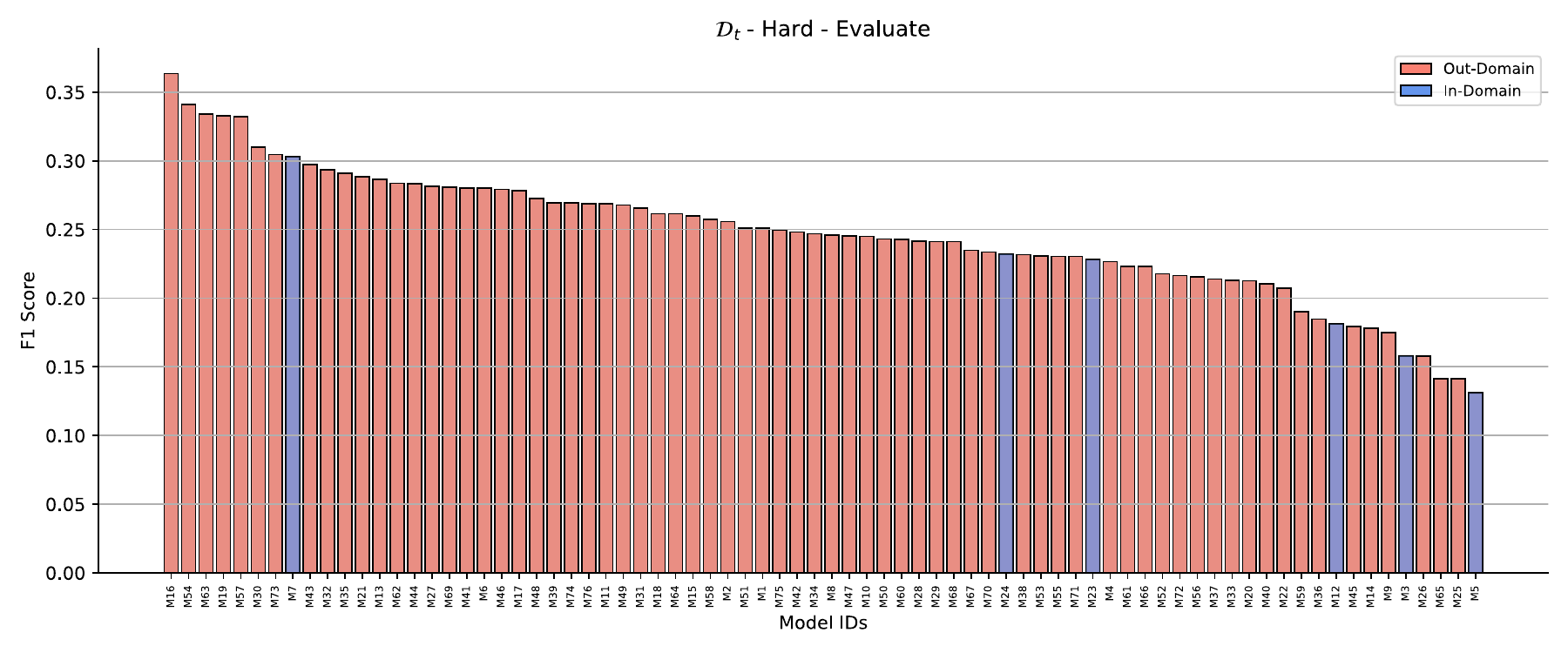} 
    \end{subfigure}
    \caption{TweetEval -- Evaluate}
    \label{fig:te_e}
\end{figure*}

Refer to Figure~\ref{fig:te_e} and other subsequent plots.

\begin{figure*}
    \centering
    \begin{subfigure}[b]{\textwidth}
        \centering
        \includegraphics[width=0.95\textwidth]{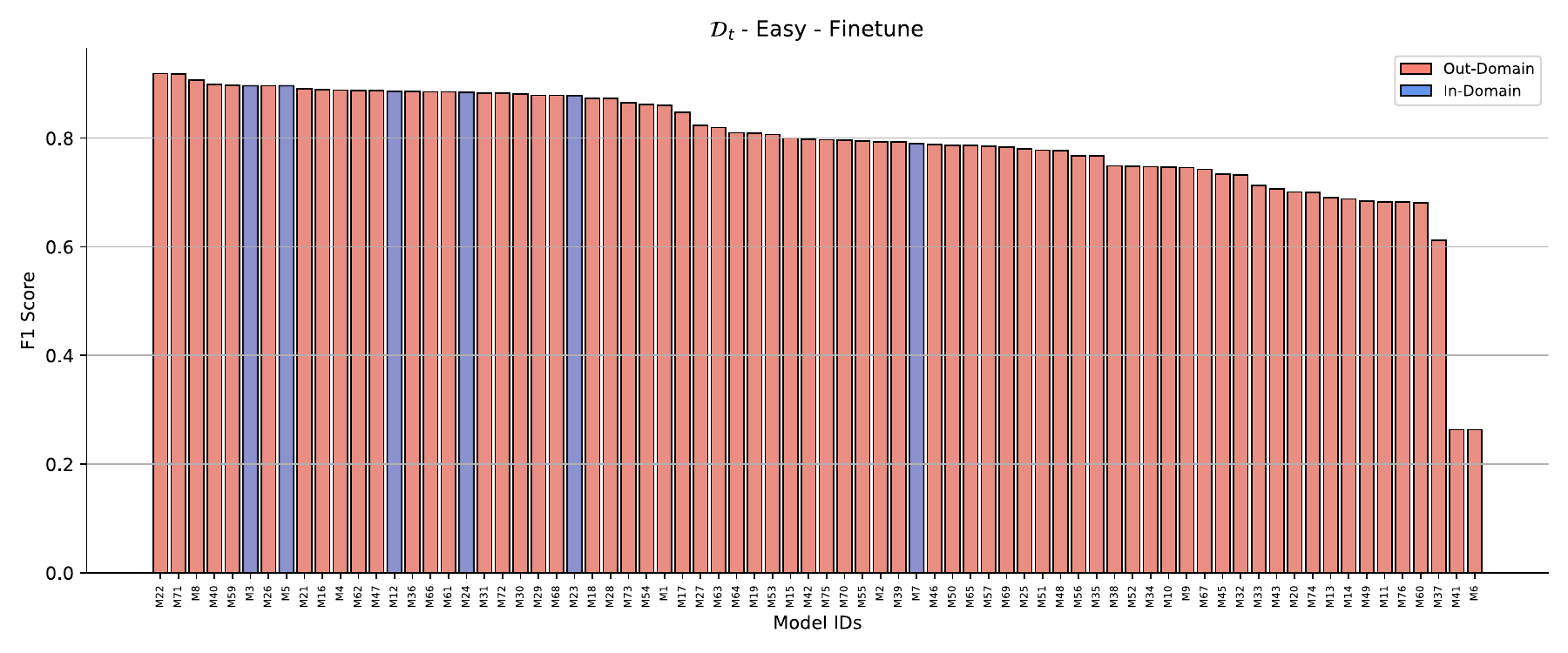}
    \end{subfigure}
    \hfill
    \begin{subfigure}[b]{\textwidth}
        \centering
        \includegraphics[width=\textwidth]{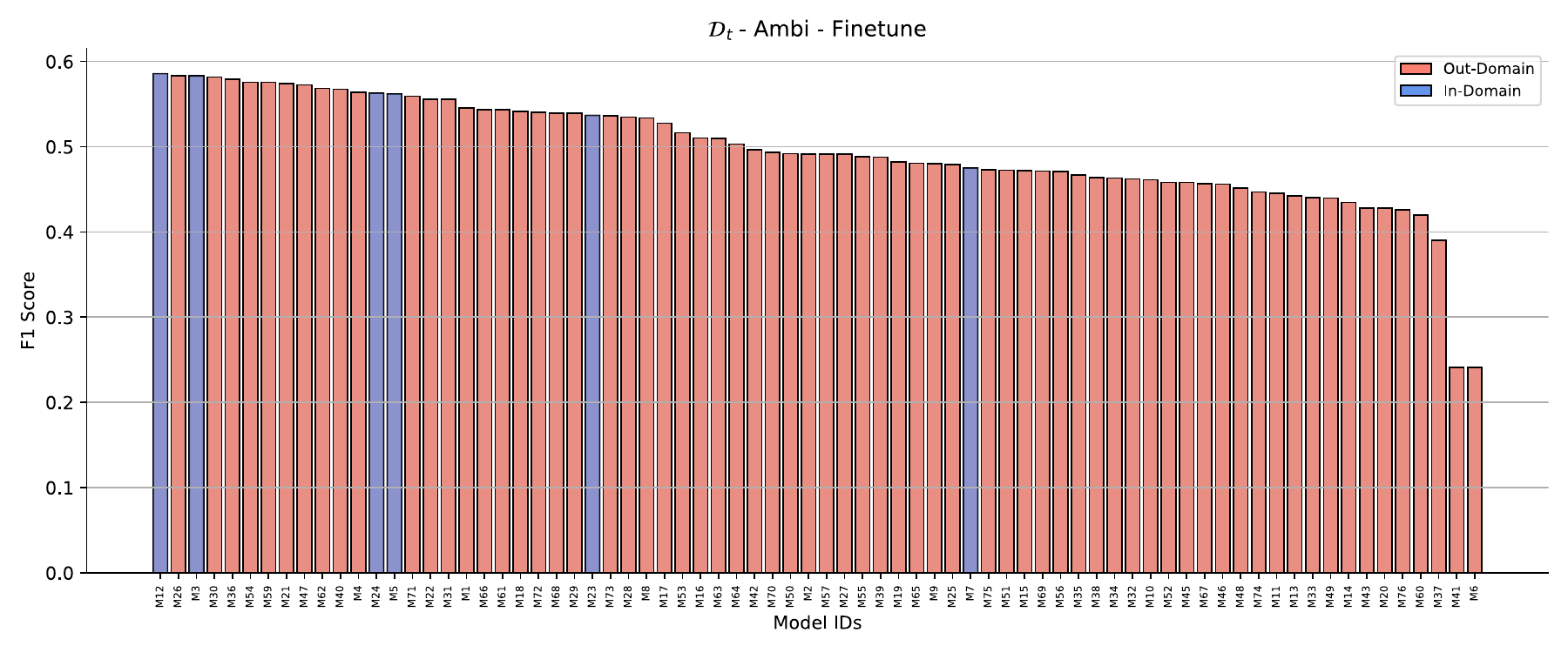}
    \end{subfigure}
    \hfill
    \begin{subfigure}[b]{\textwidth}
        \centering
        \includegraphics[width=\textwidth]{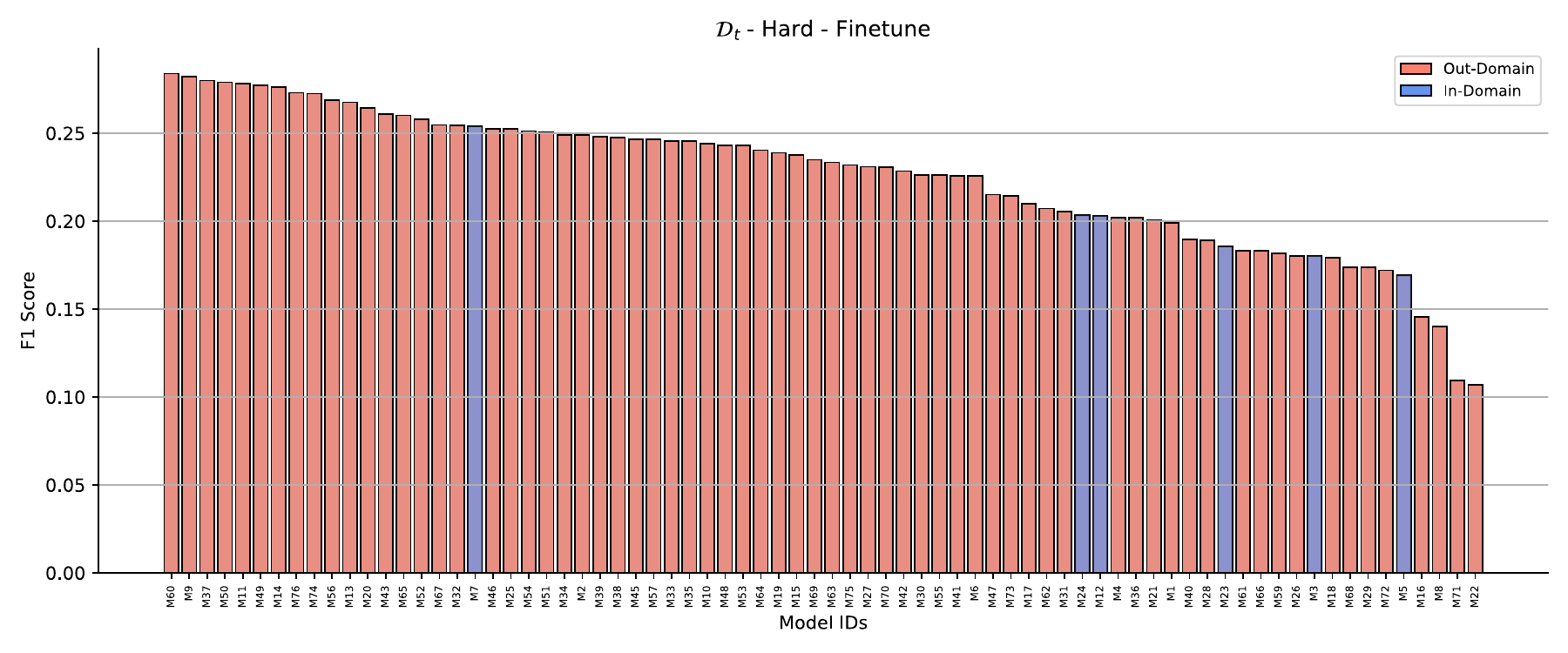} 
    \end{subfigure}
    \caption{TweetEval -- Finetune}
    \label{fig:te_f}
\end{figure*}

\begin{figure*}
    \centering
    \begin{subfigure}[b]{0.95\textwidth}
        \centering
        \includegraphics[width=\textwidth]{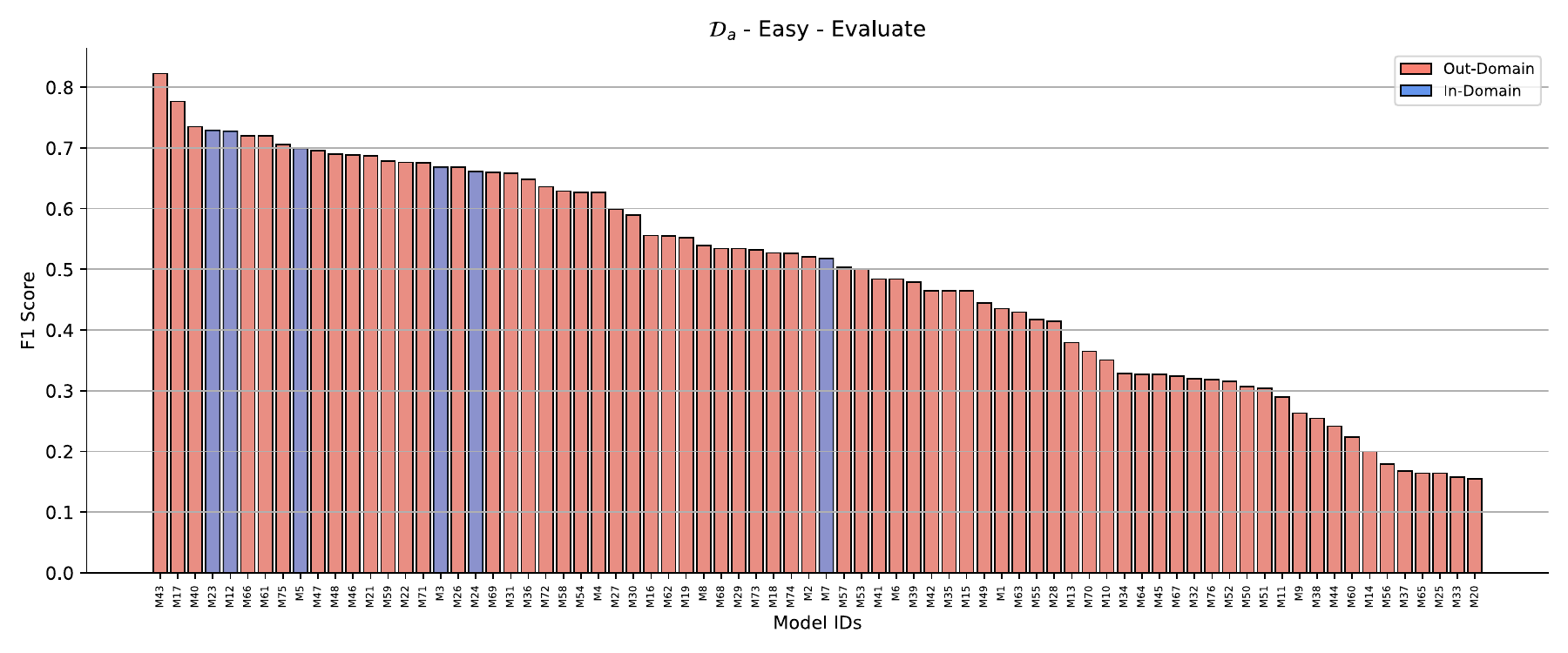}
    \end{subfigure}
    \hfill
    \begin{subfigure}[b]{\textwidth}
        \centering
        \includegraphics[width=\textwidth]{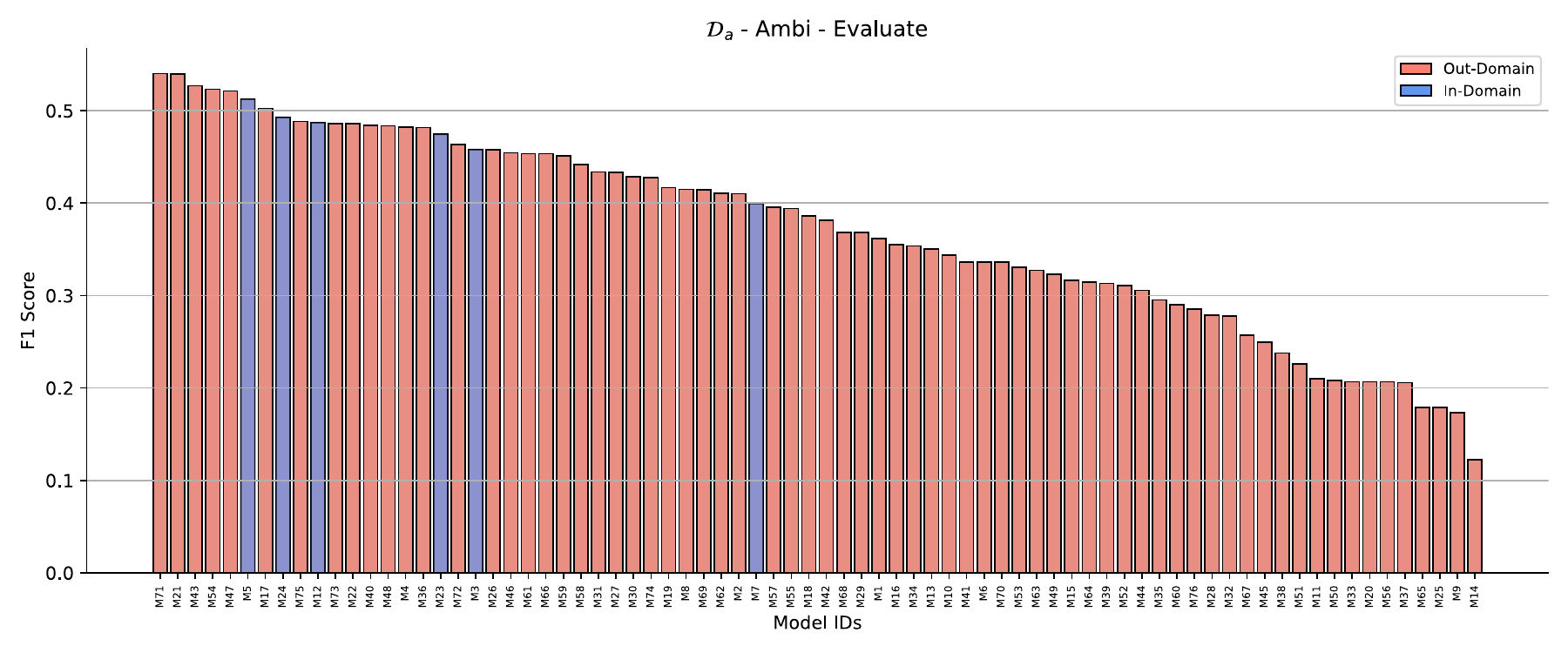}
    \end{subfigure}
    \hfill
    \begin{subfigure}[b]{\textwidth}
        \centering
        \includegraphics[width=\textwidth]{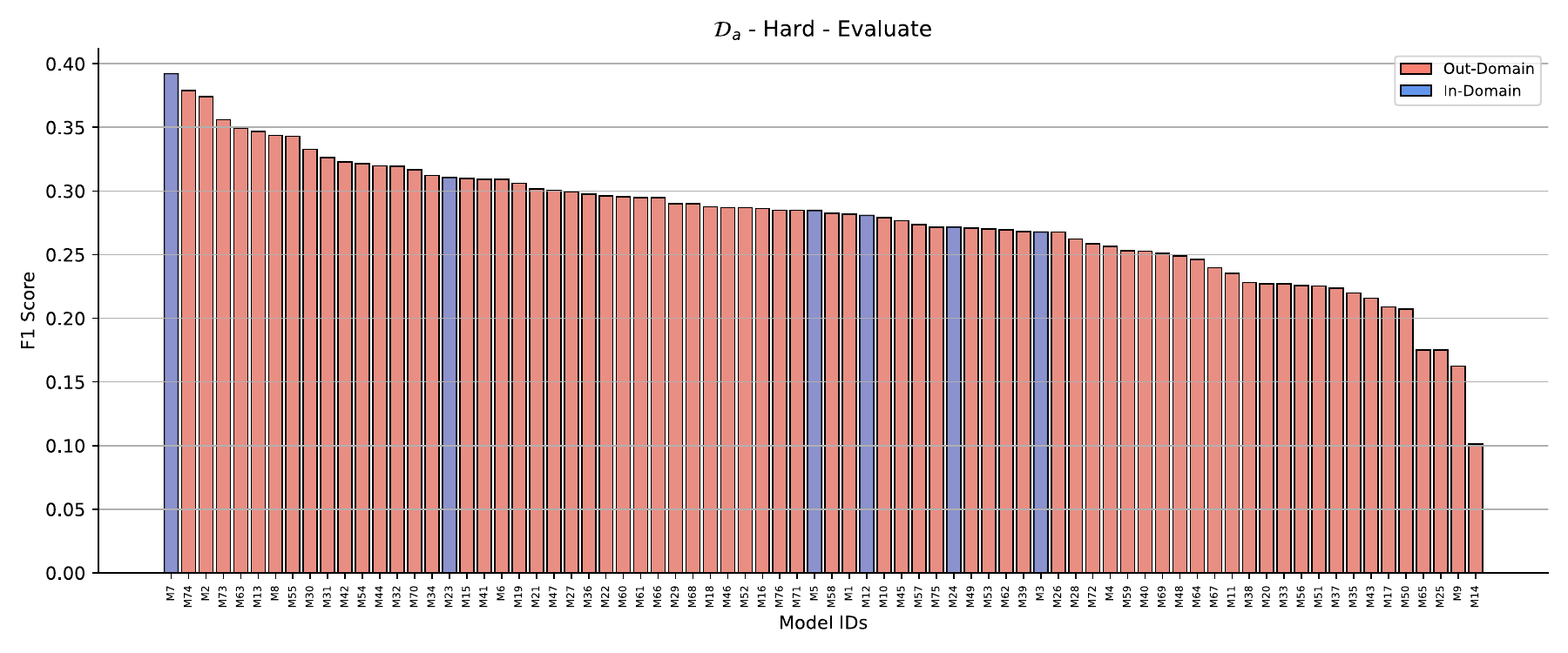} 
    \end{subfigure}
    \caption{Amazon Multi Reviews -- Evaluate}
    \label{fig:amr_e}
\end{figure*}

\begin{figure*}
    \centering
    \begin{subfigure}[b]{0.95\textwidth}
        \centering
        \includegraphics[width=\textwidth]{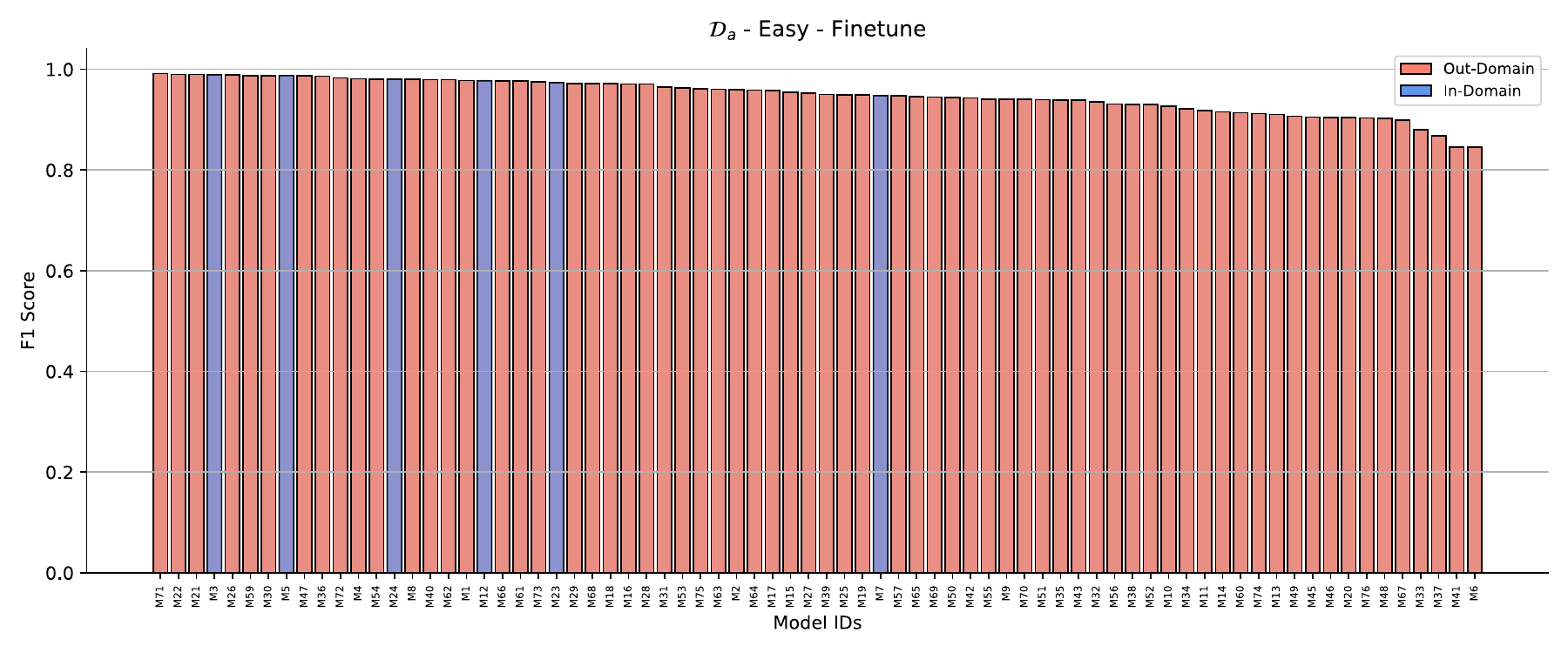}
    \end{subfigure}
    \hfill
    \begin{subfigure}[b]{\textwidth}
        \centering
        \includegraphics[width=\textwidth]{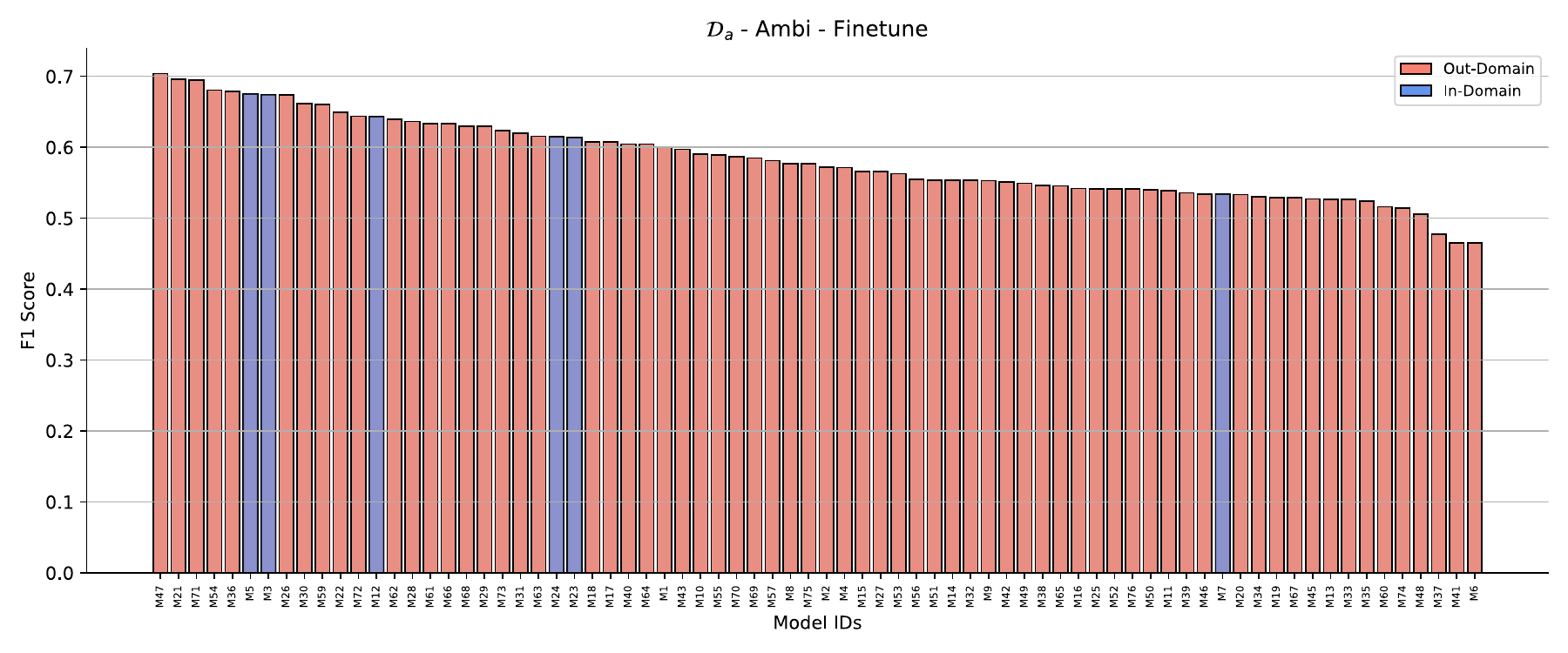}
    \end{subfigure}
    \hfill
    \begin{subfigure}[b]{\textwidth}
        \centering
        \includegraphics[width=\textwidth]{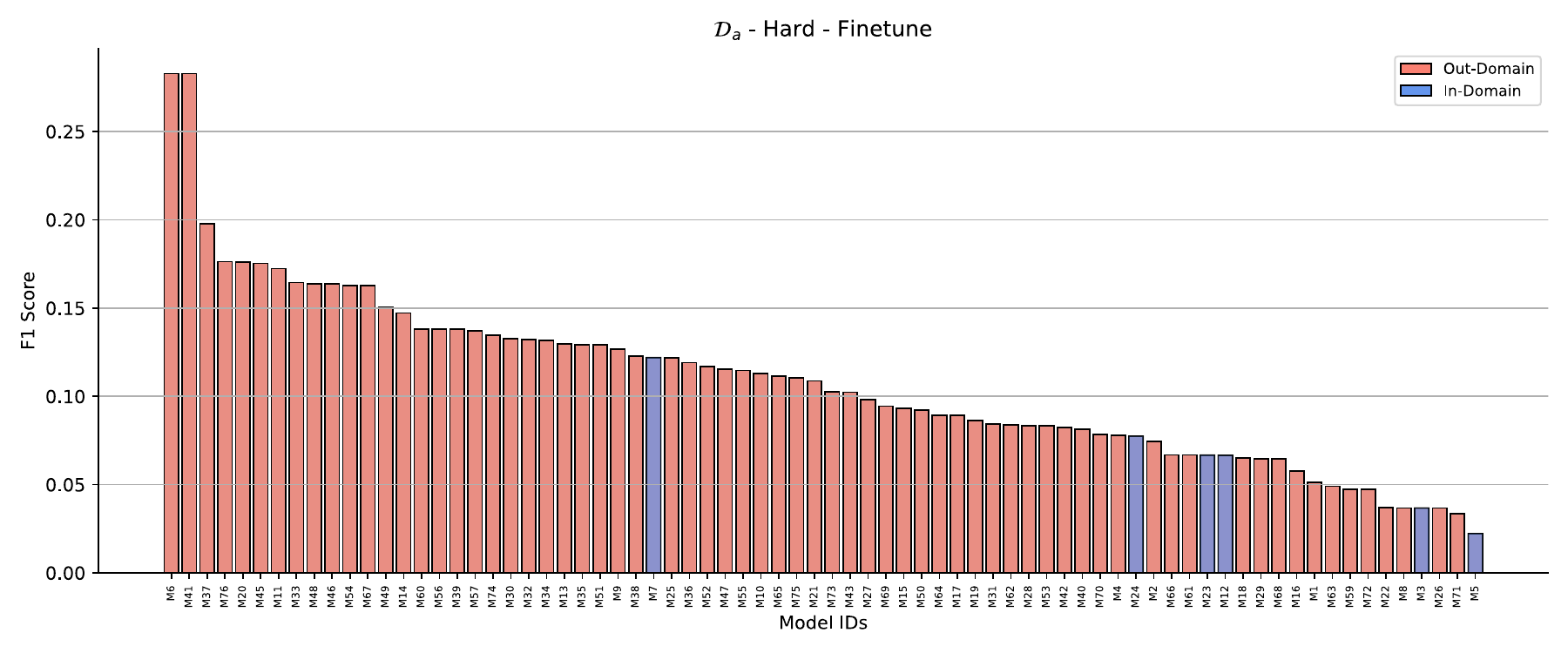} 
    \end{subfigure}
    \caption{Amazon Multi Reviews -- Evaluate}
    \label{fig:amr_f}
\end{figure*}

\begin{figure*}
    \centering
    \begin{subfigure}[b]{\textwidth}
        \centering
        \includegraphics[width=\textwidth]{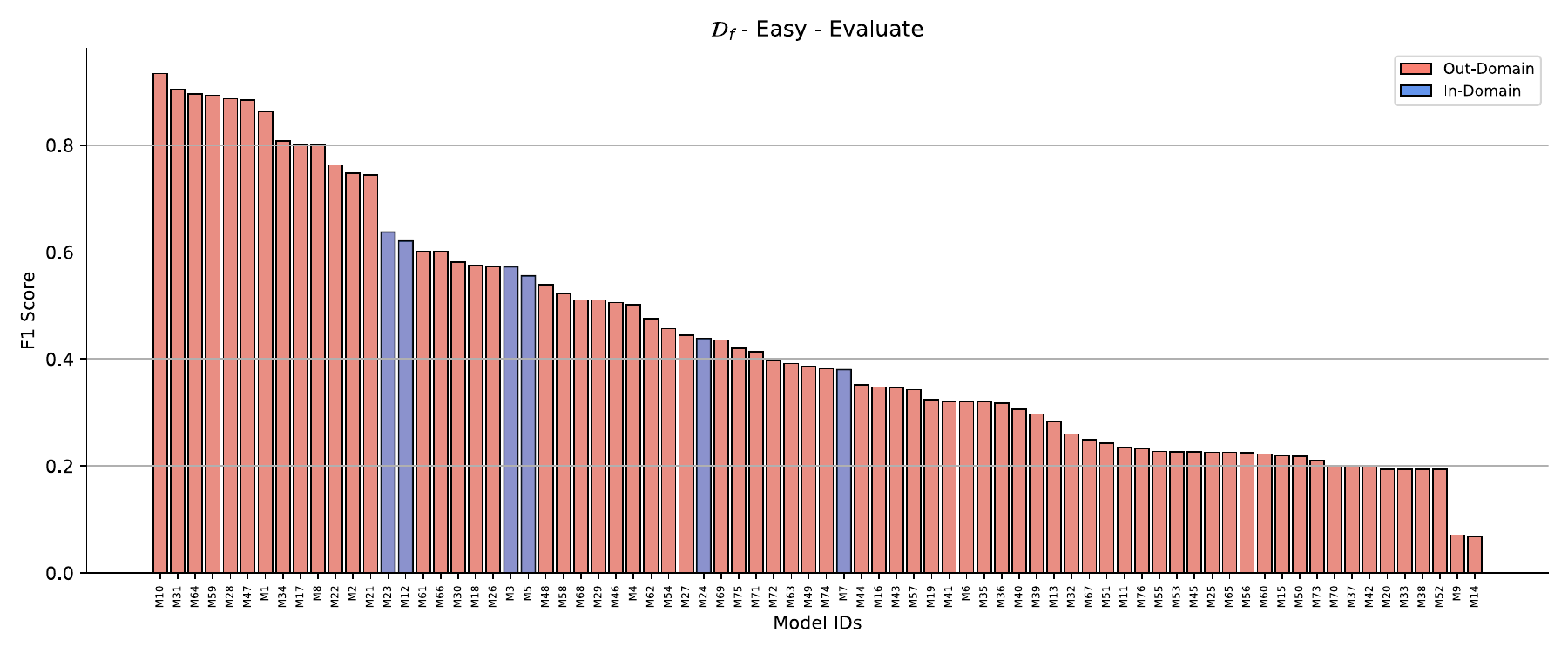}
    \end{subfigure}
    \hfill
    \begin{subfigure}[b]{\textwidth}
        \centering
        \includegraphics[width=\textwidth]{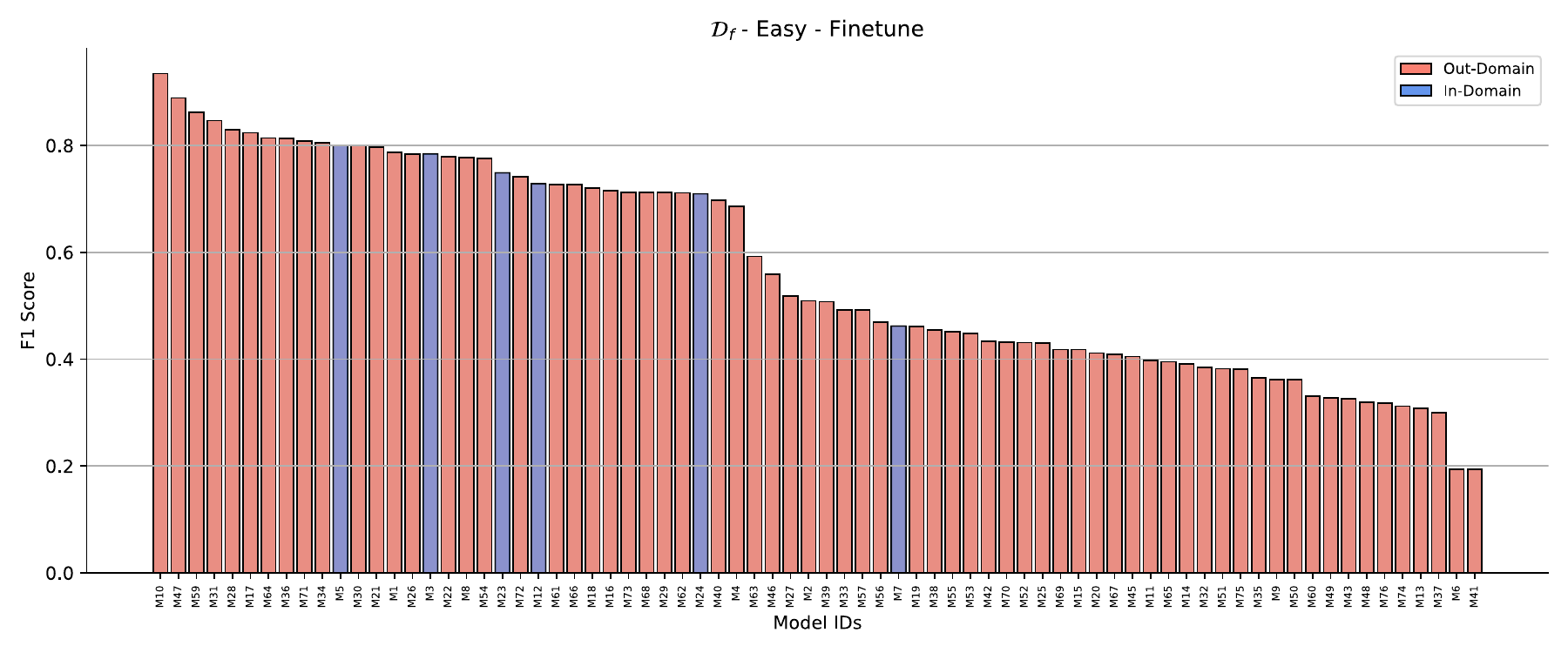}
    \end{subfigure}
    \caption{Financial Phrasebank -- Evaluate -- Finetune}
\end{figure*}

\section{Annotation Instructions}
\label{apx:annot_instruct}
\begin{figure*}[!htb]
    \centering
    \includegraphics[width=\textwidth]{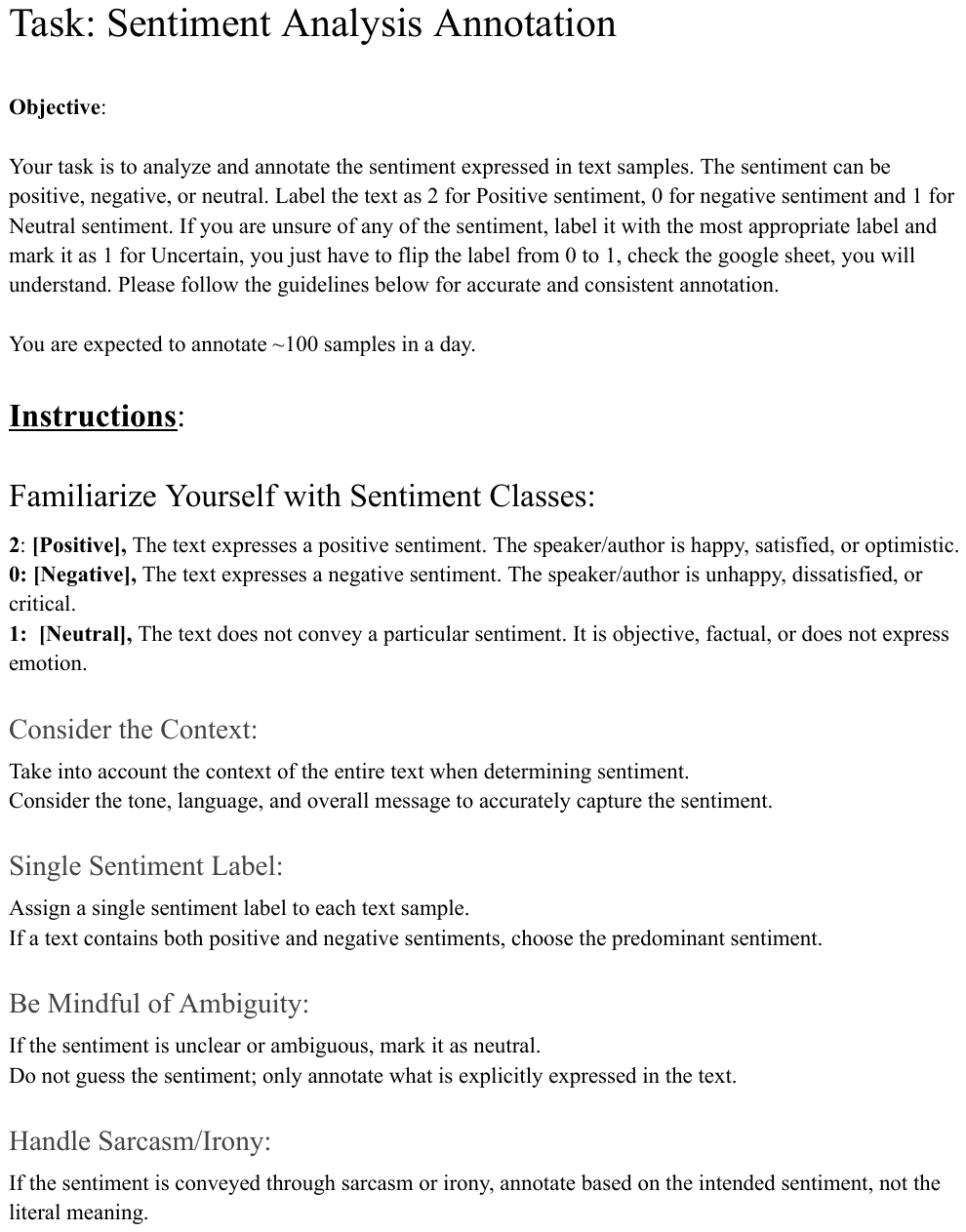}
    \caption{Annotation Instructions (Page 1)}
    \label{fig:annot1}
    \vspace{-25.26385pt}
\end{figure*}

Refer to Figure~\ref{fig:annot1} for annotation instructions.

\begin{figure*}[!htb]
    \centering
    \includegraphics[width=\textwidth]{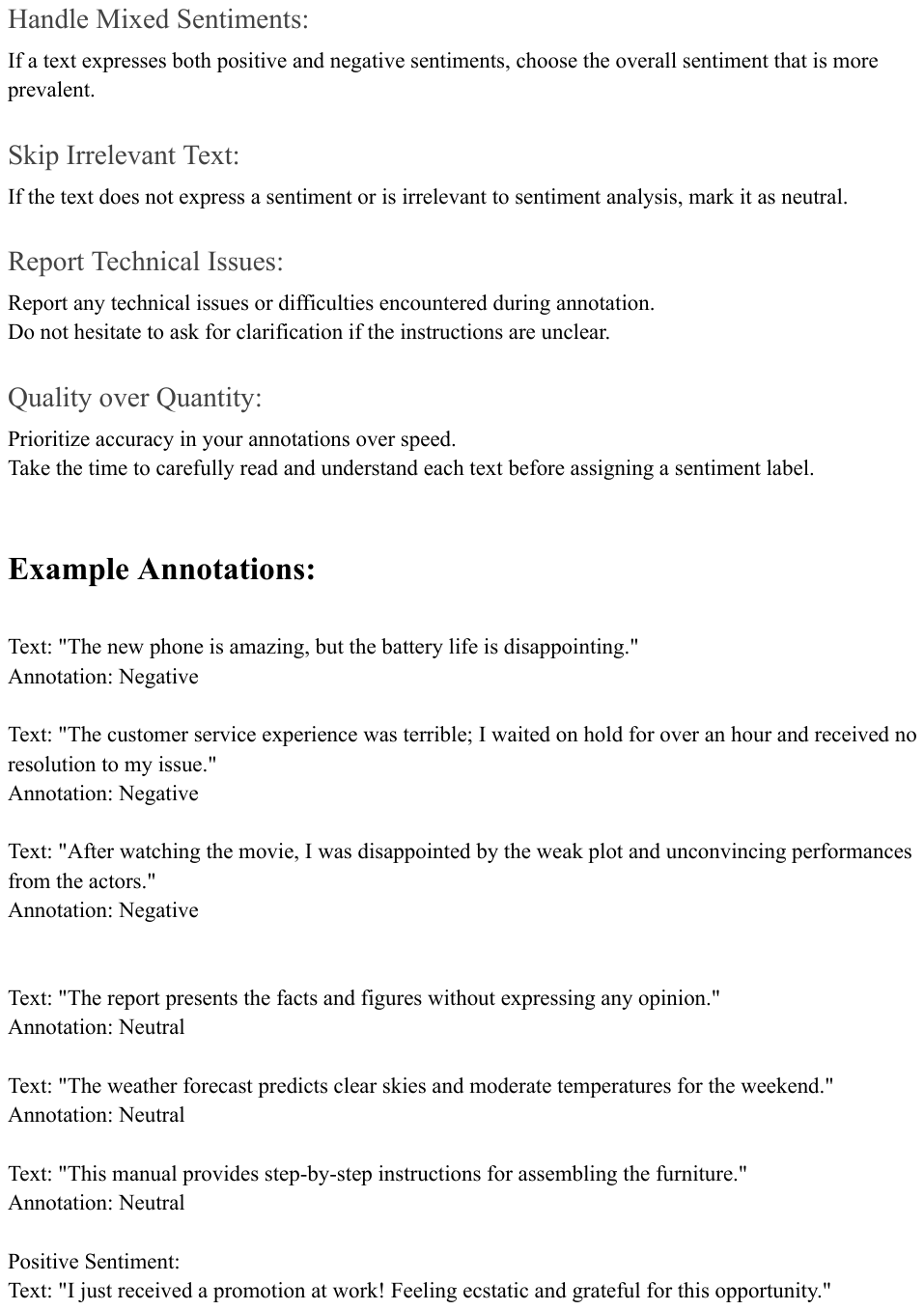}
    \caption{Annotation Instructions (Page 2)}
    \label{fig:annot2}
    \vspace{-25.26385pt}
\end{figure*}

\begin{figure*}[!htb]
    \centering
    \includegraphics[width=\textwidth]{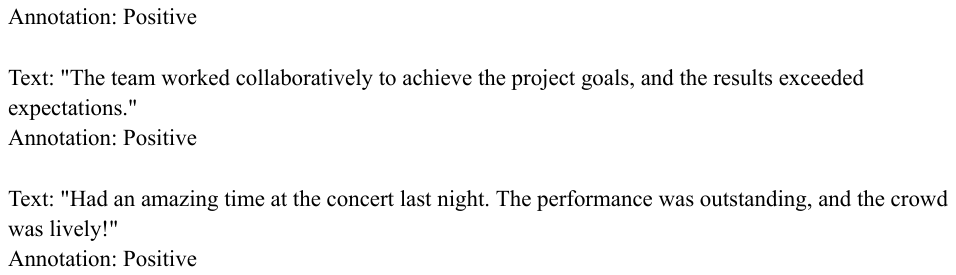}
    \caption{Annotation Instructions (Page 3)}
    \label{fig:annot3}
    \vspace{-25.26385pt}
\end{figure*}




\begin{figure*}[!htb]
    \centering
    \begin{subfigure}{\textwidth}
        \centering
        \includegraphics[width=0.48\textwidth]{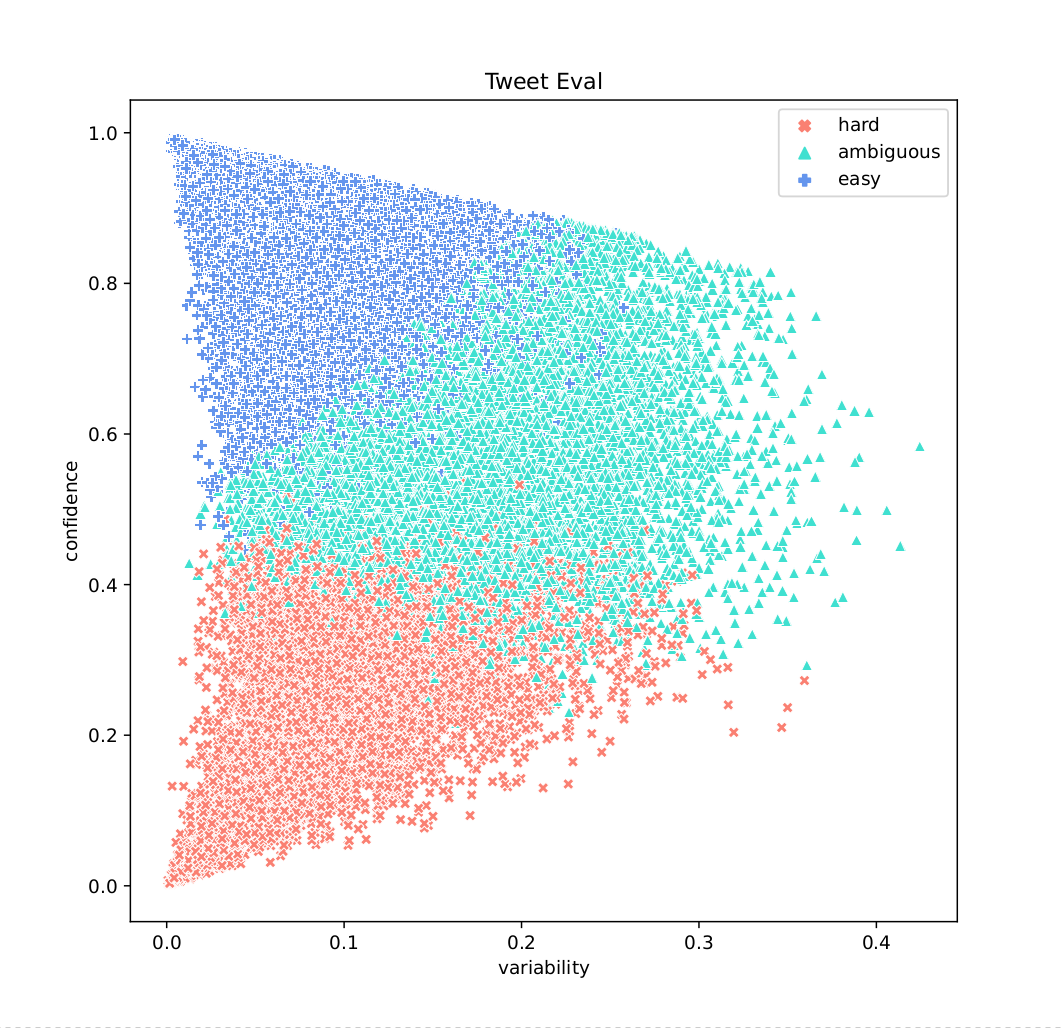}
        \hfill
        \includegraphics[width=0.48\textwidth]{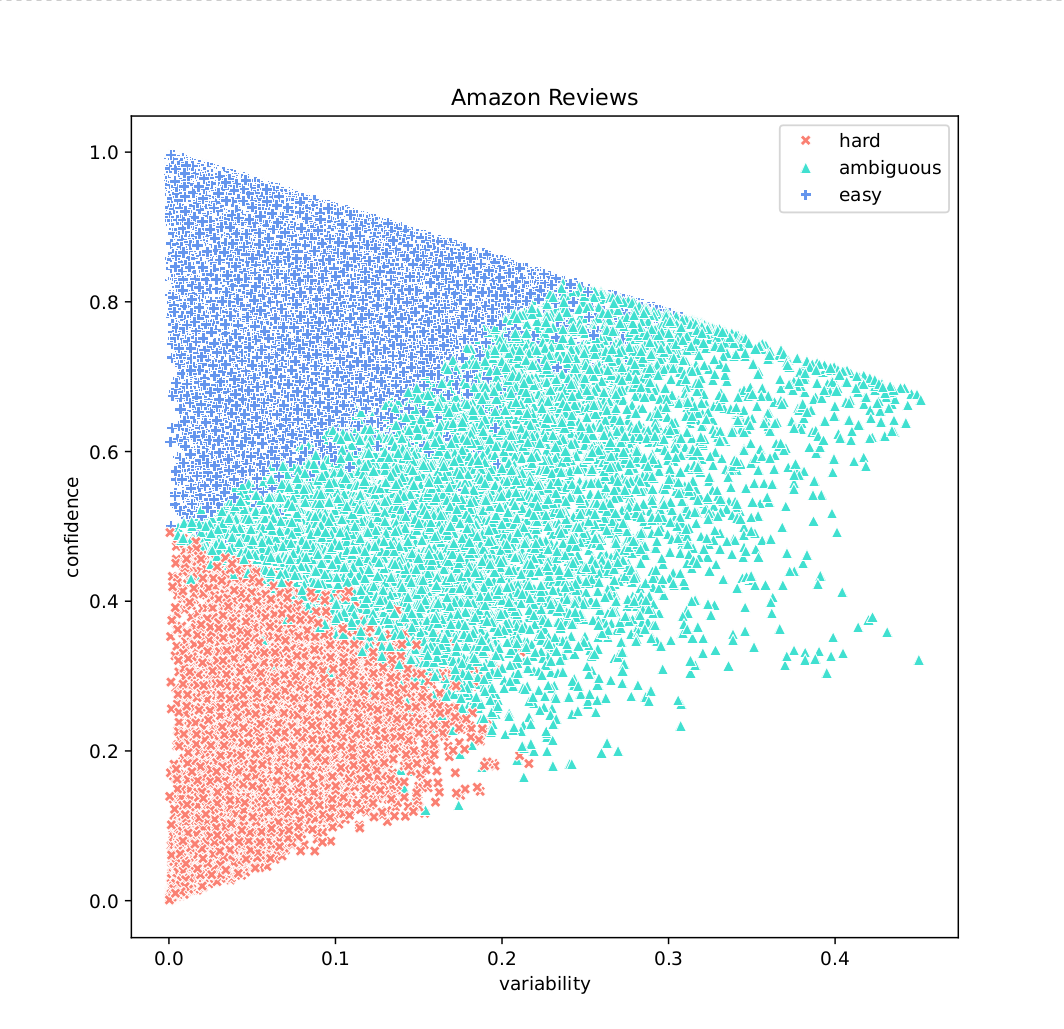}
    \end{subfigure}
    
    \vspace{\floatsep} 
    
    \begin{subfigure}{\textwidth}
        \centering
        \includegraphics[width=0.48\textwidth]{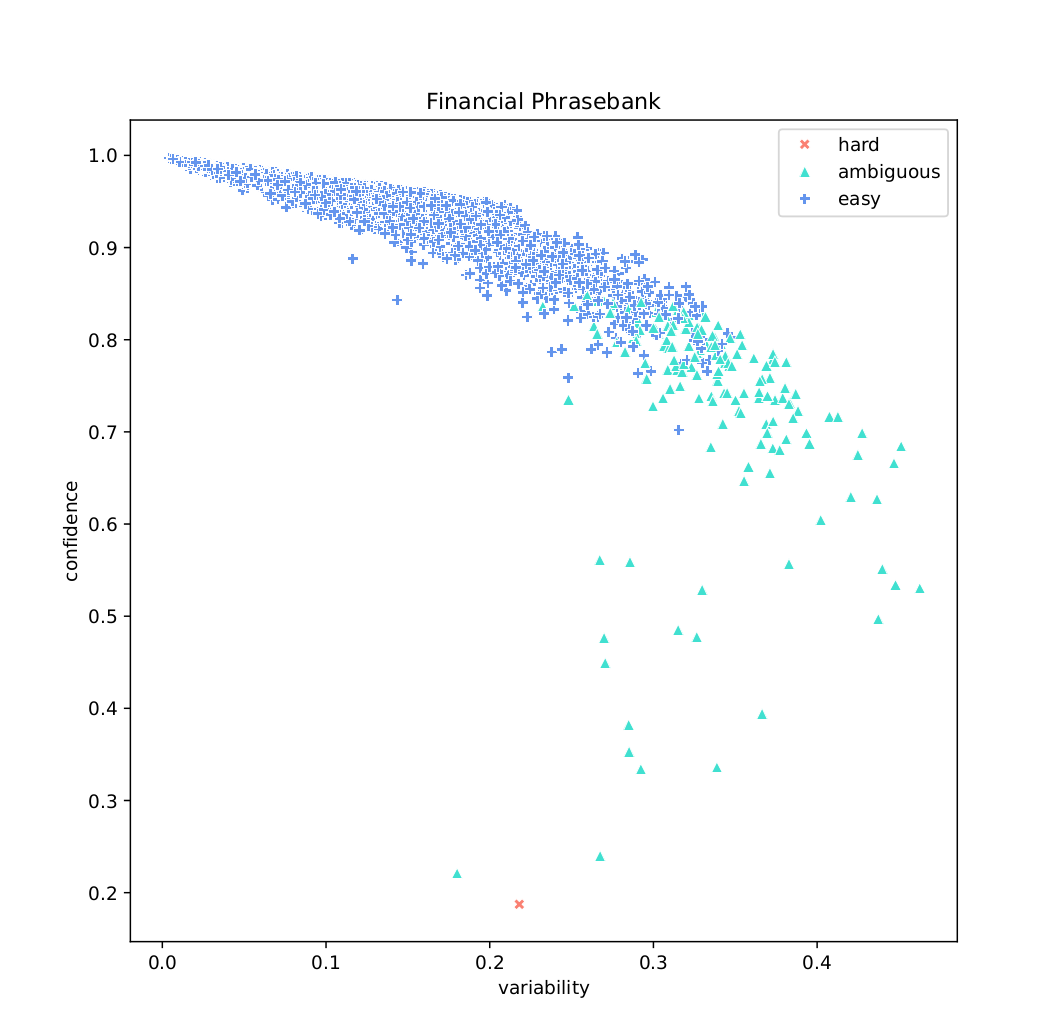}
        \label{fig:fp_cart}
    \end{subfigure}

    \caption{Datamaps across $\mathcal{D}$}
    \label{fig:cart_all}
\end{figure*}

\end{document}